\PassOptionsToPackage{table}{xcolor}
\documentclass[acmsmall,screen,nonacm]{acmart}

\usepackage[utf8]{inputenc}
\usepackage[T1]{fontenc}
\usepackage{microtype}
\usepackage{graphicx}
\usepackage{wrapfig}
\usepackage{booktabs}
\usepackage{nicefrac}
\usepackage{mathtools}
\usepackage[capitalize,noabbrev]{cleveref}
\usepackage{algorithm}
\usepackage{algorithmic}

\usepackage[normalem]{ulem}
\usepackage{float,array}
\usepackage[export]{adjustbox}
\usepackage{tikz}
\usetikzlibrary{fit,calc}
\usepackage{multirow,placeins,rotating,tabularx}

\newcolumntype{V}{>{\centering\arraybackslash}m{2.8em}}
\newcolumntype{K}{>{\centering\arraybackslash}X}
\newcommand{\vlabel}[1]{\adjustbox{valign=m}{\rotatebox{90}{\scalebox{0.8}{\begin{tabular}{@{}c@{}}\tiny\bfseries #1\end{tabular}}}}}
\newcommand{\imgcell}[1]{\begingroup\catcode`\_=12 \catcode`\~=12 \catcode`\#=12\adjustbox{valign=m}{\includegraphics[width=\linewidth]{#1}}\endgroup}
\newcommand{\imgnode}[2]{\begingroup\catcode`\_=12 \catcode`\~=12 \catcode`\#=12\adjustbox{valign=m}{\begin{tikzpicture}[baseline=(current bounding box.center),remember picture]\node[inner sep=0pt,outer sep=0pt](#1){\includegraphics[width=\linewidth]{#2}};\end{tikzpicture}}\endgroup}
\newcommand{\drawcolbox}[2]{\begin{tikzpicture}[overlay,remember picture]\draw[red,line width=1.5pt]([xshift=0.75pt,yshift=-0.75pt]#1.north west) rectangle ([xshift=-0.75pt,yshift=0.75pt]#2.south east);\end{tikzpicture}}

\newlength{\cellw}
\renewcommand{\arraystretch}{1.0}
\newcolumntype{C}{>{\centering\arraybackslash}p{\cellw}}
\newlength{\rowsep}
\newlength{\boxlw}
\newlength{\boxinset}
\setcopyright{none}
\renewcommand\footnotetextcopyrightpermission[1]{}

\title{Object-Aware Background-Controlled Editing via Weighted Velocity Guidance}

\author{Wuji Wang}
\affiliation{
  \department{Language Technologies Institute, School of Computer Science}
  \institution{Carnegie Mellon University}
  \city{Pittsburgh}
  \state{Pennsylvania}
  \country{USA}
}
\email{wujiw@andrew.cmu.edu}

\author{Yue Wu}
\affiliation{
  \department{State Key Laboratory of AI Safety}
  \institution{Institute of Computing Technology, Chinese Academy of Sciences}
  \city{Beijing}
  \country{China}
}
\email{wuyue221@mails.ucas.ac.cn}

\author{Chenhao Yi}
\affiliation{
  \department{Department of Computer Science and Technology}
  \institution{University of Chinese Academy of Sciences}
  \city{Beijing}
  \country{China}
}
\email{yichenhao23@mails.ucas.ac.cn}

\author{Shuhui Wang}
\authornote{Corresponding author.}
\affiliation{
  \department{State Key Laboratory of AI Safety}
  \institution{Institute of Computing Technology, Chinese Academy of Sciences}
  \city{Beijing}
  \country{China}
}
\email{wangshuhui@ict.ac.cn}

\renewcommand{\shortauthors}{Wang et al.}

\begin{document}
\begin{abstract}
Training-free image editing steers diffusion or flow-matching generative models
at inference time by modifying prompt-conditioned denoising velocities.
Existing velocity-based editors often apply prompt-induced residuals globally
over the latent space and rely on the model to localize semantic changes
implicitly.
For object-centric edits, these residuals are rarely zero outside the target
object, so small non-target components can accumulate during multi-step
integration, causing background drift and unstable object boundaries.
We propose \textbf{Object-Aware Velocity Control} (OAVC), a training-free
framework that introduces object-level control into the velocity-integration
process.
OAVC decouples \emph{where} semantic residuals are allowed to act from
\emph{how} they are injected into the dynamics.
It constructs a background-anchored reference interface under the source prompt
and then performs object-localized safe semantic injection under the target
prompt.
A constrained injection operator suppresses drift-inducing velocity components,
while time-adaptive spatial weighting stabilizes the transition near object
boundaries.
OAVC requires no training or modification of pretrained model parameters.
Experiments on object-centric image and video benchmarks with image and video rectified-flow backbones
show improved background preservation, structural fidelity, boundary stability,
and temporal consistency while retaining effective localized editability.
Code is available at \url{https://github.com/wwj-code/OAVC}.
\end{abstract}
\maketitle

\begin{figure}[t]
  \centering
  \includegraphics[width=\linewidth]{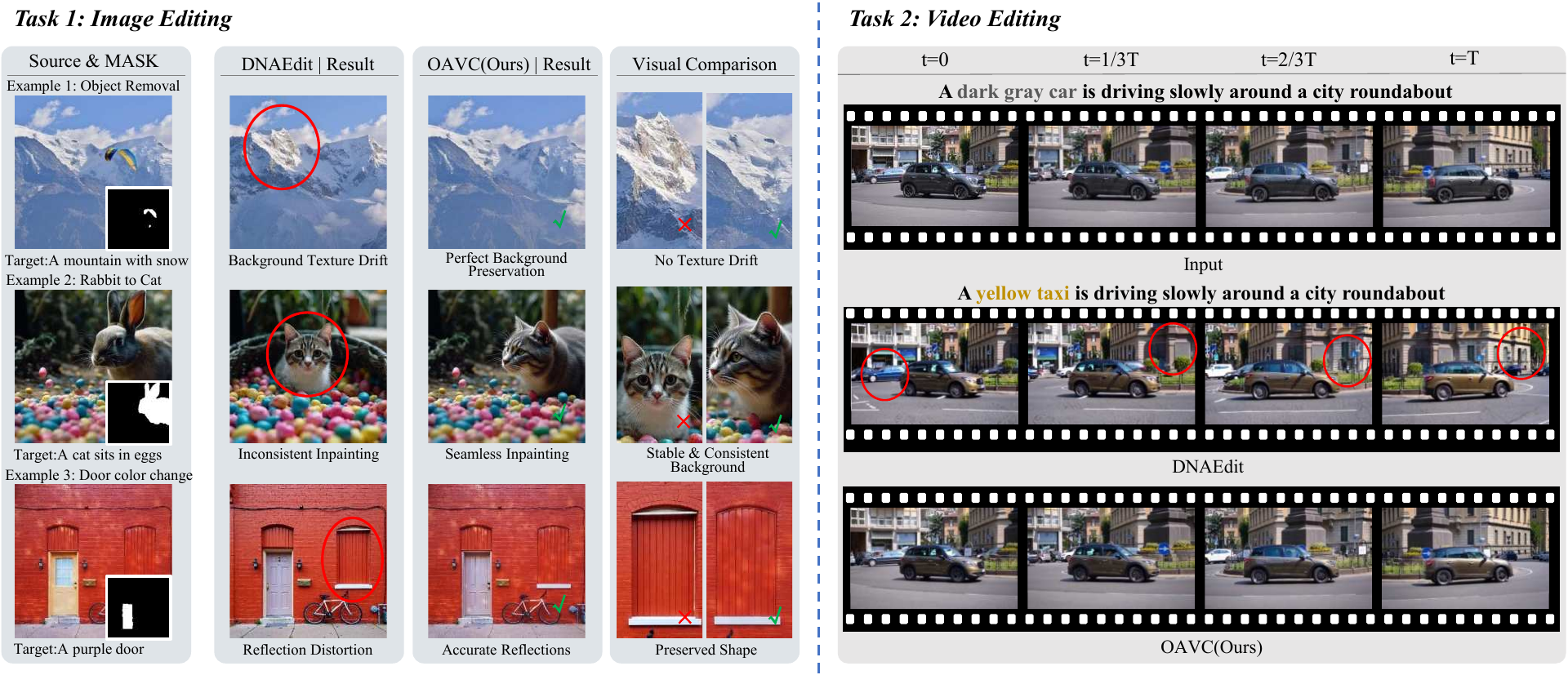}
  \caption{\textbf{Object-centric image and video editing.}
  Our approach enables localized edits while keeping non-target regions stable during multi-step integration.
  Prior training-free velocity-steering methods often accumulate small background residuals, leading to background drift, boundary artifacts, and temporal inconsistency.
  By enforcing object-localized semantic updates with explicit non-target preservation, we produce cleaner object removal/inpainting, faithful attribute or color changes, and temporally consistent video edits.}
  \label{fig:teaser}
\end{figure}

\section{Introduction}

Diffusion and flow-matching generative models enable high-fidelity synthesis with strong prompt control
\citep{ho2020denoising, song2021score, lipman2023flow}.
An increasingly important capability is training-free image editing: given an input image and a textual instruction,
the model should follow the instruction while preserving unrelated content at inference time, without finetuning or per-image optimization
\citep{kulikov2025flowedit, rout2025semantic, gong2025instantedit, wang2024taming}.
This avoids the cost and instability of training-based editing pipelines \citep{meng2022sdedit}.

A common strategy is to intervene in the sampling trajectory by comparing source- and target-conditioned denoising directions and injecting their difference as an editing signal
\citep{hertz2022prompt, Tumanyan_2023_CVPR, xie2025dnaedit, kulikov2025flowedit, wang2024taming, deng2024fireflowfastinversionrectified, kim2025flowalign}.
Existing training-free methods improve editing fidelity through different mechanisms, including trajectory or inversion correction, source-information regulation, delayed or region-aware injection, and attention/cache modulation
\citep{xie2025dnaedit, kulikov2025flowedit, jiao2025unieditflowunleashinginversionediting, ouyang2025proedit, zhu2025kv}.
However, existing methods often derive spatial cues from the model response itself, such as prompt-induced velocity differences, attention maps, or internal feature changes.
Because these cues remain coupled with the target-prompt residual, spatially imprecise responses may cause updates to leak into the background, miss parts of the object, or become unstable near boundaries.

\begin{figure}[t]
\centering
\includegraphics[width=\textwidth]{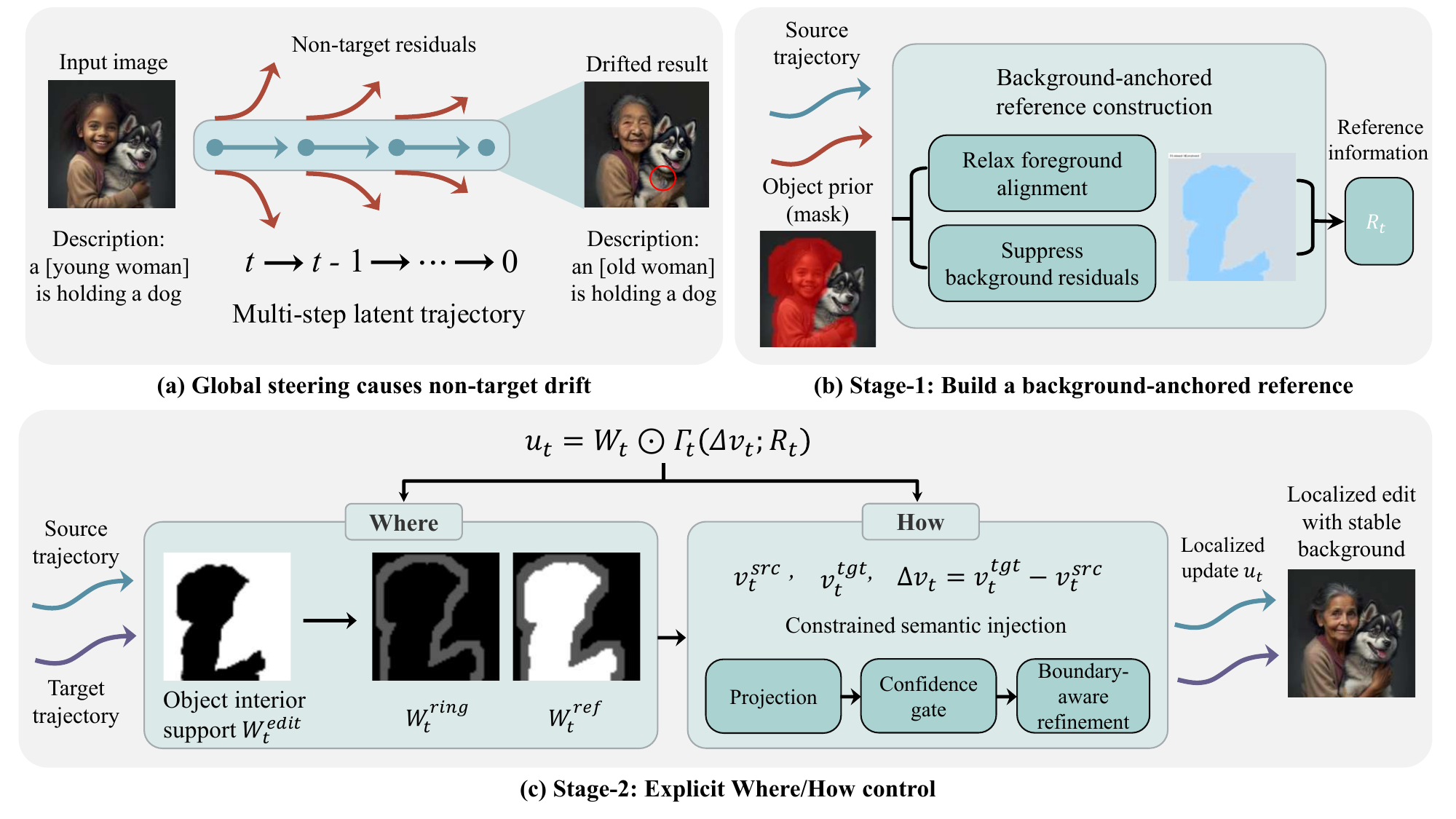} 
\caption{
\textbf{Overview of Object-Aware Velocity Control.}
Global steering applies prompt-induced residuals over the full latent trajectory, causing non-target accumulation and background drift.
OAVC first constructs a background-anchored reference interface, then applies explicit \emph{where/how} control: object supports specify where semantic residuals are integrated, while a safe injection operator specifies how they are transformed before solver integration.
}
\label{fig:pipeline}
\end{figure}

As shown in Fig.~\ref{fig:teaser}, this limitation becomes important for object-centric edits, such as object replacement, material change, deletion, or localized attribute modification.
Users expect the semantic change to occur on the target object while the surrounding regions remain stable.
However, prompt-induced velocity differences are rarely zero outside the intended object region.
When such residuals are integrated globally in a spatially coupled latent space, small non-target components can accumulate over solver steps, producing background drift, layout changes, and unstable object boundaries, as illustrated in Fig.~\ref{fig:pipeline}(a).
This suggests that object-centric editing should not rely only on implicit localization from prompt responses; it needs an object-level support that is specified independently of the prompt residual and applied during velocity integration.

In this work, we propose \textbf{Object-Aware Velocity Control} (OAVC), a training-free framework for object-centric editing with rectified-flow backbones.
Our key idea is to make the editable region a solver-level control variable and decouple it from the residual transformation itself.
Specifically, OAVC controls \emph{where} prompt-induced residuals may accumulate using an object support, and controls \emph{how} they are injected by suppressing source-flow-aligned components before integration.
It first constructs a background-anchored reference interface under the source prompt for stable source--target velocity comparison, and then injects target semantics through object-localized constrained velocity updates.
The object support is used only as a velocity-level prior during solver integration.

To instantiate the object support, we use SAM3~\citep{carion2025sam3segmentconcepts} as the default zero-shot support provider.
OAVC only requires an object-centric support prior, not a specific segmentation model, and can also use user-provided or brush-refined supports.
Localization itself is not our contribution; controlled same-support and support-robustness studies isolate the contribution of the proposed trajectory
controller and characterize its dependence on support quality.

We evaluate OAVC on object-centric image and video benchmarks under rectified-flow editing settings~\citep{rout2025semantic, xu2025unveil, deng2024fireflowfastinversionrectified}.
Unlike masked blending or inpainting, which enforce locality by replacing latent or image regions, OAVC constrains the prompt-induced residual field before solver integration.
Experiments on PIE-Bench and DAVIS show improved background preservation, structural fidelity, boundary stability, and temporal consistency while retaining effective localized editability.
We further show that the same \emph{where/how} velocity-control principle can transfer to another RF-based editor \citep{kulikov2025flowedit}, indicating that OAVC is not tied to a single base pipeline.

\begin{figure}[t]
    \centering
    \includegraphics[width=1.0\linewidth]{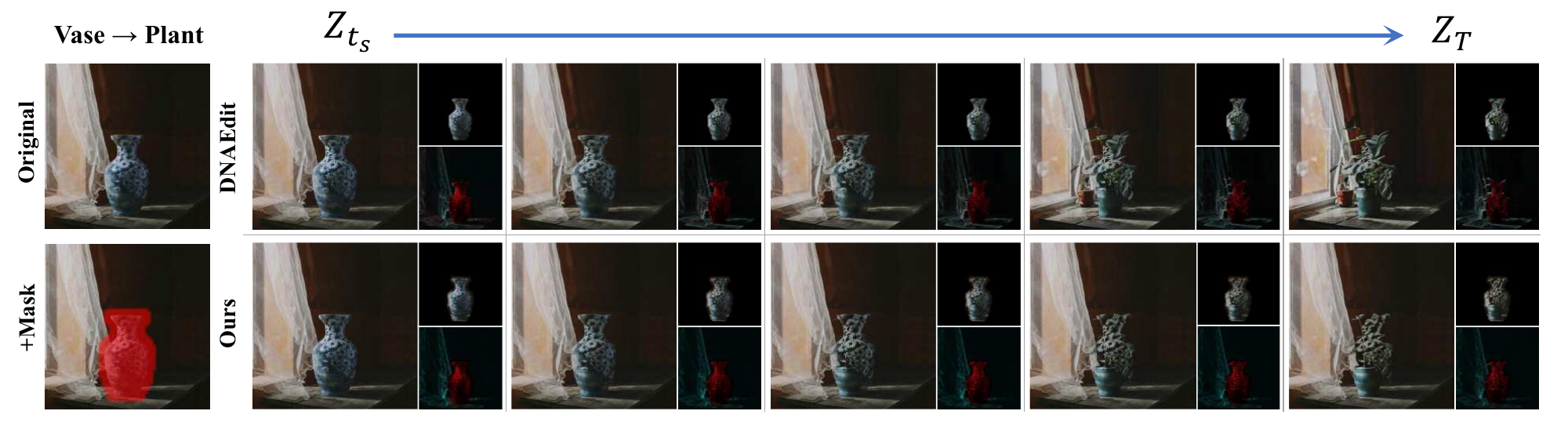}
    
    \caption{Illustration of Stage-2: Projected and Constrained Semantic Injection. We explicitly constrain the edit to the object region and a safe injection direction.}
    \label{fig:stage2_diagram}
\end{figure}
\section{Related Work}

\textbf{Training-free editing via trajectory and model control.}
Training-free image editing performs prompt-driven modifications at inference time with large pretrained
generative models, without finetuning or per-image optimization.
A classical branch trains instruction-following diffusion models for edits
\citep{brooks2022instructpix2pix, ju2024brushnet, li2025editthinkerunlockingiterativereasoning},
which enables efficient text-guided editing but relies on task-specific training.

Recent works instead achieve training-free control by intervening in a fixed pretrained model during sampling.
Attention- and token-level approaches steer semantics by modifying cross-attention or internal activations
\citep{hertz2022prompt, parmar2023zero, chefer2023attend}.
While flexible, they often depend on implicit localization and can be sensitive to prompts and schedules,
which makes stable object-level control challenging.

A complementary direction views editing as direct manipulation of generative dynamics in diffusion or
flow-matching models, leveraging prompt-conditioned velocity fields
\citep{lipman2023flow, wang2024taming, deng2024fireflowfastinversionrectified, kim2025flowalign}.
Velocity- and trajectory-based editors inject prompt-induced velocity differences along multi-step integration
\citep{xie2025dnaedit, kulikov2025flowedit, yoon2025splitflow} and have been demonstrated on modern backbones such as
SD3 \citep{esser2024scaling} and FLUX \citep{flux2024}.

Despite their algorithmic differences, existing velocity-based methods often apply semantic velocity perturbations
\emph{globally} in a spatially coupled latent space, so small non-target residuals can accumulate across steps,
causing background drift and boundary instability in object-centric edits.
Other training-free stabilization mechanisms, such as improved inversion consistency or attention/cache modulation
\citep{jiao2025unieditflowunleashinginversionediting, ouyang2025proedit, zhu2025kv},
improve robustness in some cases but do not explicitly constrain \emph{where} and \emph{how} semantic
velocity updates are spatially integrated.
Our work follows the velocity-control paradigm and addresses this limitation with object-aware spatial support and constrained injection.

\textbf{Spatial control and object-aware editing.}
Localized editing with pretrained generative models has long been studied.
Inference-time methods without explicit masks often rely on attention- or feature-level cues such as plug-and-play guidance
or key--value cache manipulation \citep{Tumanyan_2023_CVPR, ouyang2025proedit}. Other diffusion editors use soft attention masks or adaptive mask localization
to preserve source details and improve consistency in non-target regions
\citep{10.1145/3702999, 10.1145/3778175}.
Despite these mechanisms, existing methods do not directly constrain the
velocity field being integrated in rectified-flow editing.

Segmentation foundation models, including Segment Anything~\citep{kirillov2023segany} and concept-aware variants
\citep{carion2025sam3segmentconcepts}, enable high-quality zero-shot object masks
\citep{kirillov2023segany, carion2025sam3segmentconcepts}, offering explicit object-level spatial priors.
However, training-free editing methods rarely incorporate such priors into velocity manipulation or trajectory integration.
In contrast, we inject an object-level prior into the velocity-control mechanism itself, enabling spatially selective and
dynamically constrained integration while remaining fully training-free and compatible with velocity-based editing frameworks.
Unlike KV-Edit's attention-level background KV reuse~\citep{zhu2025kv}, OAVC constrains residuals at the editor/solver interface; same-support controls separate this contribution from access to a mask.

\section{Method}
\label{sec:method}

We propose Object-Aware Velocity Control, a training-free framework for object-centric editing with rectified-flow backbones.
As illustrated in Fig.~\ref{fig:pipeline}, OAVC addresses non-target residual accumulation by introducing object-aware \emph{where/how} control directly into the velocity-integration process.
It first constructs a background-anchored reference interface under the source prompt, and then injects target semantics through object-localized safe velocity updates.

\subsection{Preliminaries}
\label{sec:prelim}
\paragraph{Noise-space interpolation in rectified-flow editing.}
We consider a rectified-flow velocity field $v_\theta(Z,\psi)$ parameterized by a pretrained model,
where $Z$ is a latent state and $\psi$ is a text condition.
Let $\{\sigma_t\}_{t=0}^{T}$ be the solver noise levels and $Z_{t+1}$ be the current latent at level $\sigma_{t+1}$.
Following DNAEdit~\citep{xie2025dnaedit}, we construct a source-consistent trajectory under the source prompt $\psi_{\mathrm{src}}$ by evaluating the velocity on an interpolated proxy state.
Given an auxiliary noise state $S_{t+1}$, the proxy state at level $\sigma_t$ is:
\begin{equation}
Z_t^{*}
=
\frac{\sigma_t}{\sigma_{t+1}} Z_{t+1}
+
\Bigl(1-\frac{\sigma_t}{\sigma_{t+1}}\Bigr) S_{t+1}.
\label{eq:zt_star}
\end{equation}
The source velocity is $v_t^{\mathrm{src}}=v_\theta(Z_t^{*},\psi_{\mathrm{src}})$.
The analytic linear velocity is $v_t^{\mathrm{lin}}=(S_{t+1}-Z_{t+1})/\sigma_{t+1}$, yielding the DNA-style residual
$\Delta v_t^{\mathrm{DNA}}=v_t^{\mathrm{lin}}-v_t^{\mathrm{src}}$.
OAVC modifies this residual in Stage-1 to obtain a safe source trajectory and cached offset for aligned target evaluation.
We also inherit the reference-guided editing formulation of the base editor.
Given an edited latent $Z_t^{\mathrm{edit}}$ and a moving reference anchor $Z_{t+1}^{\mathrm{ref}}$, the reference velocity is
$v_t^{\mathrm{ref}}=(Z_t^{\mathrm{edit}}-Z_{t+1}^{\mathrm{ref}})/(1-\sigma_t)$.
The foreground editing velocity is
$v_t^{\mathrm{fg}}=\eta v_t^{\mathrm{tgt}}+(1-\eta)v_t^{\mathrm{ref}}$,
where $v_t^{\mathrm{tgt}}=v_\theta(Z_t^{*\mathrm{edit}},\psi_{\mathrm{tgt}})$ and $\eta\in[0,1]$ controls the editability--fidelity trade-off.
The aligned edited proxy $Z_t^{*\mathrm{edit}}$ is defined in Stage-2.

\paragraph{Why global velocity steering drifts.}
Training-free velocity-steering methods add a control signal $u_t$ to the model velocity.
For a discrete solver step, $Z_{t+1}=Z_t+\Delta_t\,\hat v_t$, where $\hat v_t=v_\theta(Z_t,\psi)+u_t$ and $\Delta_t$ is the scheduler step size.
Unrolling the trajectory shows that this control accumulates over time.
On a background region $\Omega_{\mathrm{bg}}$, the accumulated change is:
\begin{equation}
\Delta Z_{\mathrm{bg}}
\;\triangleq\;
\sum_{t=0}^{T-1}
\Delta_t\,u_t\big|_{\Omega_{\mathrm{bg}}}.
\label{eq:bg_drift_discrete}
\end{equation}
Thus, even weak non-target components can become visible if they are repeatedly integrated across many steps.
This is the key distinction from post-hoc masked blending or inpainting: those methods constrain the final latent or image, while OAVC constrains the residual field before solver integration.
A more detailed derivation is provided in Appendix~\ref{sec:app_full_derivations}.
\paragraph{Where/how factorization.}
We formulate object-centric editing as a controlled velocity update:
\begin{equation}
u_t
=
W_t \odot \Gamma_t(\Delta v_t;\mathcal R_t).
\label{eq:where_how_factorization}
\end{equation}
Here $W_t$ specifies \emph{where} prompt-induced residuals can accumulate, and $\Gamma_t(\cdot)$ specifies \emph{how} they are transformed before integration.
$\mathcal R_t$ is the Stage-1 reference interface, including the aligned source trajectory, cached offsets, and source velocities.
We use SAM3 by default, but OAVC is support-source agnostic and can use attention-derived, user-provided, or brush-refined supports. Details are in Appendices~\ref{sec:appendix_object_prior} and~\ref{sec:appendix_support_source}.

% ------------------------------------------------------------
\subsection{Stage-1: Background-Anchored Reference Interface}
\label{sec:stage1}

Stage-1 constructs the reference interface $\mathcal R_t$ used by the controlled injection stage.
It does not perform semantic editing.
Its purpose is to provide a source-consistent trajectory on which source and target velocities can be compared.
For object-centric editing, this reference should satisfy two asymmetric requirements: the background should stay tightly anchored to the source dynamics, while the object region should keep enough freedom for later semantic change.
This asymmetry is important because overly strong source alignment inside the object can suppress the desired edit, whereas unconstrained residuals on the background can accumulate into visible drift.

Starting from the DNA alignment residual $\Delta v_t^{\mathrm{DNA}}$, we first relax the source-alignment force inside the editable object:
\begin{equation}
\Delta v_t^{(1)}
=
(1-\lambda_{\mathrm{fg}}M_{\mathrm{fg}})
\odot
\Delta v_t^{\mathrm{DNA}}.
\label{eq:fg_relax}
\end{equation}
Here $M_{\mathrm{fg}}$ denotes the object support and $M_{\mathrm{bg}}=1-M_{\mathrm{fg}}$ denotes its complement.
The coefficient $\lambda_{\mathrm{fg}}\in[0,1]$ has a direct interpretation: it controls foreground editability.
A larger value leaves more degrees of freedom for later semantic injection, while a smaller value makes the reference more source-anchored.
Thus, $\lambda_{\mathrm{fg}}$ is not introduced as an independent engineering knob, but as the foreground side of the editability--preservation trade-off.

For the background, the residual should refine the source trajectory without introducing coherent structural motion.
We use the same flow-orthogonal projection throughout OAVC, denoted by 
$\Pi_\perp(a;b)=a-\frac{\langle a,b\rangle}{\langle b,b\rangle+\epsilon}b$, 
which removes from $a$ the component parallel to $b$.
Let $r_t=M_{\mathrm{bg}}\odot\Delta v_t^{(1)}$ be the background residual and $b_t=M_{\mathrm{bg}}\odot v_t^{\mathrm{src}}$ be the background source flow.
We then apply a soft projection:
$\tilde r_t=(1-\rho_{\mathrm{bg}})r_t+\rho_{\mathrm{bg}}\Pi_\perp(r_t;b_t)$,
where $\rho_{\mathrm{bg}}\in[0,1]$ controls the strength of background residual suppression.
Thus, $\rho_{\mathrm{bg}}=0$ keeps the original DNA-style background residual, while $\rho_{\mathrm{bg}}=1$ fully removes its source-flow-parallel component.

The final Stage-1 residual is:
\begin{equation}
\Delta v_t^{\mathrm{safe}}
=
M_{\mathrm{fg}}\odot \Delta v_t^{(1)}
+
\tilde r_t .
\label{eq:stage1_safe}
\end{equation}
This residual relaxes the editable object while preserving a background-anchored source trajectory.
Using this residual, we follow the same proxy-state refinement rule as DNAEdit:
\begin{equation}
S_t
=
S_{t+1}
+
\sigma_{t+1}\Delta v_t^{\mathrm{safe}},
\qquad
Z_t
=
Z_t^{*}
+
(\sigma_{t+1}-\sigma_t)\Delta v_t^{\mathrm{safe}}.
\label{eq:stage1_update}
\end{equation}
We then cache the offset:
\begin{equation}
\Delta x_t^{\mathrm{safe}}
=
Z_t-Z_t^{*},
\label{eq:stage1_offset}
\end{equation}
which maps the proxy state to the refined reference state.
Stage-1 therefore outputs $\mathcal R_t=\{Z_t,\Delta x_t^{\mathrm{safe}},v_t^{\mathrm{src}}\}$ for Stage-2.

The role of Stage-1 is important for distinguishing OAVC from simple masked baselines.
A final masked blend can restore the source background after generation, but it cannot prevent non-target velocity residuals from being integrated during sampling.
In contrast, Stage-1 anchors the source trajectory before target semantic injection begins, so preservation is imposed at the trajectory level rather than repaired only at the final latent or image.

\begin{figure}[t]
\centering
\vspace{-2mm}
\setlength{\tabcolsep}{0pt}

{\scriptsize
\begin{tabularx}{\linewidth}{@{} V *{10}{K} @{}}
\noalign{\vskip 2mm} 
& \textbf{Source} & \textbf{DNAEdit-SD3} & \textbf{Ours-SD3} & \textbf{DNAEdit-FLUX} & \textbf{Ours-FLUX} & \textbf{FlowEdit} & \textbf{UniEdit} & \textbf{FireFlow} & \textbf{InfEdit} & \textbf{RF-Solver} \\[-0.5mm]

\raisebox{-5pt}{\vlabel{Rocks:\\Goat$\rightarrow$Horse}} & 
\imgcell{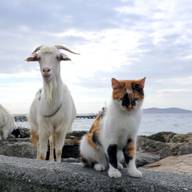}&
\imgcell{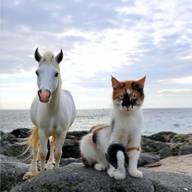}&
\imgnode{s_top}{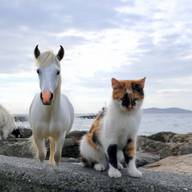}&
\imgcell{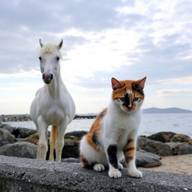}&
\imgnode{f_top}{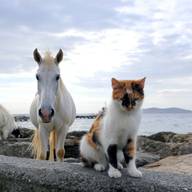}&
\imgcell{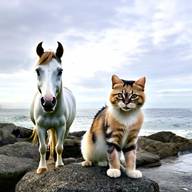}&
\imgcell{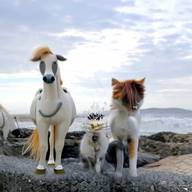}&
\imgcell{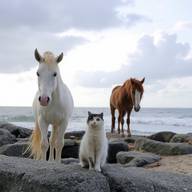}&
\imgcell{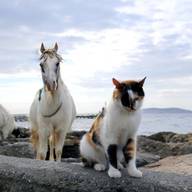}&
\imgcell{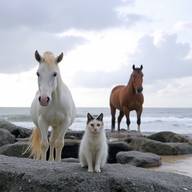} \\ \noalign{\vskip 0.5mm}

% ---------------- Row 2 ----------------
\raisebox{-0.1pt}{\vlabel{Statue:\\Front$\rightarrow$Side}} & 
\imgcell{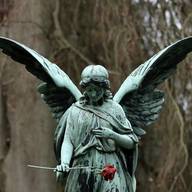}&
\imgcell{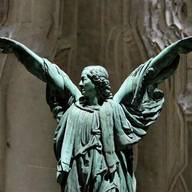}&
\imgcell{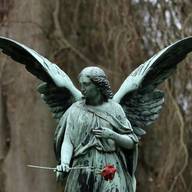}&
\imgcell{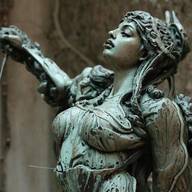}&
\imgcell{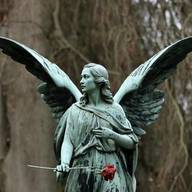}&
\imgcell{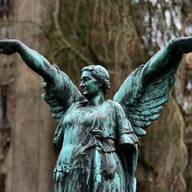}&
\imgcell{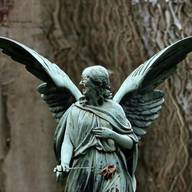}&
\imgcell{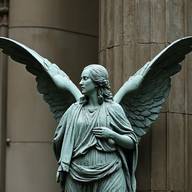}&
\imgcell{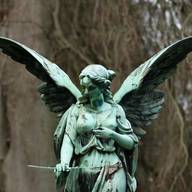}&
\imgcell{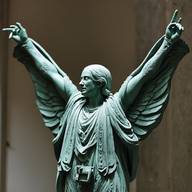} \\ \noalign{\vskip 0.5mm}

% ---------------- Row 3 ----------------
\raisebox{-10pt}{\vlabel{Countryside:\\-Church}} & 
\imgcell{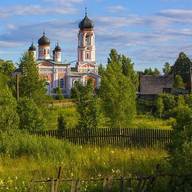}&
\imgcell{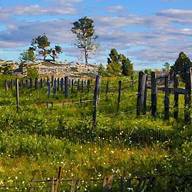}&
\imgcell{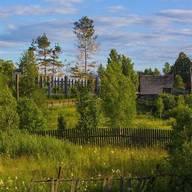}&
\imgcell{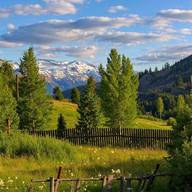}&
\imgcell{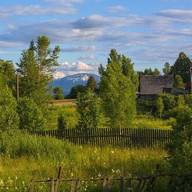}&
\imgcell{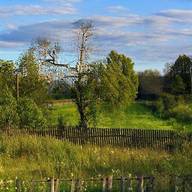}&
\imgcell{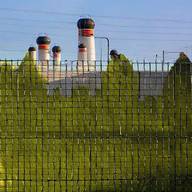}&
\imgcell{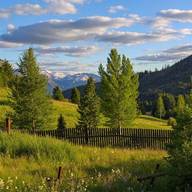}&
\imgcell{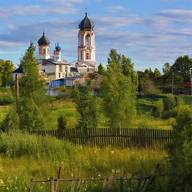}&
\imgcell{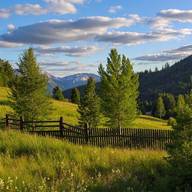} \\ \noalign{\vskip 0.5mm}

% ---------------- Row 4 ----------------
\raisebox{-10pt}{\vlabel{Table:+Books}} & 
\imgcell{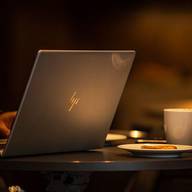}&
\imgcell{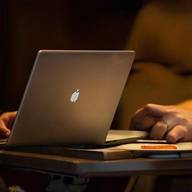}&
\imgcell{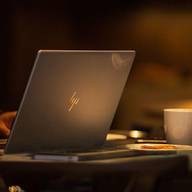}&
\imgcell{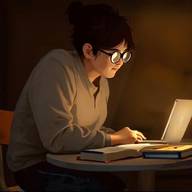}&
\imgcell{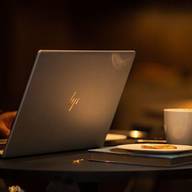}&
\imgcell{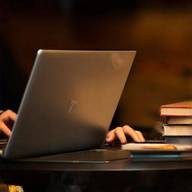}&
\imgcell{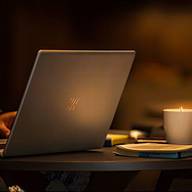}&
\imgcell{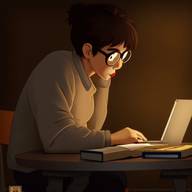}&
\imgcell{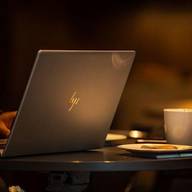}&
\imgcell{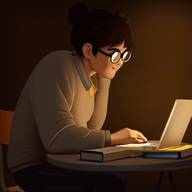} \\ \noalign{\vskip 0.5mm}

% ---------------- Row 5 ----------------
\raisebox{-10pt}{\vlabel{Angel:\\Field$\rightarrow$River}} & 
\imgcell{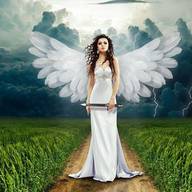}&
\imgcell{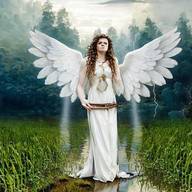}&
\imgcell{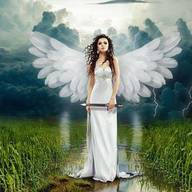}&
\imgcell{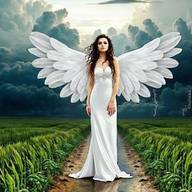}&
\imgcell{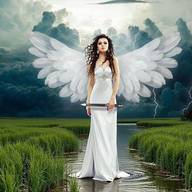}&
\imgcell{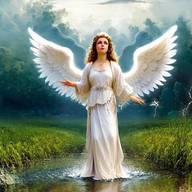}&
\imgcell{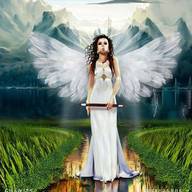}&
\imgcell{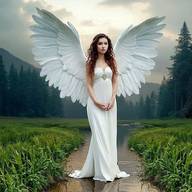}&
\imgcell{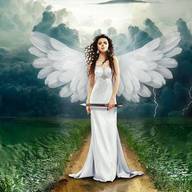}&
\imgcell{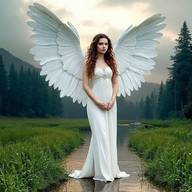} \\ \noalign{\vskip 0.5mm}

% ---------------- Row 6 ----------------
\raisebox{-10pt}{\vlabel{Makeup:\\\textcolor{yellow!50!brown}{Gold}$\rightarrow$\textcolor{blue}{Blue}}}& 
\imgcell{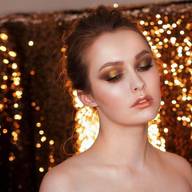}&
\imgcell{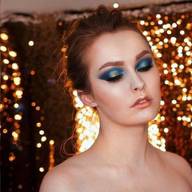}&
\imgcell{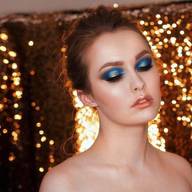}&
\imgcell{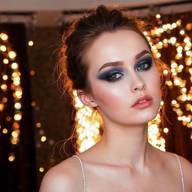}&
\imgcell{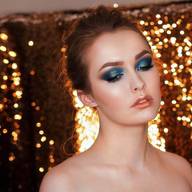}&
\imgcell{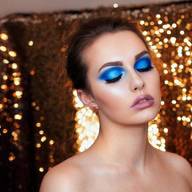}&
\imgcell{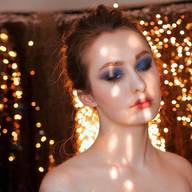}&
\imgcell{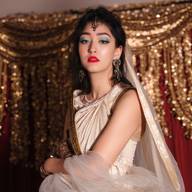}&
\imgcell{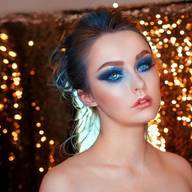}&
\imgcell{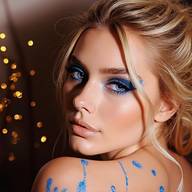} \\ \noalign{\vskip 0.5mm}

\raisebox{-1pt}{\vlabel{Woman's hair:\\Curly$\rightarrow$Straight}} & 
\imgcell{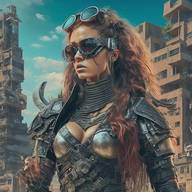}&
\imgcell{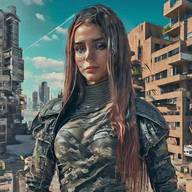}&
\imgnode{s_bot}{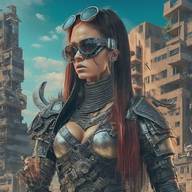}&
\imgcell{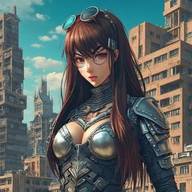}&
\imgnode{f_bot}{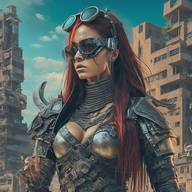}&
\imgcell{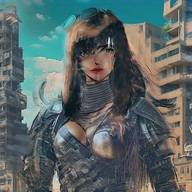}&
\imgcell{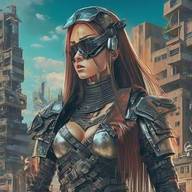}&
\imgcell{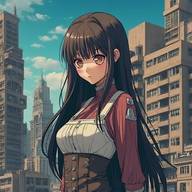}&
\imgcell{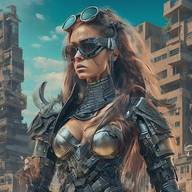}&
\imgcell{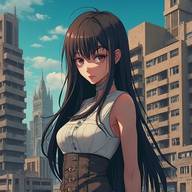}    \\ \relax

\end{tabularx}
}

\drawcolbox{s_top}{s_bot}
\drawcolbox{f_top}{f_bot}

\vspace{-8mm} 

\caption{\textbf{Qualitative comparison on PIE-Bench.}
We compare OAVC against representative training-free editors across SD3.5 and FLUX.
OAVC better preserves non-target background and object boundaries while achieving the intended edits (red boxes).}
\label{fig:qualitative_results}
\end{figure}
% ------------------------------------------------------------

\subsection{Stage-2: Object-Localized Safe Semantic Injection}
\label{sec:stage2}

Stage-2 applies the controlled update in Eq.~\eqref{eq:where_how_factorization}, as illustrated in Fig.~\ref{fig:stage2_diagram}.
It computes target semantic residuals on offset-aligned states, transforms them with the same safe injection principle introduced above, and integrates them only on the object support.
We initialize the edited trajectory from Stage-1 at a chosen starting step $t_s$ by setting $Z_{t_s}^{\mathrm{edit}}\leftarrow Z_{t_s}$.

To compare source and target velocities on consistent states, we reuse the cached Stage-1 offset and form the aligned edited proxy state $Z_t^{*\mathrm{edit}}=Z_t^{\mathrm{edit}}+\Delta x_t^{\mathrm{safe}}$.
The target semantic residual is then computed as $\Delta v_t=v_\theta(Z_t^{*\mathrm{edit}},\psi_{\mathrm{tgt}})-v_t^{\mathrm{src}}$.
This residual captures the target semantic direction, but it may still contain components that follow the source flow and induce structural motion.
Therefore, the safe injection operators $\Gamma_t^{\mathrm{ref}}$ and $\Gamma_t^{\mathrm{inj}}$ use the projection operator $\Pi_\perp(\cdot;\cdot)$ as their main drift-suppression mechanism.

In addition to flow-orthogonal suppression, OAVC uses fixed boundary smoothing and confidence normalization inside the safe injection operators.
These terms are not treated as separate conceptual modules or per-image tuned components.
They only stabilize the transition near uncertain object boundaries and avoid unstable over-injection when the semantic residual is poorly localized.
Their definitions and fixed settings are provided in Appendices~\ref{sec:appendix_gamma} and~\ref{sec:appendix_gate}.

We use $\Gamma_t^{\mathrm{ref}}(\Delta v_t;\mathcal R_t)$ to denote the safe semantic residual for updating the moving reference anchor, and $\Gamma_t^{\mathrm{inj}}(\cdot;\mathcal R_t)$ to denote the final safe residual injected into the edited velocity.
Both operators instantiate the same where/how principle in Eq.~\eqref{eq:where_how_factorization}; implementation details are provided in Appendix~\ref{sec:appendix_gamma}. OAVC also maintains a moving reference anchor $Z_t^{\mathrm{ref}}$.
This anchor accumulates only the safe semantic displacement within the object-centered reference support:
\begin{equation}
Z^{\mathrm{ref}}_{t+1}
=
Z^{\mathrm{ref}}_t
+
(\sigma_{t+1}-\sigma_t)
\left(
W_t^{\mathrm{ref}}\odot
\Gamma_t^{\mathrm{ref}}(\Delta v_t;\mathcal R_t)
\right).
\label{eq:ref_update_stage2}
\end{equation}
The anchor provides a stable image-space reference for the reference-guided velocity formulation introduced in Sec.~\ref{sec:prelim}, while preventing target semantics from accumulating freely on the background.

Because the reference-guided foreground velocity $v_t^{\mathrm{fg}}$ is the velocity signal used for editing, we apply the same safety principle after reference-guided blending.
The final edited velocity is:
\begin{equation}
v_t^{\mathrm{edit}}
=
v_t^{\mathrm{src}}
+
W_t^{\mathrm{edit}}
\odot
\Gamma_t^{\mathrm{inj}}
\left(
v_t^{\mathrm{fg}}-v_t^{\mathrm{src}};
\mathcal R_t
\right).
\label{eq:final_edit_velocity}
\end{equation}
Thus, on the background where $W_t^{\mathrm{edit}}=0$, the integrated velocity reduces to the source velocity, while the object region receives a constrained semantic update.
The edited trajectory is updated by:
\begin{equation}
Z_{t+1}^{\mathrm{edit}}
=
Z_t^{\mathrm{edit}}
+
(\sigma_{t+1}-\sigma_t)
v_t^{\mathrm{edit}} .
\label{eq:edit_integrate}
\end{equation}

This design differs from latent-space masked blending.
Masked blending enforces locality by replacing latent or image regions after they have already been generated or partially denoised.
OAVC instead constrains the velocity field being integrated, so non-target semantic residuals are prevented from accumulating on the background throughout the trajectory.

\begin{table}[!t]
\caption{Quantitative comparison on PIE-Bench. OAVC improves structural fidelity and background preservation while retaining effective localized editability. The best and second-best results are indicated by \textbf{bold} and \underline{underlined} text, respectively.}
\label{tab:pie_full}
\centering
\small
\setlength{\tabcolsep}{3.5pt}
\resizebox{\linewidth}{!}{%
\begin{tabular}{l|c|c|c|c|c|c|c|c}
\toprule
\multirow{2}{*}{Method} &
\multirow{2}{*}{\textbf{Model}} &
\multicolumn{1}{c|}{\textbf{Structure}} &
\multicolumn{4}{c|}{\textbf{Background Preservation}} &
\multicolumn{2}{c}{\textbf{CLIP Similarity}} \\
\cmidrule(lr){3-3} \cmidrule(lr){4-7} \cmidrule(lr){8-9}
& & Distance$\downarrow$ &
PSNR$\uparrow$ & LPIPS$\downarrow$ & MSE$\downarrow$ & SSIM$\uparrow$ &
Whole$\uparrow$ & Edited$\uparrow$ \\
\midrule
PnP~\citep{Tumanyan_2023_CVPR} & SD1.5 & 27.35 & 22.31 & 112.76 & 82.95 & 79.25 & 25.41 & 22.52 \\
DI+PnP~\citep{ju2023directinversion} & SD1.5 & 23.35 & 22.46 & 105.51 & 79.94 & 79.88 & 25.49 & 22.64 \\
InfEdit~\citep{xu2023infedit} & LCM & 19.31 & 27.31 & 56.32 & 47.80 & 85.30 & 24.90 & 22.14 \\
\midrule
RF-Inv~\citep{rout2025semantic} & FLUX & 42.29 & 20.20 & 179.54 & 139.20 & 69.91 & 24.57 & 22.20 \\
RFEdit~\citep{wang2024taming} & FLUX & 21.79 & 24.83 & 113.15 & 52.46 & 83.38 & 25.57 & 22.26 \\
FireFlow~\citep{deng2024fireflowfastinversionrectified} & FLUX & 29.03 & 23.33 & 133.40 & 70.83 & 81.22 & 26.19 & \underline{22.99} \\
FlowEdit~\citep{kulikov2025flowedit} & FLUX & 27.82 & 21.96 & 112.19 & 94.99 & 83.08 & 25.25 & 22.58 \\
FlowEdit~\citep{kulikov2025flowedit} & SD3 & 27.12 & 22.22 & 104.12 & 85.96 & 93.22 & \textbf{26.53} & \textbf{23.57} \\
FTEdit~\citep{xu2025unveil} & SD3.5 & 18.17 & 26.62 & 80.55 & 40.24 & 91.50 & 25.74 & 22.27 \\
FlowAlign~\citep{kim2025flowalign} & SD3 & 28.00 & 25.50 & 53.00 & 40.00 & 87.90 & 25.28 & 22.00 \\
SplitFlow~\citep{yoon2025splitflow} & SD3.5 & 11.68 & 27.12 & 52.93 & 30.61 & 89.76 & \underline{26.29} & 22.89 \\
UniEdit~\citep{jiao2025unieditflowunleashinginversionediting} & FLUX & 10.14 & 29.54 & 63.55 & 24.72 & 90.42 & 25.80 & 22.33 \\
\midrule
DNAEdit~\citep{xie2025dnaedit} & FLUX & 18.87 & 24.99 & 95.06 & 50.45 & 85.71 & 25.79 & 22.87 \\
DNAEdit~\citep{xie2025dnaedit} & SD3.5 & 14.19 & 26.66 & 74.57 & 32.76 & 88.63 & 25.63 & 22.71 \\
\rowcolor[HTML]{EAF3FF}
\textbf{OAVC (Ours)} & FLUX & \textbf{4.07} & \textbf{33.30} & \textbf{23.70} & \textbf{8.75} & \textbf{94.38} & 23.70 & 21.08 \\
\rowcolor[HTML]{DCEEFF}
\textbf{OAVC (Ours)} & SD3.5 & \underline{4.11} & \underline{32.64} & \underline{24.40} & \underline{9.61} & \underline{93.00} & 24.21 & 21.51 \\
\bottomrule
\end{tabular}%
}
\end{table}

\section{Experiments}
\label{sec:experiments}

% ------------------------------------------------------------
\subsection{Setup}
\label{sec:exp_setup}

\textbf{Datasets:}
\textbf{PIE-Bench}~\citep{ju2023directinversion} is our primary benchmark and consists of object-centric edit tuples
$(I,\psi_{\mathrm{src}},\psi_{\mathrm{tgt}})$ that cover replacement, attribute edits, and material changes.
We additionally evaluate on the \textbf{DAVIS} dataset~\citep{ponttuset20182017davischallengevideo} to assess temporal consistency under object-aware control.
Since DAVIS does not provide text-based edit instructions, we perform mask-conditioned video editing,
where the same object-level manipulation (e.g., deletion or localized replacement) is applied consistently across frames. For DAVIS sequences, we use the provided ground-truth instance masks and cache the object support for each frame,
reusing it across solver steps to decouple temporal dynamics from segmentation noise
and focus on the stability of the sampling trajectory.

\textbf{Backbones and compared methods.}
We primarily evaluate on two rectified-flow backbones, \textbf{SD3.5} and \textbf{FLUX}.
Our primary baseline is \textbf{DNAEdit}~\citep{xie2025dnaedit}, and we follow its default sampling configuration on both backbones for a fair comparison.
We compare against representative training-free text-driven editors from both diffusion and rectified-flow families.
For diffusion-based methods, we include \textbf{PnP}~\citep{Tumanyan_2023_CVPR}, \textbf{DI+PnP}~\citep{ju2023directinversion}, and \textbf{InfEdit}~\citep{xu2023infedit}.
For rectified-flow based methods, we include \textbf{RF-Inv}~\citep{rout2025semantic}, \textbf{RFEdit}~\citep{wang2024taming}, \textbf{FireFlow}~\citep{deng2024fireflowfastinversionrectified}, \textbf{FlowEdit}~\citep{kulikov2025flowedit}, and \textbf{UniEdit-Flow}~\citep{jiao2025unieditflowunleashinginversionediting}.
We report all results in Tab.~\ref{tab:pie_full}.

\textbf{Implementation details:}
We use the same MVG coefficient $\eta=0.8$ and the same sampling schedule as DNAEdit for each backbone to ensure a fair comparison: SD3.5 uses 40 solver steps with editing start step $t_s=13$ and target guidance scale 3.5, and FLUX uses 28 steps with $t_s=4$ and guidance scale 2.5.
In Stage-1 we set the foreground relaxation coefficient to $\lambda_{\mathrm{fg}}=0.6$ and the background projection mixing coefficient to $\rho_{\mathrm{bg}}=0.1$.
The object support uses $r_{\mathrm{fg}}=2$ and $r_{\mathrm{bd}}=4$ with boundary coupling $\beta_{\max}=0.45$ and $\gamma=2$ (Appendix~\ref{sec:appendix_object_prior}).
For DAVIS, we use the rectified-flow text-to-video backbone Wan2.1-T2V-1.3B on the first 41 frames at $832\times480$ resolution, with 50 solver steps, $t_s=12$, target guidance scale 3.5, $\lambda_{\mathrm{fg}}=0.1$, and $r_{\mathrm{fg}}=1$; all other OAVC settings are unchanged.
Additional implementation details and hyperparameter studies are provided in Appendices~\ref{sec:appendix_gamma}, \ref{sec:appendix_object_prior}, and~\ref{sec:ablation}.

\begin{table*}[t]
\caption{PIE-Bench ablation (all metrics scaled to match Tab.~\ref{tab:pie_full}). We report the SD3.5 setting. The best results are indicated by \textbf{bold} text.}
\label{tab:ablation_oavc}
\centering
\small
\setlength{\tabcolsep}{4pt}
\resizebox{\linewidth}{!}{
\begin{tabular}{l|c|c|c|c|c|c|c}
\toprule
\multirow{2}{*}{Setting} &
\multicolumn{1}{c|}{\textbf{Structure}} &
\multicolumn{4}{c|}{\textbf{Background Preservation}} &
\multicolumn{2}{c}{\textbf{CLIP Similarity}} \\
\cmidrule(lr){2-2} \cmidrule(lr){3-6} \cmidrule(lr){7-8}
& Distance$\downarrow$ &
PSNR$\uparrow$ & LPIPS$\downarrow$ & MSE$\downarrow$ & SSIM$\uparrow$ &
CLIP$_\text{whole}\uparrow$ & CLIP$_\text{edit}\uparrow$ \\
\midrule
DNAEdit (baseline; global) & 14.19 & 26.66 & 74.57 & 32.76 & 88.63 & 25.63 & 22.71 \\
Stage-1 only (Anchor-DNA) & 14.68 & 26.28 & 77.75 & 35.44 & 88.42 & \textbf{25.76} & \textbf{22.87} \\
\midrule
Stage-2 (Mask+Ring+Proj; w/o Gate) & 6.60 & 31.74 & 27.65 & 11.95 & 92.61 & 24.60 & 21.97 \\
Stage-2 (+Mask only; $W^{\mathrm{edit}}$) & 5.27 & 32.27 & 25.78 & 10.48 & 92.99 & 24.55 & 21.87 \\
DNAEdit + final latent hard blend & 6.28 & -- & 24.29 & 9.56 & -- & 24.46 & 21.98 \\
Stage-2 (+Mask+Ring; $W^{\mathrm{edit}}{+}W^{\mathrm{ref}}$) & 5.21 & 32.28 & 25.73 & 10.44 & 92.99 & 24.54 & 21.87 \\
Stage-2 (Mask+Ring+Gate; w/o Proj) & 6.06 & 32.15 & 26.82 & 10.70 & 92.70 & 24.38 & 21.72 \\
\midrule
\textbf{Full OAVC (Stage-1 + Stage-2)} & \textbf{4.11} & \textbf{32.64} & \textbf{24.40} & \textbf{9.61} & \textbf{93.00} & 24.21 & 21.51 \\
\bottomrule
\end{tabular}
}

\end{table*}

\textbf{Metrics:}
On PIE-Bench we report Struct.\ Dist.$\downarrow$ and background preservation metrics PSNR$\uparrow$ SSIM$\uparrow$
LPIPS$\downarrow$ and MSE$\downarrow$ computed on non-target regions defined by the object mask.
Generation and evaluation masks are distinct: SAM3 supplies only OAVC's generation support, while every method is scored with the same PIE-Bench annotation (556 valid backgrounds; Structure Distance/CLIP use all 700 examples).
Following the PIE-Bench and DNAEdit protocol, we apply scaling factors for readability:
Struct.\ Dist.\ $\times 10^{3}$, LPIPS $\times 10^{3}$, MSE $\times 10^{4}$, and SSIM $\times 10^{2}$.
PSNR is reported in dB (unscaled).
We additionally report CLIP$_{\mathrm{whole}}$ and CLIP$_{\mathrm{edit}}$ for semantic alignment on the full image and within the edited region.
For DAVIS, BG-PSNR/BG-L1 measure per-frame background fidelity between each edited frame and its source frame on non-target regions; PSNR is reported higher-is-better, while L1 is the mean absolute pixel error and lower-is-better.
TW-BG-PSNR/TW-BG-L1 measure temporal background stability by warping the previous edited frame to the current frame using source-video optical flow and computing the same errors on the valid background intersection.

% ------------------------------------------------------------

\subsection{Main Results}
\label{sec:exp_main}
\textbf{Quantitative results.}
Tab.~\ref{tab:pie_full} evaluates editing performance from structure preservation, background preservation, and CLIP similarity.
OAVC achieves the strongest preservation-oriented performance on PIE-Bench across both FLUX and SD3.5.
On FLUX, OAVC reduces the structure distance of DNAEdit from 18.87 to 4.07 and improves background PSNR from 24.99 dB to 33.30 dB.
It also outperforms UniEdit, which is already a strong flow-based baseline, indicating that explicit object-level velocity control better suppresses non-target drift than implicit trajectory localization.
On SD3.5, OAVC shows the same trend, reducing the structure distance from 14.19 to 4.11 and increasing PSNR from 26.66 dB to 32.64 dB over DNAEdit.
Although FlowEdit-SD3 obtains higher CLIP scores, its weaker structure and background metrics suggest that global velocity steering can spread target semantics beyond the intended object region.
In contrast, OAVC favors object-centric preservation and localized editability, leading to stronger structural and background fidelity.
We further analyze this CLIP preservation--editability trade-off in Sec.~\ref{sec:appendix_clip_tradeoff}.

We also evaluate video stability on DAVIS in Tab.~\ref{tab:davis_video_quant} using four sequences, identical source clips and prompts, the first 41 frames, and the same resolution and evaluation protocol.
Relative to the temporal editor UniEdit-Flow, OAVC improves BG-PSNR by 4.66 dB, reduces BG-L1 by 39.2\%, improves TW-BG-PSNR by 1.68 dB, and reduces TW-BG-L1 by 6.6\%.
These results show that OAVC suppresses both static background drift and frame-to-frame jitter.

\begin{wraptable}{r}{0.48\linewidth}
\vspace{-3mm}
\centering
\scriptsize
\setlength{\tabcolsep}{4pt}
\caption{
DAVIS quantitative video results.
BG metrics measure per-frame background fidelity, while TW-BG metrics measure temporal background stability using source-video optical flow.
}
\label{tab:davis_video_quant}
\vspace{-2mm}
\resizebox{\linewidth}{!}{
\begin{tabular}{l|c|c|c|c}
\toprule
Method &
BG-PSNR$\uparrow$ &
BG-L1$\downarrow$ &
TW-BG-PSNR$\uparrow$ &
TW-BG-L1$\downarrow$ \\
\midrule
DNAEdit & 19.85 & 0.0692 & 26.56 & 0.0281 \\
UniEdit-Flow & 22.87 & 0.0467 & 27.97 & 0.0227 \\
OAVC    & \textbf{27.53} & \textbf{0.0284} & \textbf{29.65} & \textbf{0.0212} \\
\bottomrule
\end{tabular}
}
\vspace{-4mm}
\end{wraptable}
\textbf{Qualitative results.}
Fig.~\ref{fig:qualitative_results} confirms the quantitative trends on diverse image edits, including object replacement, viewpoint change, object deletion/addition, background-region change, and fine-grained attribute editing.
Global-steering baselines such as DNAEdit, FlowEdit, UniEdit, and FireFlow often exhibit background drift, boundary artifacts, or off-target semantic leakage.
OAVC better localizes the edit to the target object while preserving non-target regions and cleaner boundaries.

\begin{figure}[t]
    \centering
    \includegraphics[width=\textwidth]{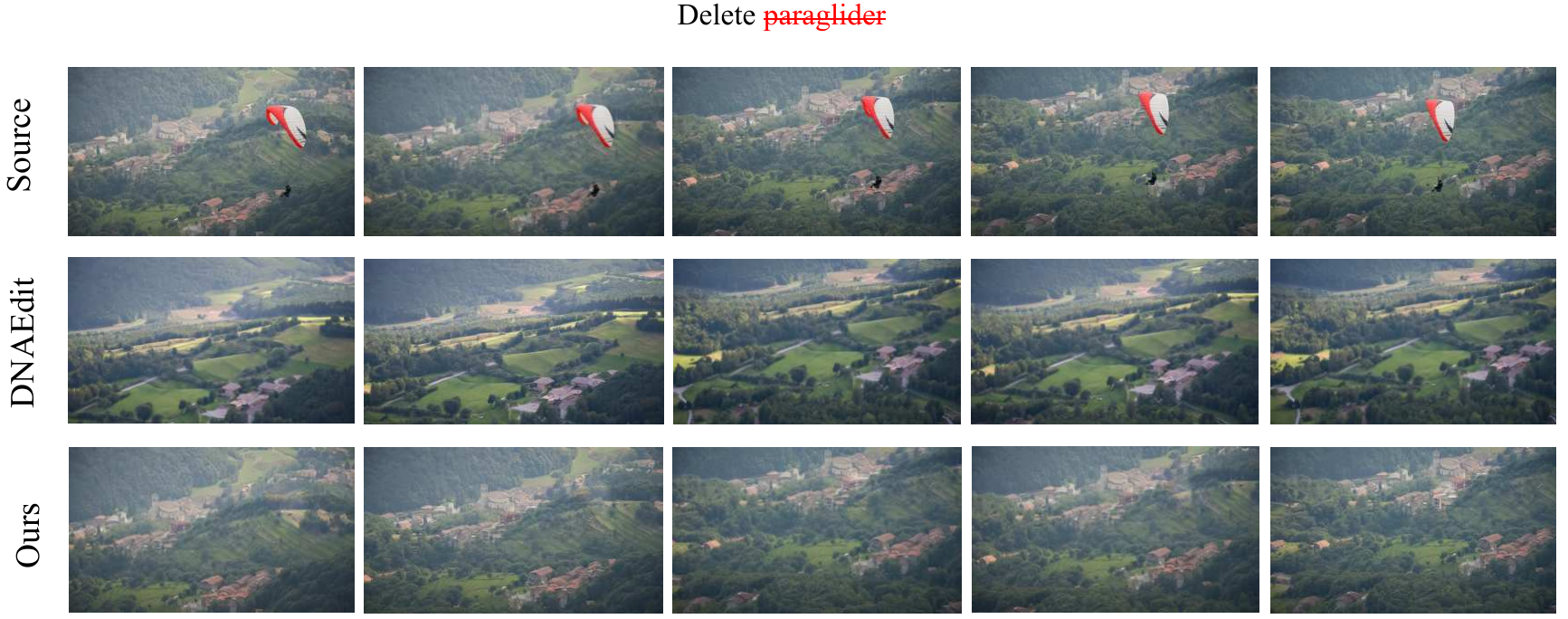}
    \caption{DAVIS video editing example for target-object removal across consecutive frames.}
    \label{fig:davis_delete_paraglider}
\end{figure}

For video editing, Fig.~\ref{fig:teaser} and Fig.~\ref{fig:davis_delete_paraglider} show consecutive-frame results on DAVIS.
Since each frame is edited independently, global steering can amplify small frame-wise residual differences and produce temporal inconsistency.
OAVC constrains prompt-induced residuals within the object support during integration, leading to more stable object removal and better preservation of the surrounding scene, consistent with the DAVIS metrics in Tab.~\ref{tab:davis_video_quant}.

% ------------------------------------------------------------

\begin{table}[t]
\caption{Transfer to FlowEdit backbones on PIE-Bench (all metrics scaled to match Tab.~\ref{tab:pie_full}).}
\label{tab:flowedit_transfer}
\centering
\small
\setlength{\tabcolsep}{4pt}
\resizebox{\linewidth}{!}{
\begin{tabular}{l|c|c|c|c|c|c|c|c}
\toprule
\multirow{2}{*}{Method} &
\multirow{2}{*}{\textbf{Backbone}} &
\multicolumn{1}{c|}{\textbf{Structure}} &
\multicolumn{4}{c|}{\textbf{Background Preservation}} &
\multicolumn{2}{c}{\textbf{CLIP Similarity$\uparrow$}} \\
\cmidrule(lr){3-3} \cmidrule(lr){4-7} \cmidrule(lr){8-9}
& & Distance$\downarrow$ &
PSNR$\uparrow$ & LPIPS$\downarrow$ & MSE$\downarrow$ & SSIM$\uparrow$ &
CLIP$_\text{whole}\uparrow$ & CLIP$_\text{edit}\uparrow$ \\
\midrule
FlowEdit & FLUX & 27.82 & 21.96 & 112.19 & 94.99 & 83.08 & 25.25 & 22.58 \\
FlowEdit & SD3 & 27.12 & 22.22 & 104.12 & 85.96 & 93.22 & \textbf{26.53} & 23.57 \\
\rowcolor[HTML]{DCEEFF}
Ours (OAVC on FlowEdit) & SD3 & 35.72 & 27.65 & 81.05 & 21.57 & 86.98 & 24.46 & 22.20 \\
\rowcolor[HTML]{EAF3FF}
Ours (OAVC on FlowEdit) & FLUX & \textbf{9.32} & \textbf{30.77} & \textbf{30.09} & \textbf{16.80} & \textbf{93.57} & 24.18 & \textbf{23.66} \\
\bottomrule
\end{tabular}
}
\end{table}
\subsection{Ablation Study}
\label{sec:exp_ablation}

We report additional ablations in Appendix~\ref{sec:ablation} and summarize the main findings here.
Table~\ref{tab:ablation_oavc} validates the effectiveness of the proposed \emph{where/how} velocity control.

\textbf{Component ablation.}
Stage-1 alone slightly improves prompt faithfulness, as reflected by higher CLIP scores, but does not by itself improve structure or background preservation (it is marginally worse than DNAEdit on these metrics), since it only constructs the reference interface and performs no localized injection.
The major improvement comes from Stage-2, which sharply reduces Struct.\ Dist.\ and improves background metrics, confirming that drift is primarily caused by step-wise accumulation during integration.
Within Stage-2, object-support masking accounts for most of the preservation gain, while the boundary ring provides a smaller but consistent improvement by softening the foreground/background transition.
Removing either the flow-orthogonal projection or the confidence gate degrades structure and preservation, showing that the \emph{how} constraint suppresses drift-aligned or unstable residuals.
With identical support, full OAVC reduces Structure Distance by 22.0\% versus mask-only gating and 34.6\% versus final latent blending while matching the latter's LPIPS/MSE within 0.6\%; Sec.~\ref{sec:appendix_fair_controls} gives the full attribution and human study.

\textbf{Hyperparameter robustness.}
We sweep the main Stage-1 hyperparameters, foreground relaxation $\lambda_{\mathrm{fg}}$ and background suppression $\rho_{\mathrm{bg}}$, in Fig.~\ref{fig:fg_bg}.
Across SD3.5 and FLUX, moderate foreground relaxation with small background suppression gives stable LPIPS/MSE performance.
We therefore use $\lambda_{\mathrm{fg}}{=}0.6$ and $\rho_{\mathrm{bg}}{=}0.1$ by default.
Additional support-related hyperparameter studies are provided in Appendix~\ref{sec:ablation}.

\textbf{Transfer robustness to another velocity editor.}
We further apply OAVC-style Stage-2 controlled injection to FlowEdit~\citep{kulikov2025flowedit} to test whether the proposed control principle is tied to our base pipeline.
As shown in Tab.~\ref{tab:flowedit_transfer}, the transfer strongly improves FlowEdit-FLUX, reducing Struct.\ Dist.\ from 27.82 to 9.32 and increasing PSNR from 21.96 dB to 30.77 dB.
On SD3, it improves background preservation, e.g., PSNR increases from 22.22 dB to 27.65 dB and MSE decreases from 85.96 to 21.57, but Struct.\ Dist.\ worsens from 27.12 to 35.72.
This suggests that the \emph{where/how} control is broadly useful, while stable structure preservation also benefits from the Stage-1 reference interface.

\section{Extended Evaluation and Analysis}
\label{sec:extended_evaluation}
\subsection{Controlled Same-Support Baselines and Human Evaluation}
\label{sec:appendix_fair_controls}

\paragraph{Disentangling object localization from trajectory control.}
The comparison between global DNAEdit and OAVC conflates two sources of
improvement: the introduction of an explicit object support and the way in
which the prompt-induced residual is integrated along the generation
trajectory.
To disentangle these effects, Table~\ref{tab:same_support_controls} compares
three localized variants using the \emph{identical cached SAM3 support}.
Consequently, differences among these variants arise from how the shared
support is applied, rather than from differences in localization quality.

As a post-hoc spatial-control baseline, we first run unconstrained DNAEdit and
then apply a single hard blend in the final latent space:
\begin{equation}
Z_0^{\mathrm{blend}}
=
M\odot Z_0^{\mathrm{DNA}}
+
(1-M)\odot E(I_{\mathrm{src}}),
\end{equation}
where $M$ denotes the thresholded support at the latent resolution and
$E(I_{\mathrm{src}})$ is the VAE encoding of the source image.
This baseline directly restores the source latent outside the support, but,
unlike OAVC, does not constrain intermediate velocities or numerical solver
steps.

\begin{table}[t]
\centering
\small
\setlength{\tabcolsep}{4pt}
\caption{
Controlled comparison on PIE-Bench using SD3.5.
All localized variants use the identical cached SAM3 support, thereby
isolating the effect of trajectory-level control from that of object
localization.
Metrics follow the scaling convention of Table~\ref{tab:pie_full}.
}
\label{tab:same_support_controls}
\resizebox{\linewidth}{!}{
\begin{tabular}{l|ccccc}
\toprule
Method
& Struct.$\downarrow$
& BG LPIPS$\downarrow$
& BG MSE$\downarrow$
& CLIP$_{\rm whole}\uparrow$
& CLIP$_{\rm edit}\uparrow$ \\
\midrule
DNAEdit (global)
& 14.19 & 74.57 & 32.76 & \textbf{25.63} & \textbf{22.71} \\
Final latent hard blend
& 6.28 & \textbf{24.29} & \textbf{9.56} & 24.46 & 21.98 \\
Mask-only velocity gating
& 5.27 & 25.78 & 10.48 & 24.55 & 21.87 \\
\textbf{Full OAVC}
& \textbf{4.11} & 24.40 & 9.61 & 24.21 & 21.51 \\
\bottomrule
\end{tabular}}
\end{table}

The transition from global DNAEdit to mask-only velocity gating reduces
Structure Distance from 14.19 to 5.27, confirming that explicit object support
accounts for a substantial part of the preservation improvement.
However, localization alone does not explain the full gain.
Under the identical support, full OAVC further reduces Structure Distance from
5.27 to 4.11, corresponding to a 22.0\% relative improvement over mask-only
velocity gating.

Final latent blending has a construction-level advantage in background pixel
metrics because it explicitly overwrites the final background with the encoded
source latent.
Despite operating entirely within the generation trajectory and performing no
final overwrite, OAVC matches its BG LPIPS and BG MSE within 0.6\%, while
reducing Structure Distance from 6.28 to 4.11, a 34.6\% relative improvement.
This structural improvement is accompanied by a modest decrease in CLIP-based
editability, reflecting the preservation--editability trade-off analyzed in
Section~\ref{sec:appendix_clip_tradeoff}.

Overall, these controlled comparisons separate the two contributions:
explicit support provides the primary localization benefit, whereas OAVC's
constrained, spatially weighted in-trajectory integration yields an additional
improvement in structural preservation under the same support.

\paragraph{Human evaluation.}
We conduct a blinded study on 100 PIE-Bench examples.
Three independent annotators first choose which method best realizes the edit and which best preserves the background; we report per-example majority votes, with ties and cases lacking two agreeing votes separated.
Two annotators additionally label every output as Yes, Partially, or No for absolute task success, yielding 200 judgments per method.

\begin{table}[t]
\centering
\small
\setlength{\tabcolsep}{4pt}
\caption{Human evaluation. Top: majority-vote preference counts over 100 examples. Bottom: absolute task-success rates over 200 judgments per method.}
\label{tab:human_study}
\resizebox{\linewidth}{!}{
\begin{tabular}{l|ccc|cc}
\toprule
Criterion & OAVC & DNAEdit & Final blend & None/Tie & No consensus \\
\midrule
Edit realization & 15 & 22 & 28 & 17 & 18 \\
Background preservation & 74 & 0 & 20 & 0 & 6 \\
\midrule
Absolute task success & Yes & Partially & Yes+Partially & No & -- \\
\midrule
OAVC & 59.0\% & 20.5\% & 79.5\% & 20.5\% & -- \\
DNAEdit & 65.5\% & 14.5\% & 80.0\% & 20.0\% & -- \\
Final blend & 55.5\% & 21.5\% & 77.0\% & 23.0\% & -- \\
\bottomrule
\end{tabular}}
\end{table}

OAVC wins 74 of 94 decisive background-preservation comparisons (78.7\%; exact binomial $p<10^{-7}$).
For edit realization, the 65 decisive outcomes favor OAVC/DNAEdit/final blend 15/22/28; an exact multinomial test against equal preference gives $p=0.147$, and exploratory exact binomial comparisons give $p=0.324$ for OAVC versus DNAEdit and $p=0.066$ for OAVC versus final blending.
Thus, OAVC tends to be more conservative, but the relative edit-realization differences are not significant at 0.05 in this study; non-significance is not evidence of equivalence.
The absolute Yes-or-Partially rate is 79.5\% for OAVC versus 80.0\% for DNAEdit (paired exact McNemar $p=1.00$), indicating that lower CLIP does not correspond to widespread task failure.

\subsection{Robustness to Input-Support Corruption}
\label{sec:appendix_support_corruption}

We evaluate robustness to errors in the input object prior on a fixed subset of 100 PIE-Bench examples.
Starting from the cached SAM3 prior $M_0$, we construct a perturbed prior
\begin{equation}
M_r=
\begin{cases}
\operatorname{Erode}(M_0;|r|), & r<0,\\
M_0, & r=0,\\
\operatorname{Dilate}(M_0;r), & r>0,
\end{cases}
\qquad r\in\{-4,-2,0,+2,+4\},
\label{eq:support_corruption}
\end{equation}
where each morphology step is applied on the VAE latent grid.
The perturbation is introduced before OAVC constructs its editable and boundary supports.
Every condition subsequently uses the same support-construction parameters, including
$r_{\mathrm{fg}}=2$ and $r_{\mathrm{bd}}=4$, as well as identical generation hyperparameters and sample-wise random seeds.
Thus, $r$ measures corruption of the input prior and should not be confused with the fixed foreground-support margin $r_{\mathrm{fg}}$.

\begin{figure}[t]
\centering
\includegraphics[width=\linewidth]{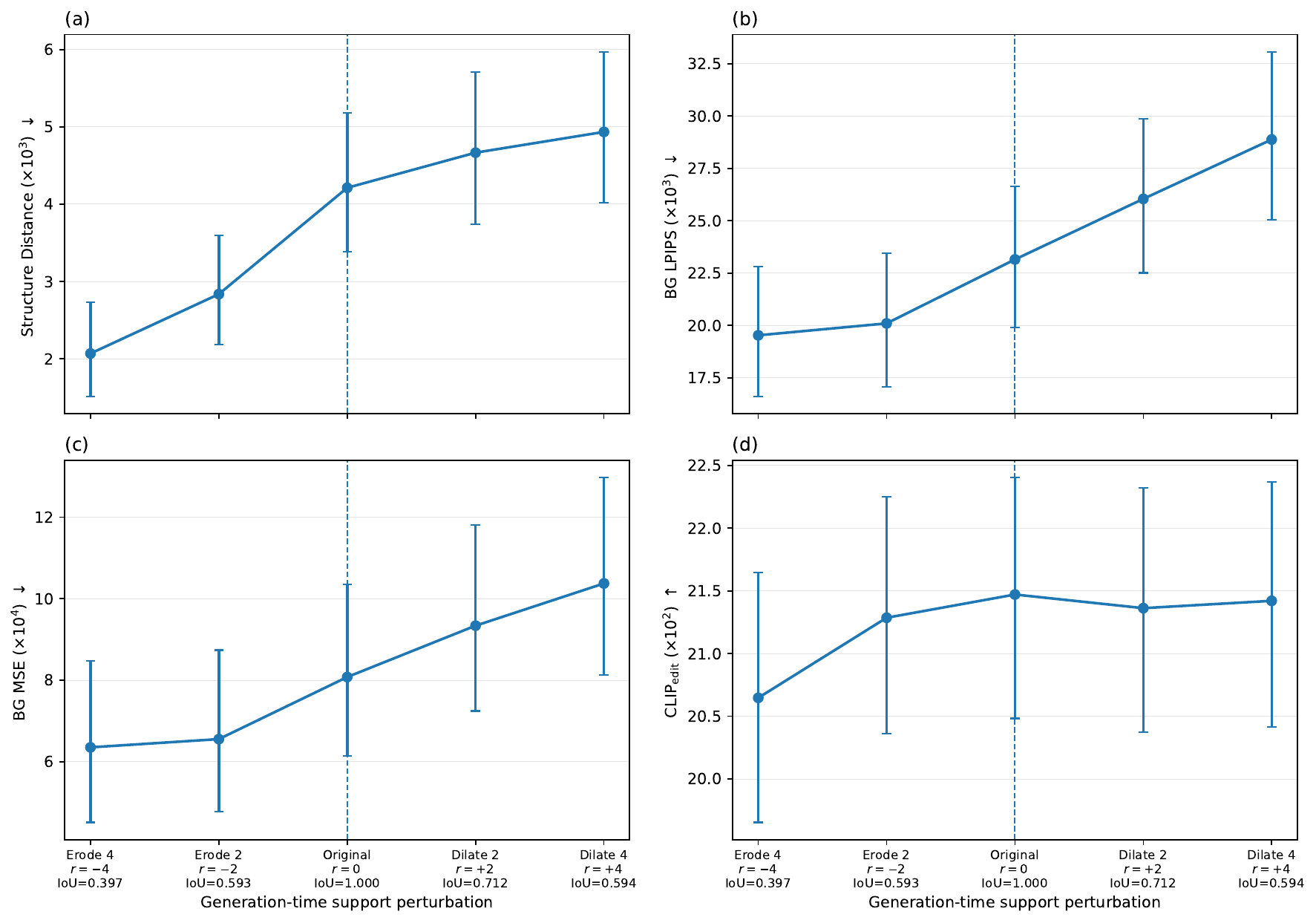}
\caption{Robustness to generation-time input-prior corruption on a fixed 100-example PIE-Bench subset. Points denote means and error bars denote bootstrap 95\% confidence intervals. The dashed line marks the unperturbed input prior ($r=0$). All conditions retain the same OAVC support-construction parameters, including $r_{\mathrm{fg}}=2$ and $r_{\mathrm{bd}}=4$. Region-conditioned metrics use the same independent PIE-Bench annotation; Structure Distance is global. Erosion restricts edit coverage, whereas dilation increasingly exposes non-target regions to semantic residuals.}
\label{fig:support_corruption_curve}
\end{figure}

Only the generation-time prior is perturbed:
\begin{equation}
M_{\mathrm{gen}}=M_r,
\qquad
M_{\mathrm{eval}}=M_{\mathrm{PIE}},
\label{eq:support_generation_evaluation}
\end{equation}
where $M_{\mathrm{PIE}}$ is the fixed PIE-Bench annotation shared by all conditions.
IoU and area ratio characterize $M_r$ relative to $M_0$; all region-conditioned output metrics use the independent PIE-Bench annotation, while Structure Distance remains a global, mask-independent metric.
Table~\ref{tab:support_corruption} and Fig.~\ref{fig:support_corruption_curve} jointly show direction-dependent degradation: erosion restricts edit coverage, whereas dilation exposes more non-target area to semantic residuals.

\begin{table}[t]
\centering
\small
\setlength{\tabcolsep}{3.5pt}
\caption{Robustness to input-prior corruption on 100 PIE-Bench examples. The cached SAM3 prior $M_0$ is eroded or dilated before the unchanged OAVC support-construction step. All conditions retain $r_{\mathrm{fg}}=2$ and $r_{\mathrm{bd}}=4$. IoU and area ratio characterize $M_r$ relative to $M_0$; region-conditioned metrics use the same independent PIE-Bench annotation, while Structure Distance is global. Metrics are scaled as in Table~\ref{tab:pie_full}.}
\label{tab:support_corruption}
\resizebox{\linewidth}{!}{
\begin{tabular}{l|ccc|cccc}
\toprule
Input-prior perturbation & IoU w.r.t. $M_0$ & Area ratio & Empty & Struct.$\downarrow$ & BG LPIPS$\downarrow$ & BG MSE$\downarrow$ & CLIP$_{\rm edit}\uparrow$ \\
\midrule
Erode 4 & 0.397 & 0.397$\times$ & 13\% & 2.07 & 19.53 & 6.35 & 20.65 \\
Erode 2 & 0.593 & 0.593$\times$ & 6\% & 2.84 & 20.10 & 6.55 & 21.29 \\
Original & 1.000 & 1.000$\times$ & 0\% & 4.21 & 23.15 & 8.08 & 21.47 \\
Dilate 2 & 0.712 & 1.581$\times$ & 0\% & 4.67 & 26.04 & 9.34 & 21.36 \\
Dilate 4 & 0.594 & 2.236$\times$ & 0\% & 4.93 & 28.88 & 10.38 & 21.42 \\
\bottomrule
\end{tabular}}
\end{table}

The results show a clear direction-dependent degradation pattern.
Erosion restricts the spatial coverage of semantic injection and makes editing increasingly conservative.
Relative to the unperturbed condition, Erode-2 changes the aggregate CLIP$_{\rm edit}$ score by only 0.84\% (21.47 to 21.29), but 6/100 input supports become empty.
Under Erode-4, the mean support area falls to 39.7\% of the original prior, 13/100 supports become empty, and CLIP$_{\rm edit}$ decreases to 20.65.
These results indicate that aggregate CLIP scores can understate sample-specific incomplete-edit failures caused by under-segmentation.

Dilation produces the complementary failure mode.
At Dilate-4, the input support expands to 2.236 times its original area.
Compared with the unperturbed condition, BG LPIPS increases by 24.7\% (23.15 to 28.88) and BG MSE increases by 28.4\% (8.08 to 10.38), while CLIP$_{\rm edit}$ changes by only 0.23\%.
Thus, excessive support expansion provides little aggregate semantic benefit while exposing substantially more non-target content to semantic residuals.
Notably, Erode-2 and Dilate-4 have nearly identical mean IoUs (0.593 versus 0.594) but substantially different background errors.
This demonstrates that IoU alone does not characterize support corruption: under-segmentation mainly limits edit coverage, whereas over-segmentation weakens spatial locality and increases non-target drift.
Overall, the aggregate metrics are relatively insensitive to moderate input-prior perturbations, but OAVC remains dependent on support quality.
Severe erosion can remove the editable region and cause incomplete edits, whereas excessive dilation increases non-target drift.

\subsection{Qualitative Generalization Check on GEdit-Bench}
\label{sec:appendix_gedit}

\begin{figure}[t]
\centering
\includegraphics[width=.82\linewidth]{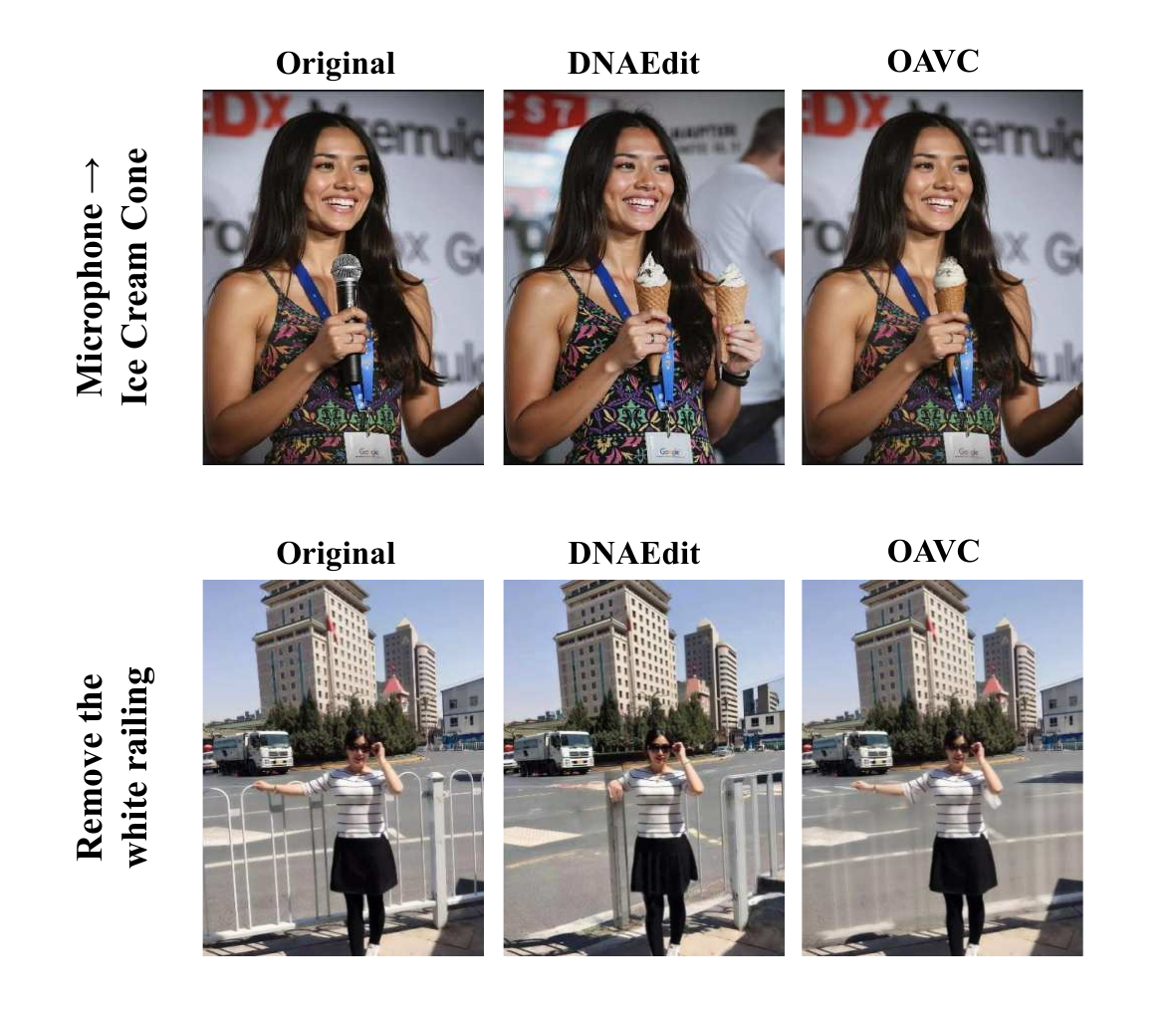}
\caption{Qualitative examples on GEdit-Bench. Columns show Original, DNAEdit, and OAVC. Top: microphone $\rightarrow$ ice-cream cone. Bottom: remove the white railing. OAVC better localizes the requested edit and preserves surrounding content. These two cases provide a qualitative generalization check, not full-benchmark evaluation.}
\label{fig:gedit_examples}
\end{figure}

We additionally test two representative real-world instructions from GEdit-Bench~\citep{jiang2026geditbench}, covering localized object replacement and removal.
Figure~\ref{fig:gedit_examples} compares the source, global DNAEdit, and OAVC for changing a microphone into an ice-cream cone and removing a white railing.
In both examples, OAVC confines the requested change more closely to the intended region while preserving non-target content.
These examples extend the qualitative scope beyond PIE-Bench, but they are not a substitute for benchmark-wide quantitative evaluation.

\subsection{Editability--Preservation Trade-off and CLIP Similarity}
\label{sec:appendix_clip_tradeoff}

This section further analyzes the CLIP similarity trade-off observed in the main results.
Although OAVC substantially improves background preservation and structural stability, its CLIP$_\text{whole}$ and sometimes CLIP$_\text{edit}$ scores can be lower than global-steering baselines.
This should not be interpreted as OAVC simply weakening the edit.
Instead, the difference reflects a stricter object-centric objective: target semantics should be injected into the editable object, but should not leak into non-target regions.

\paragraph{Why global CLIP can favor background leakage.}
CLIP similarity measures image-text alignment between a generated image and the target prompt.
Let $f_I(\cdot)$ and $f_T(\cdot)$ denote the CLIP image and text encoders, respectively.
For an image $X$ and target prompt $\psi_{\mathrm{tgt}}$, we compute
\begin{equation}
\mathrm{Sim}(X,\psi_{\mathrm{tgt}})
=
\left\langle
\frac{f_I(X)}{\|f_I(X)\|_2},
\frac{f_T(\psi_{\mathrm{tgt}})}{\|f_T(\psi_{\mathrm{tgt}})\|_2}
\right\rangle .
\label{eq:clip_sim}
\end{equation}
For object-centric editing, however, the target prompt usually describes the edited object while the desired output should preserve the non-target background.
A global editor may increase CLIP similarity by spreading target-related visual changes beyond the object support.
Such changes can improve whole-image target-prompt alignment while simultaneously damaging background preservation.
Therefore, a higher CLIP$_\text{whole}$ score does not necessarily indicate a better localized edit.

\paragraph{OAVC does not suppress the full edited field.}
OAVC does not project or suppress the entire edited velocity field.
The final edited velocity has the form
\begin{equation}
v_t^{\mathrm{edit}}
=
v_t^{\mathrm{src}}
+
W_t^{\mathrm{edit}}
\odot
\left(g_t\Delta v_t^{\mathrm{inj}}\right),
\label{eq:clip_tradeoff_velocity}
\end{equation}
where $v_t^{\mathrm{src}}$ is preserved as the base transport field and only the additional prompt-induced residual is spatially localized and constrained.
Thus, OAVC is not a low-guidance variant that uniformly weakens the edit.
It keeps the source transport and controls where the additional target semantic residual is allowed to accumulate.

This also clarifies the role of the flow-orthogonal projection.
The projection is applied to the residual direction, not to the whole edited field.
Therefore, structural information from the source trajectory is still retained through $v_t^{\mathrm{src}}$, while target-specific changes are introduced through offset-aligned target evaluation, reference-guided blending, and object-localized residual injection.
In our ablation, removing projection only slightly recovers CLIP similarity but clearly worsens structure and background preservation, suggesting that projection mainly acts as a stabilizer rather than the primary source of the CLIP trade-off.

\paragraph{Foreground/background semantic gain decomposition.}
To test whether OAVC merely weakens editing, we decompose target-prompt similarity gain using the method-independent PIE-Bench foreground/background annotation, not OAVC's generation support.
Let $M_{\mathrm{fg}}$ be the edited object support and $M_{\mathrm{bg}}=1-M_{\mathrm{fg}}$ be the non-target background support.
We define a region extraction operator $\mathcal C_M(X)$ that crops or masks image $X$ to the region indicated by $M$ using the same preprocessing for all methods.
The region-level target similarity is
\begin{equation}
\mathrm{Sim}_{M}(X,\psi_{\mathrm{tgt}})
=
\mathrm{Sim}\bigl(\mathcal C_M(X),\psi_{\mathrm{tgt}}\bigr).
\label{eq:regional_clip_sim}
\end{equation}
Given the source image $X^{\mathrm{src}}$ and the edited image $X^{\mathrm{edit}}$, we define foreground and background target-semantic gains as
\begin{equation}
\mathrm{FG\text{-}Gain}
=
\mathrm{Sim}_{M_{\mathrm{fg}}}(X^{\mathrm{edit}},\psi_{\mathrm{tgt}})
-
\mathrm{Sim}_{M_{\mathrm{fg}}}(X^{\mathrm{src}},\psi_{\mathrm{tgt}}),
\label{eq:fg_gain}
\end{equation}
\begin{equation}
\mathrm{BG\text{-}Gain}
=
\mathrm{Sim}_{M_{\mathrm{bg}}}(X^{\mathrm{edit}},\psi_{\mathrm{tgt}})
-
\mathrm{Sim}_{M_{\mathrm{bg}}}(X^{\mathrm{src}},\psi_{\mathrm{tgt}}).
\label{eq:bg_gain}
\end{equation}
FG-Gain measures how much target semantics are added to the intended object region.
BG-Gain measures how much target semantics leak into the non-target background.
We further define the localization gap and leakage ratio as
\begin{equation}
\mathrm{FG\text{-}BG\ Gap}
=
\mathrm{FG\text{-}Gain}
-
\mathrm{BG\text{-}Gain},
\qquad
\mathrm{Leakage\ Ratio}
=
\frac{\mathrm{BG\text{-}Gain}}
{\mathrm{FG\text{-}Gain}+\epsilon}.
\label{eq:fg_bg_gap_leakage}
\end{equation}
A larger FG-BG Gap and a smaller Leakage Ratio indicate stronger object-localized semantic editing.

\begin{table}[t]
\centering
\footnotesize
\setlength{\tabcolsep}{5pt}
\renewcommand{\arraystretch}{1.08}
\caption{
Foreground/background target-semantic decomposition.
FG-Gain and BG-Gain measure target-prompt gain in foreground and background
regions, respectively, while FG-BG Gap measures localization strength.
}
\label{tab:clip_fg_bg_gain}
\begin{tabular}{l|cc|cc|cc}
\toprule
\multirow{2}{*}{Method} &
\multicolumn{2}{c|}{FG-Gain} &
\multicolumn{2}{c|}{BG-Gain} &
\multicolumn{2}{c}{FG-BG Gap} \\
\cmidrule(lr){2-3}
\cmidrule(lr){4-5}
\cmidrule(lr){6-7}
& Mean$\uparrow$ & Std
& Mean$\downarrow$ & Std
& Mean$\uparrow$ & Std \\
\midrule
DNAEdit (global)
& \textbf{0.0189} & 0.0299
& 0.0108 & 0.0245
& 0.0080 & 0.0336 \\
OAVC
& 0.0130 & 0.0261
& \textbf{0.0008} & 0.0123
& \textbf{0.0122} & 0.0260 \\
\bottomrule
\end{tabular}
\end{table}

\paragraph{Analysis.}
Tab.~\ref{tab:clip_fg_bg_gain} shows that OAVC does not uniformly remove target semantics.
Compared with DNAEdit, the foreground target gain decreases moderately from $0.0189$ to $0.0130$.
In contrast, the background target gain is reduced much more strongly, from $0.0108$ to $0.0008$.
Using Eq.~\eqref{eq:fg_bg_gap_leakage}, the background leakage ratio decreases from approximately $57.5\%$ for DNAEdit to $6.3\%$ for OAVC.
This indicates that the main effect of OAVC is not global edit suppression.
Instead, OAVC mainly removes target semantics that global steering tends to leak into non-target regions.

The larger FG-BG Gap further supports this interpretation.
Although DNAEdit can obtain higher whole-image target similarity, part of this gain comes from changing the background.
OAVC produces a more localized semantic update: target semantics are concentrated in the foreground while non-target regions remain source-consistent.
Therefore, the observed CLIP trade-off reflects the difference between global target-prompt alignment and object-centric editing quality.

\paragraph{Relation to CLIP$_\text{whole}$ and CLIP$_\text{edit}$.}
CLIP$_\text{whole}$ evaluates target-prompt alignment on the entire image and is therefore sensitive to both desired foreground changes and undesired background leakage.
CLIP$_\text{edit}$ focuses more on the edited region, but it can still be affected by mask quality, boundary content, and the fact that object-centric editing intentionally preserves some source structure.
For this reason, CLIP similarity should be interpreted together with structure and background-preservation metrics.
A method that maximizes CLIP by changing the whole image may be undesirable when the instruction asks for a localized edit.

\paragraph{Relation to projection and localization.}
The CLIP decrease is not solely caused by the flow-orthogonal projection.
Projection is an auxiliary stabilizer that suppresses drift-inducing residual components.
The larger CLIP change appears when Stage-2 object-localized control is enabled, because target semantics outside the object support are explicitly blocked.
This is consistent with the purpose of OAVC: the method prioritizes localized semantic editing and background stability over globally maximizing target-prompt similarity.

\begin{figure}[t]
\centering
\includegraphics[width=\linewidth]{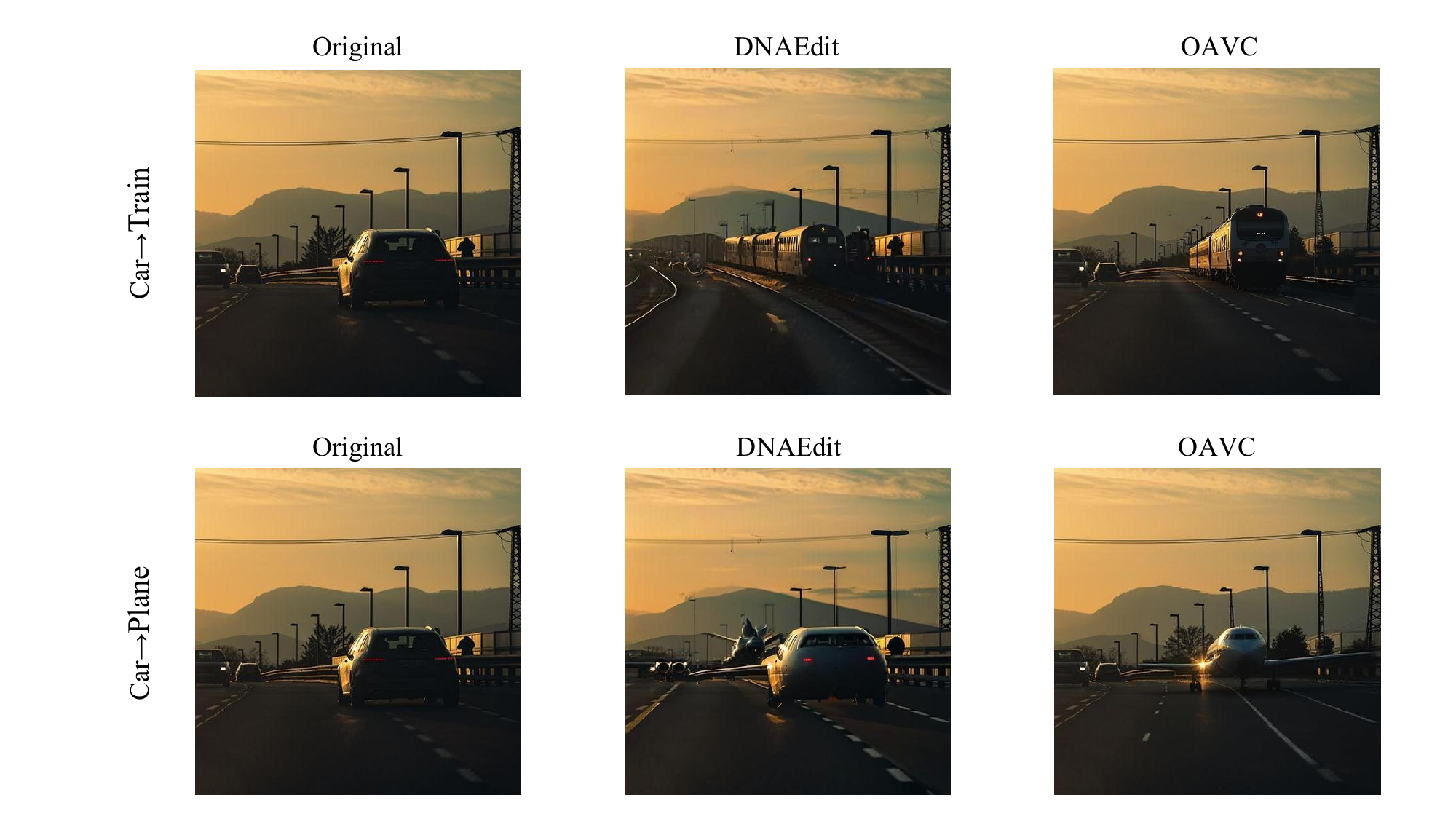}
\caption{
Large-deformation object replacements from the same source image.
Rows show car-to-train and car-to-plane edits, while columns compare the original image, DNAEdit, and OAVC.
DNAEdit introduces more extensive changes to the road and surrounding scene and, in the car-to-plane case, incompletely removes the source car.
OAVC concentrates the transformation around the source object while more closely preserving non-target scene content.
}
\label{fig:large_geometry_edits}
\end{figure}

\paragraph{Substantial object-centric replacements.}
Fig.~\ref{fig:large_geometry_edits} examines two substantial object-centric replacements involving pronounced semantic and geometric changes.
Starting from the same source image, we perform car-to-train and car-to-plane edits and compare OAVC with the global DNAEdit baseline.
For car-to-train, DNAEdit produces the target object but also noticeably modifies the road layout and surrounding scene structures.
For car-to-plane, DNAEdit retains much of the source car while introducing additional aircraft-related content.
In comparison, OAVC concentrates the transformation around the source object and more closely preserves the surrounding road, mountains, sky, and lamp posts.
These examples demonstrate that the proposed velocity projection can accommodate substantial semantic and geometric transformations when the target object is adequately covered by the editable support.
Nevertheless, the achievable edit coverage remains dependent on support quality: an overly tight support may restrict target geometry that needs to extend substantially beyond the source object.

\paragraph{Practical implication.}
OAVC operates on the editability--preservation frontier.
When stronger foreground transformation is desired, the object support can be slightly enlarged or foreground anchoring can be relaxed.
When background leakage occurs, the support can be tightened or background suppression can be strengthened.
This provides a structured way to navigate the trade-off, rather than treating the CLIP decrease as an uncontrolled failure mode.
For aggressive shape changes, the fixed foreground-relaxation coefficient may be conservative when the desired geometry must extend outside a tight support.
A time-dependent schedule $\lambda_{\mathrm{fg}}(t)$ that permits greater foreground freedom during early high-noise steps and strengthens anchoring later is directly compatible with OAVC, but was not evaluated here.
Likewise, because the orthogonal projection constrains only the added residual and retains $v_t^{\mathrm{src}}$ as the base transport, it primarily stabilizes structure; it may nevertheless suppress a useful target component aligned with the source flow, motivating a future confidence-adaptive soft projection.
\subsection{Multi-Object and Compositional Editing}
\label{sec:appendix_multi_object}

This section discusses how OAVC can be naturally extended from single-object editing to multi-object and simple compositional edits.
The main formulation of OAVC does not assume that the editable support contains only one object.
Since the method operates on velocity-level spatial supports, multiple object supports can be incorporated without changing the underlying where/how control principle.

\begin{figure}[t]
    \centering
    \includegraphics[width=\linewidth]{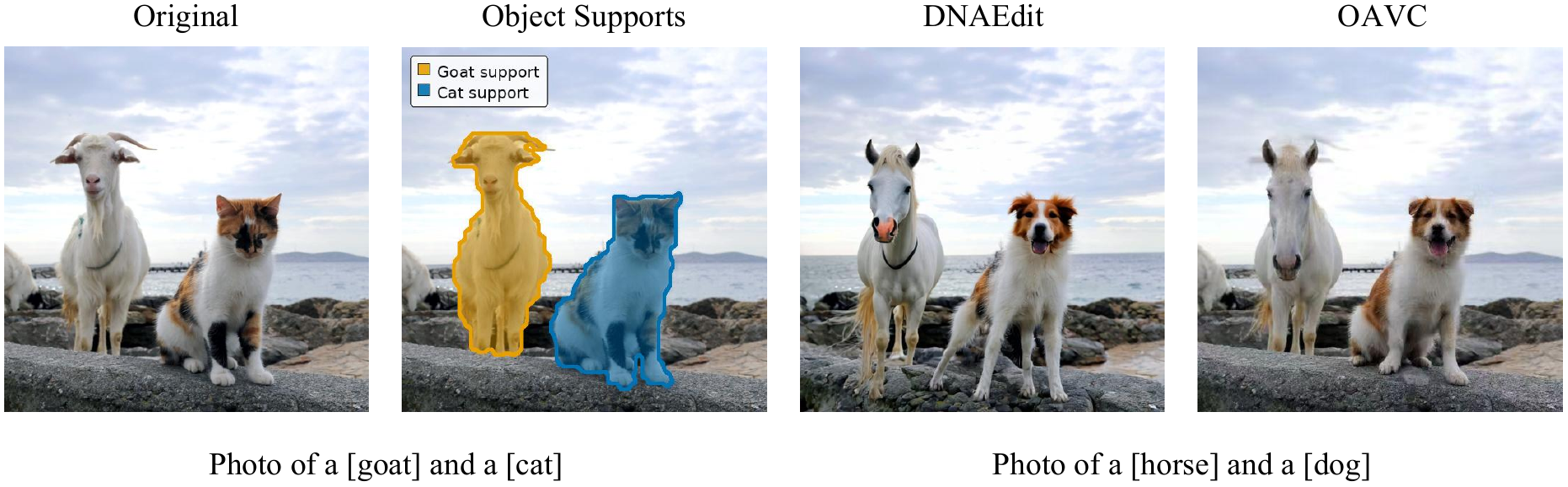}
    \caption{
    Representative two-object edit from a goat and a cat to a horse and a dog.
    The orange and blue overlays indicate the source goat and cat supports, respectively; their union is used as the editable support.
    Both methods modify the two target objects, while OAVC more closely retains the original scene layout and non-target background.
    }
    \label{fig:multi_object_edit}
\end{figure}

\paragraph{Multi-object editing.}
Fig.~\ref{fig:multi_object_edit} presents a representative two-object edit from
``photo of a [goat] and a [cat]''
to
``photo of a [horse] and a [dog]''.
The second column visualizes the object supports identified for the source goat and cat in orange and blue, respectively; their union defines the editable region used during generation.
Both DNAEdit and OAVC respond to the two target concepts, demonstrating that the underlying flow-based editor can perform compositional object replacement.
However, the global semantic injection of DNAEdit produces more pronounced changes in object appearance, layout, and nearby scene content.
OAVC confines the prompt-induced velocity correction to the combined object support, changing both foreground objects while more closely retaining the original spatial arrangement and non-target background, including the rocks, ocean, and sky.
This example provides a proof of concept that the same velocity-level control mechanism can operate on multiple object supports, although broader evaluation of multi-object correspondence and interactions remains future work.

\paragraph{Object-wise semantic injection.}
In Stage-2, each object support is converted into its own edit and reference supports,
$W_{t,i}^{\mathrm{edit}}$ and $W_{t,i}^{\mathrm{ref}}$, following the same construction as Appendix~\ref{sec:appendix_object_prior}.
For each object, we compute an object-specific semantic residual and apply the same safe injection operator:
\begin{equation}
u_t^{\mathrm{multi}}
=
\sum_{i=1}^{K}
W_{t,i}^{\mathrm{edit}}
\odot
\Gamma_t\!\left(\Delta v_{t}^{(i)};\mathcal R_t\right),
\label{eq:multi_object_injection}
\end{equation}
where $\Delta v_{t}^{(i)}$ denotes the prompt-induced residual for the $i$-th local edit.
The final edited velocity follows the same form as the single-object case,
except that the controlled residual is summed over object supports:
\begin{equation}
v_t^{\mathrm{edit}}
=
v_t^{\mathrm{src}}
+
u_t^{\mathrm{multi}} .
\label{eq:multi_object_edit_velocity}
\end{equation}
If object supports overlap, the weights can be normalized by
$\bar W_{t,i}^{\mathrm{edit}}
=
W_{t,i}^{\mathrm{edit}}/
(\sum_j W_{t,j}^{\mathrm{edit}}+\epsilon)$
to avoid over-injection in overlapping regions.
In the simple case where a single joint target prompt is used, the formulation reduces to applying OAVC on the union support.

\paragraph{Reference update.}
The moving reference anchor can be updated analogously by accumulating only object-supported semantic displacement:
\begin{equation}
Z_{t+1}^{\mathrm{ref}}
=
Z_t^{\mathrm{ref}}
+
(\sigma_{t+1}-\sigma_t)
\sum_{i=1}^{K}
\left(
W_{t,i}^{\mathrm{ref}}
\odot
\Gamma_t\!\left(\Delta v_t^{(i)};\mathcal R_t\right)
\right).
\label{eq:multi_object_ref_update}
\end{equation}
Thus, the background remains anchored by the union complement, while each editable object receives its own localized semantic update.

\paragraph{Scope.}
While this example demonstrates the feasibility of multi-object OAVC, more complex compositional scenarios remain an important direction for future work.
Challenging cases include strongly overlapping objects, interactions between edited objects, and edits requiring global scene-level reasoning.
These cases may require more structured semantic layouts or object-relation priors beyond independent object supports.

\begin{table}[t]
  \caption{Runtime and peak GPU memory (per image) for DNAEdit vs.\ OAVC on two rectified-flow backbones.}
  \label{tab:efficiency}
  \begin{center}
    \begin{small}
      \setlength{\tabcolsep}{3.8pt} % tighten columns to avoid overflow
      \begin{tabular}{lcc}
        \toprule
        Backbone/Method & Time (s/img)$\downarrow$ & Peak Mem (MiB)$\downarrow$ \\
        \midrule
        SD3.5 / DNAEdit & 3.08 & 19{,}459 \\
        SD3.5 / Ours    & 3.12 & 23{,}139 \\
        FLUX / DNAEdit  & 9.01 & 35{,}839 \\
        FLUX / Ours     & 9.85 & 39{,}571 \\
        \bottomrule
      \end{tabular}
    \end{small}
  \end{center}
  \vskip -0.1in
\end{table}
\subsection{Efficiency and Overhead}
\label{sec:appendix_efficiency}

Tab.~\ref{tab:efficiency} reports end-to-end runtime and peak GPU memory per image for DNAEdit and OAVC on SD3.5 and FLUX.
OAVC introduces additional inference-time operations (reference-anchor maintenance, spatial weighting, ring high-pass filtering, and confidence gating),
which lead to a modest runtime and memory overhead over DNAEdit under the same backbone and sampling configuration.
In our implementation, the SAM3 mask is computed once per image (or per video) and cached across solver steps; thus the reported overhead primarily reflects OAVC's control logic during sampling rather than repeated segmentation calls.

\subsection{Limitations and Scope}
\label{sec:main_limitations}
OAVC relies on a meaningful object support that localizes where semantic velocity updates may act.
When the support becomes nearly global, the spatial selectivity of Stage~2 vanishes and the method approaches global velocity steering; it is therefore not designed for full-image transformations such as global style transfer, overall color grading, or domain-level appearance changes.
Performance also depends on support quality: severe under-segmentation can produce incomplete edits, while excessive over-segmentation exposes non-target regions to semantic residuals, as quantified in Sec.~\ref{sec:appendix_support_corruption}.
The current post-hoc control uses a single final latent hard blend; stronger pixel-space or multi-step compositing controls remain to be evaluated.
Finally, Structure Distance is a global correspondence metric, and the present human study was not powered for reliable per-edit-category analysis.
Representative weak-benefit cases and broader qualitative results are provided in Appendices~\ref{sec:app_limitations} and~\ref{sec:appendix_more_qual}.

\section{Conclusion}

We presented \textbf{OAVC}, a training-free framework for object-centric image and video editing with rectified-flow backbones.
Rather than relying on post-hoc output masking, OAVC controls both \emph{where} prompt-induced residuals accumulate and \emph{how} they are injected during velocity integration.
By combining a background-anchored reference interface with object-localized safe semantic injection, OAVC reduces non-target drift and improves structural fidelity and boundary stability.
Experiments on PIE-Bench and video evaluations on DAVIS, complemented by qualitative checks on GEdit-Bench, demonstrate improved background preservation and temporal consistency while maintaining effective localized editability.
Controlled ablations, baseline comparisons, and human evaluation further characterize the preservation--editability trade-off and the dependence of OAVC on object-support quality.
Future work will explore adaptive support discovery and refinement together with multi-step compositing, reducing sensitivity to segmentation errors and better accommodating target objects with substantial geometric changes.
Integrating vision-language models for instruction grounding and multi-object correspondence may further extend OAVC toward more general and compositional editing scenarios.

% 
% ------------------------------------------------------------

\bibliographystyle{ACM-Reference-Format}
\bibliography{references}
\FloatBarrier
\clearpage
\appendix
\section*{Appendix}
\addcontentsline{toc}{section}{Appendix}
\section{Details of Algorithm}
\begin{algorithm}[H]
\caption{Stage-1: Background-Anchored Reference Interface}
\label{alg:stage1_main}
\begin{algorithmic}[1]
\STATE \textbf{Input:} source latent $Z_T$; source prompt $\psi_{\mathrm{src}}$; initial noise state $S_T\sim\mathcal N(0,I)$; object support $M_{\mathrm{fg}}$ and $M_{\mathrm{bg}}=1-M_{\mathrm{fg}}$; timesteps $\{\sigma_t\}_{t=0}^{T}$.
\STATE \textbf{Output:} reference interface $\mathcal R_t=\{Z_t,\Delta x_t^{\mathrm{safe}},v_t^{\mathrm{src}}\}$.

\vspace{2pt}
\FOR{$t=T-1,\dots,0$}
    \STATE $Z_t^{*}
    \leftarrow
    \dfrac{\sigma_t}{\sigma_{t+1}}Z_{t+1}
    +
    \left(1-\dfrac{\sigma_t}{\sigma_{t+1}}\right)S_{t+1}$
    \hfill\COMMENT{proxy state}
    
    \STATE $v_t^{\mathrm{src}}
    \leftarrow
    v_\theta(Z_t^{*},\psi_{\mathrm{src}})$,\quad
    $\Delta v_t^{\mathrm{DNA}}
    \leftarrow
    \dfrac{S_{t+1}-Z_{t+1}}{\sigma_{t+1}}
    -
    v_t^{\mathrm{src}}$
    \hfill\COMMENT{source residual}
    
    \STATE $\Delta v_t^{(1)}
    \leftarrow
    (1-\lambda_{\mathrm{fg}}M_{\mathrm{fg}})
    \odot
    \Delta v_t^{\mathrm{DNA}}$
    \hfill\COMMENT{foreground relaxation}
    
    \STATE $r_t
    \leftarrow
    M_{\mathrm{bg}}\odot \Delta v_t^{(1)}$,\quad
    $b_t
    \leftarrow
    M_{\mathrm{bg}}\odot v_t^{\mathrm{src}}$.
    
    \STATE $\tilde r_t
    \leftarrow
    (1-\rho_{\mathrm{bg}})r_t
    +
    \rho_{\mathrm{bg}}\Pi_\perp(r_t;b_t)$
    \hfill\COMMENT{background anchoring}
    
    \STATE $\Delta v_t^{\mathrm{safe}}
    \leftarrow
    M_{\mathrm{fg}}\odot \Delta v_t^{(1)}
    +
    \tilde r_t$.
    
    \STATE $S_t
    \leftarrow
    S_{t+1}
    +
    \sigma_{t+1}\Delta v_t^{\mathrm{safe}}$,\quad
    $Z_t
    \leftarrow
    Z_t^{*}
    +
    (\sigma_{t+1}-\sigma_t)\Delta v_t^{\mathrm{safe}}$
    \hfill\COMMENT{reference update}
    
    \STATE $\Delta x_t^{\mathrm{safe}}
    \leftarrow
    Z_t-Z_t^{*}$
    \hfill\COMMENT{cache offset}
\ENDFOR

\STATE \textbf{return} $\mathcal R_t=\{Z_t,\Delta x_t^{\mathrm{safe}},v_t^{\mathrm{src}}\}$
\end{algorithmic}
\end{algorithm}

\begin{algorithm}[H]
\caption{Stage-2: Object-Localized Safe Semantic Injection}
\label{alg:stage2_main}
\begin{algorithmic}[1]
\STATE \textbf{Input:} prompts $(\psi_{\mathrm{src}},\psi_{\mathrm{tgt}})$; reference interface $\mathcal R_t=\{Z_t,\Delta x_t^{\mathrm{safe}},v_t^{\mathrm{src}}\}$; supports $W_t^{\mathrm{edit}},W_t^{\mathrm{ref}}$; start step $t_s$.
\STATE \textbf{Output:} edited latent $Z_T^{\mathrm{edit}}$.

\vspace{2pt}
\STATE $Z_{t_s}^{\mathrm{edit}}\leftarrow Z_{t_s}$,\quad
$Z_{t_s}^{\mathrm{ref}}\leftarrow Z_T$
\hfill\COMMENT{initialize}

\vspace{2pt}
\FOR{$t=t_s,\dots,T-1$}
    \STATE $Z_t^{*\mathrm{edit}}
    \leftarrow
    Z_t^{\mathrm{edit}}
    +
    \Delta x_t^{\mathrm{safe}}$,\quad
    $v_t^{\mathrm{tgt}}
    \leftarrow
    v_\theta(Z_t^{*\mathrm{edit}},\psi_{\mathrm{tgt}})$
    \hfill\COMMENT{aligned target evaluation}
    
    \STATE $\Delta v_t
    \leftarrow
    v_t^{\mathrm{tgt}}
    -
    v_t^{\mathrm{src}}$,\quad
    $\bar{\Delta v}_t
    \leftarrow
    \Gamma_t^{\mathrm{ref}}(\Delta v_t;\mathcal R_t)$
    \hfill\COMMENT{safe semantic residual}
    
    \STATE $Z_{t+1}^{\mathrm{ref}}
    \leftarrow
    Z_t^{\mathrm{ref}}
    +
    (\sigma_{t+1}-\sigma_t)
    \left(
    W_t^{\mathrm{ref}}
    \odot
    \bar{\Delta v}_t
    \right)$
    \hfill\COMMENT{reference anchor}
    
    \STATE $v_t^{\mathrm{ref}}
    \leftarrow
    \dfrac{Z_t^{\mathrm{edit}}-Z_{t+1}^{\mathrm{ref}}}{1-\sigma_t}$,\quad
    $v_t^{\mathrm{fg}}
    \leftarrow
    \eta v_t^{\mathrm{tgt}}
    +
    (1-\eta)v_t^{\mathrm{ref}}$.
    
    \STATE $u_t
    \leftarrow
    W_t^{\mathrm{edit}}
    \odot
    \Gamma_t^{\mathrm{inj}}
    \left(
    v_t^{\mathrm{fg}}-v_t^{\mathrm{src}};
    \mathcal R_t
    \right)$
    \hfill\COMMENT{where/how injection}
    
    \STATE $v_t^{\mathrm{edit}}
    \leftarrow
    v_t^{\mathrm{src}}
    +
    u_t$,\quad
    $Z_{t+1}^{\mathrm{edit}}
    \leftarrow
    Z_t^{\mathrm{edit}}
    +
    (\sigma_{t+1}-\sigma_t)v_t^{\mathrm{edit}}$
    \hfill\COMMENT{integrate}
\ENDFOR

\STATE \textbf{return} $Z_T^{\mathrm{edit}}$
\end{algorithmic}
\end{algorithm}
\section{Implementation of the Safe Injection Operator}
\label{sec:appendix_gamma}

This section specifies the implementation of the safe injection operator
$\Gamma_t(\cdot;\mathcal R_t)$ used in Algorithms~\ref{alg:stage1_main} and~\ref{alg:stage2_main}.
The main text keeps $\Gamma_t$ abstract to emphasize the unified where/how control principle.
Here we provide the implementation-aligned form used in all experiments.

Recall that the Stage-1 reference interface is
$\mathcal R_t=\{Z_t,\Delta x_t^{\mathrm{safe}},v_t^{\mathrm{src}}\}$.
We use the same flow-orthogonal projection operator throughout OAVC:
\begin{equation}
\Pi_\perp(a;b)
=
a
-
\frac{\langle a,b\rangle}
{\langle b,b\rangle+\epsilon}b .
\label{eq:app_projection_operator}
\end{equation}
This operator removes from $a$ the component parallel to $b$.

\paragraph{Reference-anchor semantic operator.}
Given the target semantic residual
$\Delta v_t=v_t^{\mathrm{tgt}}-v_t^{\mathrm{src}}$,
we first remove the component aligned with the source flow:
\begin{equation}
\Delta v_t^{\perp}
=
\Pi_\perp(\Delta v_t;v_t^{\mathrm{src}}).
\label{eq:app_ref_projection}
\end{equation}
The boundary transition support is defined from the two spatial supports:
\begin{equation}
W_t^{\mathrm{ring}}
=
\mathrm{clip}\!\left(W_t^{\mathrm{ref}}-W_t^{\mathrm{edit}},0,1\right).
\label{eq:app_w_ring}
\end{equation}
We then apply boundary smoothing only on this ring:
\begin{equation}
\Delta \bar v_t
=
(1-W_t^{\mathrm{ring}})\odot \Delta v_t^{\perp}
+
W_t^{\mathrm{ring}}\odot \mathrm{HP}(\Delta v_t^{\perp}).
\label{eq:app_boundary_aware_residual}
\end{equation}
Finally, a scalar confidence gate $g_t\in[0,1]$ is applied:
\begin{equation}
\Gamma_t^{\mathrm{ref}}(\Delta v_t;\mathcal R_t)
=
g_t\,\Delta \bar v_t .
\label{eq:app_gamma_ref}
\end{equation}
This is the operator used to update the moving reference anchor:
\begin{equation}
Z_{t+1}^{\mathrm{ref}}
=
Z_t^{\mathrm{ref}}
+
(\sigma_{t+1}-\sigma_t)
\left(
W_t^{\mathrm{ref}}
\odot
\Gamma_t^{\mathrm{ref}}(\Delta v_t;\mathcal R_t)
\right).
\label{eq:app_ref_update_gamma}
\end{equation}

\paragraph{Final field-level injection operator.}
The reference-guided foreground velocity $v_t^{\mathrm{fg}}$ is an intermediate velocity signal.
Since the final integrated field is built from this velocity, we apply the same flow-orthogonal safety constraint at the field level:
\begin{equation}
\Delta v_t^{\mathrm{inj}}
=
\Pi_\perp
\left(
v_t^{\mathrm{fg}}-v_t^{\mathrm{src}};
v_t^{\mathrm{src}}
\right).
\label{eq:app_final_field_projection}
\end{equation}
The final safe injection operator is
\begin{equation}
\Gamma_t^{\mathrm{inj}}
\left(
v_t^{\mathrm{fg}}-v_t^{\mathrm{src}};
\mathcal R_t
\right)
=
g_t\,\Delta v_t^{\mathrm{inj}} .
\label{eq:app_gamma_inj}
\end{equation}
Thus the final edited velocity becomes
\begin{equation}
v_t^{\mathrm{edit}}
=
v_t^{\mathrm{src}}
+
W_t^{\mathrm{edit}}
\odot
\Gamma_t^{\mathrm{inj}}
\left(
v_t^{\mathrm{fg}}-v_t^{\mathrm{src}};
\mathcal R_t
\right).
\label{eq:app_final_edit_velocity_gamma}
\end{equation}
Therefore, outside the editable support where $W_t^{\mathrm{edit}}=0$, the integrated velocity is exactly the source velocity.
Inside the object support, only a gated and flow-orthogonal semantic residual is injected.

\paragraph{Unified operator view.}
Stage-2 uses the same safe-injection principle at two control points.
For the reference anchor, $\Gamma_t^{\mathrm{ref}}$ transforms the prompt residual
$\Delta v_t=v_t^{\mathrm{tgt}}-v_t^{\mathrm{src}}$ with flow-orthogonal projection and boundary-aware smoothing before it is accumulated through $W_t^{\mathrm{ref}}$.
For the final edited velocity, $\Gamma_t^{\mathrm{inj}}$ transforms the reference-guided residual
$v_t^{\mathrm{fg}}-v_t^{\mathrm{src}}$ with the field-level projection before it is injected through $W_t^{\mathrm{edit}}$.
Thus, $\Gamma_t^{\mathrm{ref}}$ and $\Gamma_t^{\mathrm{inj}}$ are not separate conceptual modules, but two applications of the same where/how control principle to the reference anchor and the final integrated velocity, respectively.

\section{Object Prior and Mask-Induced Velocity Constraints}
\label{sec:appendix_object_prior}

This appendix provides implementation-aligned details of the object support prior used by OAVC and the induced velocity-level spatial constraints.
The purpose of this appendix is to clarify two points.
First, OAVC is not tied to a specific mask provider such as SAM3; it only requires an object-centric support prior.
Second, the support is used to gate velocity updates during integration, not to overwrite latents or composite the final image.

\paragraph{Support prior, not hard editing mask.}
Given an input image $I$ and an edit request $(\psi_{\mathrm{src}},\psi_{\mathrm{tgt}})$, OAVC constructs an object support $M$ corresponding to the region where the semantic change is expected to occur.
In our default implementation, this support is estimated by SAM3 using the edited object concept extracted from the source/target prompts.
For prompt formats with explicit brackets, e.g., ``a photo of a [goat]'' $\rightarrow$ ``a photo of a [horse]'', the bracketed source object is used as the query concept.
Otherwise, we extract the edited noun or phrase from the prompt difference.

Importantly, SAM3 is only the default zero-shot support provider.
The OAVC formulation itself is support-source agnostic: the same velocity-control mechanism can use SAM3 masks, attention-derived masks, user-provided masks, soft masks, or interactive brush-refined masks.
In all cases, the resulting support is used only to construct $W_t$ in Eq.~\eqref{eq:where_how_factorization}.
It does not directly replace image pixels, overwrite latent regions, or perform final background compositing.

\paragraph{Latent-space support construction.}
The pixel-space support is resized to the latent resolution, producing
$M\in[0,1]^{1\times H\times W}$.
When a binary support is needed, we threshold $M$ with a fixed threshold; otherwise, the soft mask can be used directly.
Let $\mathcal{D}_{r}(\cdot)$ denote morphological dilation with radius $r$ on the latent grid, implemented by max pooling in practice, and let $\mathrm{clip}(\cdot,0,1)$ denote element-wise clipping.
We define the foreground, boundary, ring, and background supports as
\begin{equation}
M_{\mathrm{fg}}=\mathcal{D}_{r_{\mathrm{fg}}}(M),\quad
M_{\mathrm{bd}}=\mathcal{D}_{r_{\mathrm{bd}}}(M_{\mathrm{fg}}),\quad
M_{\mathrm{ring}}=\mathrm{clip}(M_{\mathrm{bd}}-M_{\mathrm{fg}},0,1),\quad
M_{\mathrm{bg}}=1-M_{\mathrm{fg}}.
\label{eq:app_regions}
\end{equation}
Here $M_{\mathrm{fg}}$ is the editable object interior, $M_{\mathrm{ring}}$ is a thin transition band around the object, and $M_{\mathrm{bg}}$ denotes the non-target region.
This construction intentionally avoids treating the object boundary as a brittle binary cut: moderate uncertainty near the mask boundary is absorbed into the ring support.
In Stage-1, the support used for foreground relaxation and background anchoring (Eqs.~\eqref{eq:fg_relax} and~\eqref{eq:stage1_safe}) is dilated by one additional latent cell, i.e., $\mathcal{D}_{1}(M_{\mathrm{fg}})$ and its complement, which keeps background anchoring slightly away from the object boundary; for brevity we use the same symbols $M_{\mathrm{fg}}$ and $M_{\mathrm{bg}}$ in both stages.

\paragraph{Velocity-level weight maps.}
The support maps are used only to gate velocity updates.
We define two weight maps, broadcast over channels:
\begin{equation}
W_t^{\mathrm{edit}} = M_{\mathrm{fg}},\qquad
W_t^{\mathrm{ref}} = M_{\mathrm{fg}} + \beta(\sigma_t)\,M_{\mathrm{ring}}.
\label{eq:app_weights}
\end{equation}
$W_t^{\mathrm{edit}}$ specifies where semantic injection is allowed in the final edited velocity.
$W_t^{\mathrm{ref}}$ is used for the moving reference anchor and additionally includes the boundary ring.
This lets the reference anchor transition smoothly near object boundaries while still keeping the non-target background protected from target-semantic accumulation.

\paragraph{Boundary coupling schedule.}
We use a fixed monotone decay schedule for the boundary coupling:
\begin{equation}
\beta(\sigma_t)=\beta_{\max}\Big(\frac{\sigma_t}{\sigma_{t_s}}\Big)^{\gamma},
\qquad \text{for } \sigma_t\le\sigma_{t_s}.
\label{eq:app_beta}
\end{equation}
Here $t_s$ is the editing start step, $\beta_{\max}$ controls the maximum ring coupling, and $\gamma$ controls the decay rate.
The intuition is simple: at earlier, noisier steps, the object boundary is less stable and benefits from soft coupling; at later steps, the boundary becomes more detailed, so the ring weight decays to avoid unnecessary leakage.
In all main experiments, these boundary parameters are fixed to $\beta_{\max}=0.45$ and $\gamma=2$ and are not tuned per image.

\paragraph{Robustness to imperfect supports.}
OAVC does not assume that the support prior is perfect.
Since the mask is not used for hard latent overwrite, moderate support errors do not directly force incorrect pixels into the final image.
Instead, the support only controls where velocity residuals are allowed to accumulate during integration.
Small under-segmentation errors are partially mitigated by foreground dilation, while boundary uncertainty is buffered by the ring support in $W_t^{\mathrm{ref}}$.
Moreover, the framework is not restricted to a single binary mask: in practice, the support can be corrected or replaced using soft thresholds, attention-derived support, full-mask fallback, or interactive refinement.

This also clarifies the role of SAM3.
If SAM3 produces a high-quality object support, OAVC uses it as the default zero-shot prior.
If the automatic support is imperfect, the same OAVC update can still be applied with a corrected or alternative support source.
Thus, the method is not a ``SAM3 editing pipeline''; it is a velocity-control framework that accepts a replaceable object support prior.
\begin{figure}[t]
    \centering
    \includegraphics[width=1.0\linewidth]{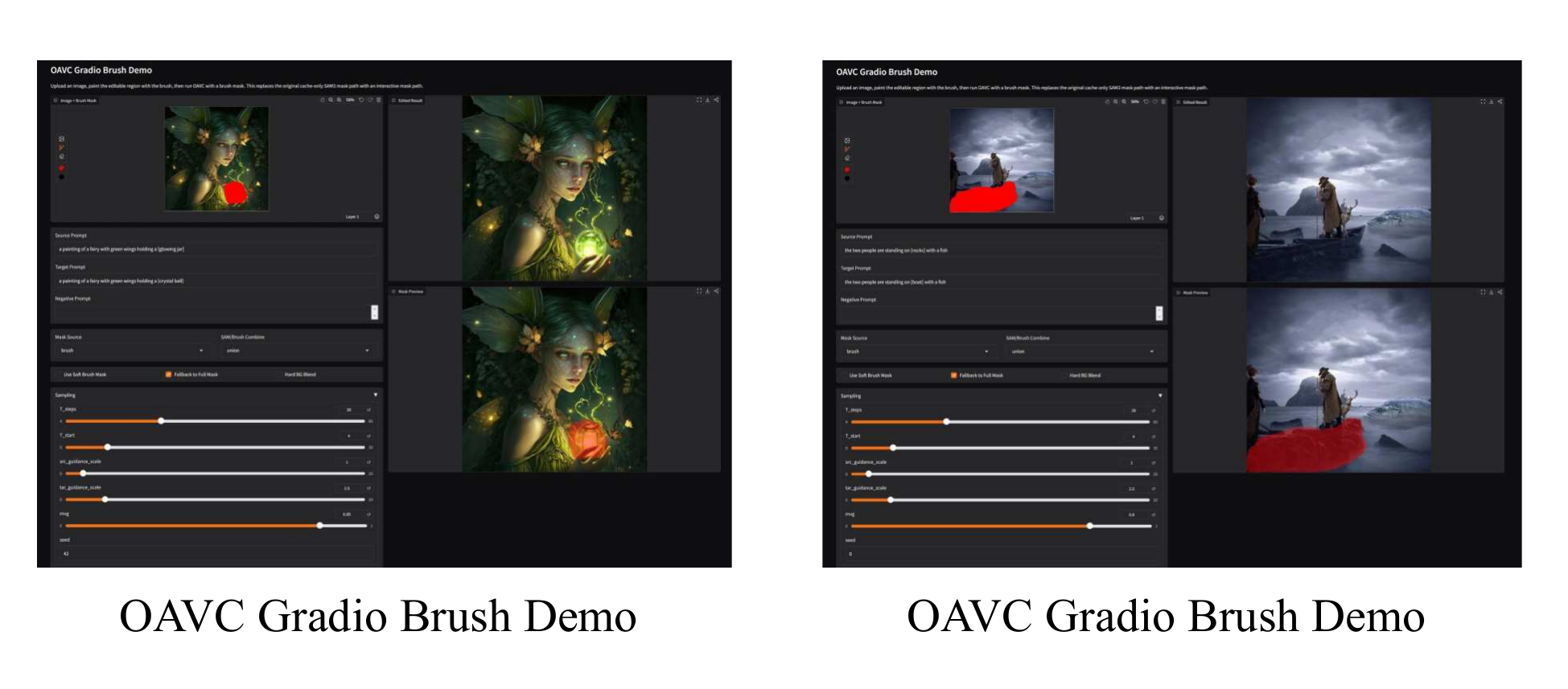}
    \caption{
    Interactive brush refinement for challenging support extraction cases.
    When the automatic SAM3 support is imperfect, users can manually refine the editable region with a brush, and OAVC applies the same velocity-control mechanism using the corrected support.
    }
    \label{fig:brush_refinement}
\end{figure}
\paragraph{Interactive refinement for hard cases.}
For challenging cases in which SAM3 fails to extract a reliable object support (e.g., thin structures, ambiguous boundaries, or visually complex regions), we additionally support interactive brush-based refinement.
Concretely, the user can manually mark the intended editing region and use the brush mask directly, or combine it with the automatic support through modes such as brush-only, SAM-only, or SAM+brush refinement.
This provides a simple and practical correction mechanism for failure cases without changing the OAVC formulation itself.
We include representative examples in Fig.~\ref{fig:brush_refinement}, where automatic support is insufficient but brush refinement recovers a usable object prior for object-centric velocity control.

\section{Support Source Robustness}
\label{sec:appendix_support_source}

This section further clarifies that OAVC is not tied to a specific support provider such as SAM3.
The method requires an explicit object support for velocity-level control, but the support can come from different sources.
In our default implementation, we use SAM3 as a strong zero-shot support provider for object-centric edits.
However, the same OAVC update can also use attention-derived supports, soft masks, user-provided masks, or interactively refined brush masks.
This supports our main claim that the contribution is object-aware velocity control rather than SAM-based editing.

\paragraph{Attention-derived support.}
To test whether an external segmenter is necessary, we replace the SAM3 support with an attention-derived support extracted from the editing backbone itself.
For FLUX, we follow the common strategy used in attention-based editing methods: we collect the attention response associated with the edited concept from a late double-attention block, average it across heads, normalize it to $[0,1]$, and threshold it to obtain a coarse editing region.
The resulting attention map is then expanded using the same dilation and boundary-ring construction described in Appendix~\ref{sec:appendix_object_prior}.
After this support is constructed, the rest of OAVC is unchanged.
Thus, this comparison isolates the support source while keeping the same velocity-control mechanism.

\paragraph{Quantitative comparison.}
Tab.~\ref{tab:attention_support} compares OAVC using attention-derived support with the default SAM3 support.
The attention-based variant remains substantially better than global DNAEdit on structure and background preservation metrics.
Compared with SAM3, it shows a moderate degradation, which is expected because attention maps are usually coarser and less boundary-accurate than segmentation masks.
Nevertheless, the result confirms that OAVC is not fundamentally dependent on SAM3: what the method needs is a usable support prior for per-step velocity control.

\begin{table}[t]
\centering
\small
\setlength{\tabcolsep}{5pt}
\caption{
Support-source comparison on PIE-Bench.
All metrics are scaled to match the main tables.
Attention-derived support can replace SAM3 and still provides strong preservation gains over global DNAEdit, showing that OAVC is support-source agnostic.
}
\label{tab:attention_support}
\resizebox{\linewidth}{!}{
\begin{tabular}{l|ccccc|cc}
\toprule
\multirow{2}{*}{Support source} &
\multicolumn{1}{c}{Struct.} &
\multicolumn{4}{c|}{Background Preservation} &
\multicolumn{2}{c}{CLIP Similarity} \\
\cmidrule(lr){2-2} \cmidrule(lr){3-6} \cmidrule(lr){7-8}
& Distance$\downarrow$ &
PSNR$\uparrow$ & LPIPS$\downarrow$ & MSE$\downarrow$ & SSIM$\uparrow$ &
CLIP$_\text{whole}\uparrow$ & CLIP$_\text{edit}\uparrow$ \\
\midrule
DNAEdit (global) 
& 18.87 & 24.99 & 95.06 & 50.45 & 85.71 & \textbf{25.79} & \textbf{22.87} \\
Attention-derived support 
& 4.58 & 30.55 & 35.74 & 16.41 & 92.53 & 23.60 & 20.76 \\
SAM3 support (default) 
& \textbf{4.07} & \textbf{33.30} & \textbf{23.70} & \textbf{8.75} & \textbf{94.38} & 23.70 & 21.08 \\
\bottomrule
\end{tabular}
}
\end{table}

\begin{figure}[t]
    \centering
    \includegraphics[width=\linewidth]{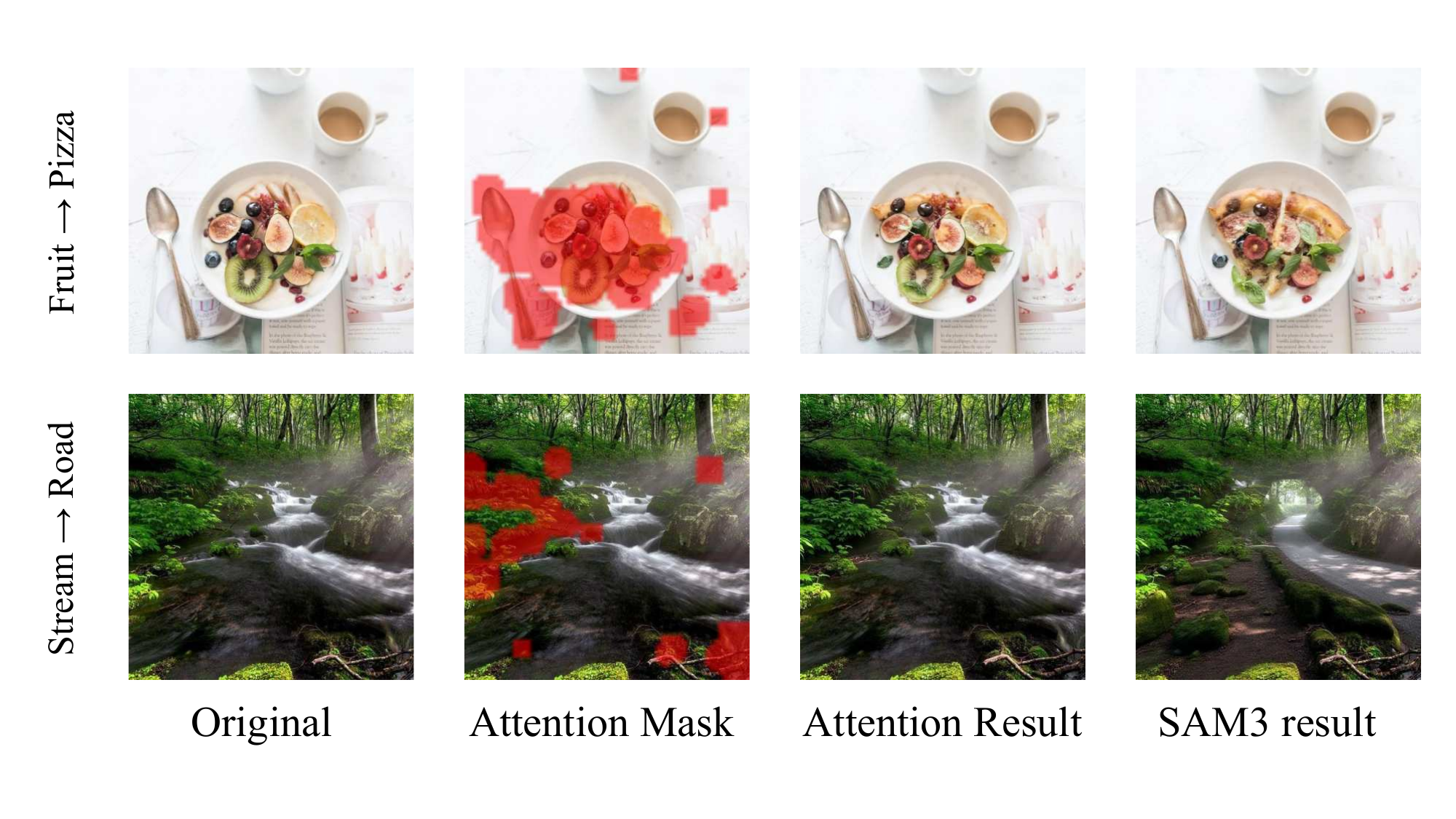}
    \caption{
    Qualitative examples using attention-derived support instead of SAM3.
    Attention maps provide an internal support prior for OAVC, enabling object-localized velocity control without an external segmenter.
    Although attention-derived supports are coarser than SAM3 masks, they remain sufficient to reduce background drift and preserve non-target regions compared with global steering.
    }
    \label{fig:attention_support_examples}
\end{figure}

\paragraph{Visualization.}
Fig.~\ref{fig:attention_support_examples} visualizes representative examples using attention-derived support.
Although the attention maps are coarser than SAM3 masks, they still provide meaningful object-centered regions for OAVC.
The resulting edits preserve the main advantage of velocity-level spatial control: target semantics are injected into the intended object region while non-target regions remain more stable than with global steering.

\paragraph{Discussion.}
The attention-derived support provides a fully internal alternative to segmentation-based support.
Its performance is weaker than SAM3 because attention maps are lower-resolution, more diffuse, and less reliable around fine boundaries.
However, the large gap between global DNAEdit and OAVC with attention-derived support shows that the benefit does not come from SAM3 itself.
Instead, the benefit comes from applying the same object-aware where/how control to a usable spatial support.

This observation also complements the interactive-refinement setting in Appendix~\ref{sec:appendix_object_prior}.
When automatic segmentation is accurate, SAM3 provides a strong zero-shot default.
When segmentation is ambiguous, users can refine the support with a brush.
When one wants to avoid an external segmenter, attention-derived support provides an internal alternative.
All three cases use the same OAVC update rule, confirming that OAVC is a support-source-agnostic velocity-control framework rather than a SAM3-specific editing pipeline.

\section{Full Derivations for the Two-Stage Recursions}
\label{sec:app_full_derivations}

This appendix provides a complete derivation of the two-stage recursions
used in our method.
Unlike the main text, which focuses on modeling intuition and algorithmic
structure, this appendix is dedicated to establishing the mathematical
origin and correctness of the update rules in Stage-1 and Stage-2.

Our goals are threefold:
(i) to show that the Stage-1 correction arises as the optimizer of a
regularized reference-trajectory alignment problem,
(ii) to explain why the resulting residual offsets form a natural and
well-conditioned interface to Stage-2, and
(iii) to demonstrate that the projection and constrained updates in
Stage-2 correspond to optimal solutions of constrained control problems.

Throughout, we follow the rectified-flow sampling formulation used in DNAEdit.
The reverse process is discretized into $T$ solver steps indexed by $t\in\{0,\ldots,T-1\}$.
The scheduler provides a monotonically ordered sequence of integration levels
$\{\sigma_0,\ldots,\sigma_T\}\subset[0,1]$, which we use only to construct evaluable proxy/interpolation states
(e.g., $Z_t^{*}$ in the main text).
Independently, the numerical integrator uses a per-step update magnitude (effective step size),
which we denote abstractly by $\Delta_t>0$ for each step $t$.
We do not assume a specific analytic relationship between $\Delta_t$ and $\{\sigma_t\}$, since $\Delta_t$ depends on the
chosen solver implementation and backbone-specific parameterization; in our derivations, $\Delta_t$ is simply the scalar
that multiplies the applied velocity in the timestep update.

\subsection{Setup: Noise-Parameterized Discrete Dynamics}
\label{sec:app_setup}

Let $Z_t\in\mathbb{R}^{C\times H\times W}$ denote the latent variable at noise
level $\sigma_t$.
A reverse-time Euler update takes the form
\begin{equation}
Z_t
=
Z_{t+1} + \Delta\sigma_t\,\widehat v_t,
\qquad
\Delta\sigma_t=\sigma_{t+1}-\sigma_t>0,
\label{eq:app_euler}
\end{equation}
where $\widehat v_t$ is the velocity applied at step $t$.

Given a text prompt $\psi$, the pretrained model predicts a velocity field
$v_\psi(Z)=v_\theta(Z,\psi)$.
We denote by
$v_t^{\mathrm{src}}$ and $v_t^{\mathrm{tgt}}$
the velocities induced by the source and target prompts, respectively.

\paragraph{Recall: spatial priors.}
As defined in the main text, an object prior induces spatial masks
$M_{\mathrm{fg}}$, $M_{\mathrm{ring}}$, and $M_{\mathrm{bg}}$,
as well as two associated spatial weights
$W_t^{\mathrm{edit}}$ and $W_t^{\mathrm{ref}}$.
In this appendix, these masks are treated as fixed and we focus solely on
how they modulate the velocity updates.

\subsection{DNA Discrepancy from Linear Transport}
\label{sec:app_dna_discrepancy}

DNAEdit constructs a source-consistent reference trajectory by comparing the
model-predicted source velocity against an analytic linear-transport
velocity induced by the sampler's interpolation rule.

At each step $t$, the sampler defines an interpolation point $Z_t^{*}$ on a
linear path in noise space.
The corresponding linear velocity $v_t^{\mathrm{linear}}$ is the unique
velocity that maps $Z_{t+1}$ to $Z_t^{*}$ under this parameterization.
Evaluating the model under the source prompt yields
$v_t^{\mathrm{src}}=v_\theta(Z_t^{*},\psi_{\mathrm{src}})$.
The DNA discrepancy is then defined as
\begin{equation}
\Delta v_t^{\mathrm{DNA}}
\;\triangleq\;
v_t^{\mathrm{linear}} - v_t^{\mathrm{src}}.
\label{eq:app_dna_residual}
\end{equation}
This discrepancy measures the deviation between the learned velocity field
and the idealized linear transport implied by rectified flow.

% ------------------------------------------------------------
\subsection{Additional Preliminaries: Control View and Spatial Support}
\label{sec:app_prelim_control}

This subsection complements the concise preliminaries in the main text by (i) making explicit the
trajectory-integration interpretation of velocity control, and (ii) formalizing why \emph{where} a
perturbation acts (spatial support) is as important as \emph{how} strongly it acts (temporal strength),
especially for object-centric edits.

\paragraph{Discrete accumulation and drift.}
Training-free editors intervene by adding a step-wise control signal $u_t$ to the sampler update.
For a discrete trajectory indexed by timesteps $t\in\{0,\ldots,T-1\}$, the reverse update can be written as
\begin{equation}
Z_{t+1} = Z_t + \Delta_t\,\widehat v_t,
\qquad
\widehat v_t = v_\theta(Z_t,\psi) + u_t,
\label{eq:app_discrete_control_update}
\end{equation}
where $\Delta_t$ denotes the step size implied by the scheduler (e.g., a function of adjacent integration levels).
Unrolling the recursion shows that the final sample aggregates the injected control over all steps:
\begin{equation}
Z_T = Z_0 + \sum_{t=0}^{T-1}\Delta_t\, v_\theta(Z_t,\psi)
\;+\;
\sum_{t=0}^{T-1}\Delta_t\, u_t.
\label{eq:app_discrete_accum}
\end{equation}
The second term makes explicit why even small, persistent perturbations can accumulate into visible artifacts.

Let $\Omega$ denote the spatial domain and let $\Omega_{\mathrm{bg}}\subset\Omega$ be background locations that should remain unchanged.
If the control is applied globally, i.e., $u_t(x)\neq 0$ on $x\in\Omega_{\mathrm{bg}}$ for many steps, then the accumulated background change satisfies
\begin{equation}
\Delta Z_{\mathrm{bg}}
\;\triangleq\;
\sum_{t=0}^{T-1}\Delta_t\, u_t\big|_{\Omega_{\mathrm{bg}}},
\label{eq:app_bg_drift_discrete}
\end{equation}
which can become non-negligible even when each per-step background component is small.
This motivates explicitly controlling the \emph{spatial support} of velocity injection to prevent repeated background contamination across timesteps.

\paragraph{Decoupling \emph{where} and \emph{how}.}
Many velocity-steering methods implicitly control only the \emph{strength} of a global semantic difference, e.g.,
$u_t=\alpha_t\,\Delta v_t$ with a scalar schedule $\alpha_t$.
In contrast, we make the control structure explicit by separating \emph{where} the update is allowed to act from \emph{how} the semantic difference is transformed into a safe injection direction:
\begin{equation}
u_t
=
W_t \odot \Gamma_t(\Delta v_t;\mathcal{R}_t).
\label{eq:app_where_how}
\end{equation}
Here $W_t\in[0,1]^{1\times H\times W}$ is a spatial support (broadcast across channels) that gates updates to the intended object region (Appendix~\ref{sec:appendix_object_prior}).
$\Gamma_t(\cdot)$ is a constrained operator that removes drift-inducing components (e.g., by projection, boundary handling, and gating), and $\mathcal{R}_t$ denotes the Stage-1 reference information ensuring that $\Delta v_t$ is evaluated on consistent states.
Eq.~\eqref{eq:app_where_how} directly prevents step-by-step perturbations from repeatedly affecting background locations while preserving strong semantic steering within the object support.

\subsection{Notation and Operators}
\label{sec:app_notation}

\paragraph{Latents and timesteps.}
We discretize the reverse trajectory into $T$ timesteps $t\in\{0,\ldots,T-1\}$.
Let $Z_t\in\mathbb{R}^{C\times H\times W}$ denote the latent at timestep $t$.
The scheduler induces a per-step integration increment, denoted by $\Delta_t>0$, so a generic Euler-type update takes the form
\begin{equation}
Z_{t+1} = Z_t + \Delta_t\,\widehat v_t.
\label{eq:app_euler_timestep}
\end{equation}
We use $S_t$ to denote the auxiliary Gaussian noise state aligned with timestep $t$ (its recursion is defined in Appendix~\ref{sec:app_stage1_recursion}).

\paragraph{Pointwise and spatial inner products.}
For two latent-shaped tensors $a,b\in\mathbb{R}^{C\times H\times W}$ and a spatial location $x\in\Omega$,
\begin{equation}
a(x)\cdot b(x) \;\triangleq\; \sum_{c=1}^{C} a_c(x)\,b_c(x).
\end{equation}
We define the spatial-sum inner product
\begin{equation}
\langle a,b\rangle \;\triangleq\; \sum_{x\in\Omega} a(x)\cdot b(x),
\end{equation}
and the background-weighted inner product (used in Stage-1)
\begin{equation}
\langle a,b\rangle_{\mathrm{bg}}
\;\triangleq\;
\langle M_{\mathrm{bg}}\odot a,\; M_{\mathrm{bg}}\odot b\rangle.
\label{eq:app_bg_inner}
\end{equation}
Here $\odot$ denotes element-wise multiplication; spatial masks are broadcast across channels.

\paragraph{Projection operators.}
Given a nonzero reference field $v$ and a tensor $r$,
the projection of $r$ onto $v$ under $\langle\cdot,\cdot\rangle$ is
\begin{equation}
\mathrm{Proj}_{v}(r)
=
\frac{\langle r,v\rangle}{\langle v,v\rangle+\epsilon}\,v,
\end{equation}
and the orthogonal component is
\begin{equation}
\mathrm{Proj}^{\perp}_{v}(r)
=
r-\mathrm{Proj}_{v}(r).
\label{eq:app_proj_perp}
\end{equation}
For Stage-1 background anchoring, we use the same form but with
$\langle\cdot,\cdot\rangle_{\mathrm{bg}}$ (Eq.~\eqref{eq:app_bg_inner}) and with $v$ masked by $M_{\mathrm{bg}}$.

% ------------------------------------------------------------
\section{Stage-1 Derivation: Regularized reference-trajectory Alignment}
\label{sec:app_stage1_derivation}

Stage-1 follows the DNAEdit recursion backbone but replaces the raw discrepancy
$\Delta v_t^{\mathrm{DNA}}$ with a modified correction $\Delta v_t^{\mathrm{safe}}$ used in the Stage-1 state updates.
This appendix provides a principled interpretation of the two Stage-1 modifications in the main text:
(i) \emph{foreground weakening}, which reduces the alignment force inside the editable object region to preserve
degrees of freedom for Stage-2 semantic injection; and
(ii) \emph{background anchoring}, which suppresses drift-inducing background components aligned with the source flow.
For (i), we first derive the exact closed-form shrinkage implied by a diagonal Tikhonov (ridge) regularizer,
and then explain how the lightweight linear shrinkage used in the main text (Eq.~\eqref{eq:app_linear_shrink})
approximates this exact shrinkage while preserving the same masking structure.

\subsection{Foreground Weakening via Diagonal Tikhonov Regularization}
\label{sec:app_stage1_ridge}

\paragraph{What is being optimized.}
At timestep $t$, DNAEdit defines the alignment discrepancy
$\Delta v_t^{\mathrm{DNA}} \triangleq v_t^{\mathrm{linear}} - v_t^{\mathrm{src}}$ (see Eq.~\eqref{eq:app_dna_residual}),
which serves as a step-wise \emph{alignment correction velocity}.
Stage-1 seeks a modified correction $\delta v$ that remains close to $\Delta v_t^{\mathrm{DNA}}$ (so that source alignment is retained),
but is \emph{attenuated on the editable foreground} so that Stage-2 can inject semantics without fighting an overly rigid foreground anchor.
This motivates a regularized objective over the correction variable $\delta v$.

\paragraph{Why naive weighting is insufficient.}
A naive weighted least-squares term $\min \|w\odot(\delta v-\Delta v)\|_2^2$ does not change the minimizer when $w>0$ elementwise,
because it only rescales the same residual.
To actually shrink the correction magnitude on selected regions, we use a diagonal Tikhonov penalty on $\delta v$.

\paragraph{Exact Tikhonov shrinkage (closed form).}
Consider the diagonal Tikhonov-regularized objective
\begin{equation}
\Delta v_t^{(1)\star}
=
\arg\min_{\delta v}\;
\|\delta v-\Delta v_t^{\mathrm{DNA}}\|_2^2
+
\|\Gamma^{1/2}\odot\delta v\|_2^2,
\label{eq:app_tikh}
\end{equation}
where $\Gamma\succeq 0$ is diagonal and $\odot$ denotes element-wise multiplication (masks are broadcast over channels).
Choosing $\Gamma=\lambda_{\mathrm{fg}}M_{\mathrm{fg}}$ penalizes correction energy only on the foreground region
$M_{\mathrm{fg}}\in\{0,1\}^{H\times W}$.
Since the objective is separable over spatial locations (and channels), the optimizer is a pointwise shrinkage:
\begin{equation}
\Delta v_t^{(1)\star}
=
\frac{1}{1+\Gamma}\odot \Delta v_t^{\mathrm{DNA}}
=
\frac{1}{1+\lambda_{\mathrm{fg}}M_{\mathrm{fg}}}\odot \Delta v_t^{\mathrm{DNA}}.
\label{eq:app_tikh_sol}
\end{equation}
Eq.~\eqref{eq:app_tikh_sol} makes the effect explicit:
on background pixels ($M_{\mathrm{fg}}=0$), $\Delta v_t^{(1)\star}=\Delta v_t^{\mathrm{DNA}}$;
on foreground pixels ($M_{\mathrm{fg}}=1$), the correction is shrunk by a factor $\frac{1}{1+\lambda_{\mathrm{fg}}}$.

\paragraph{Connection to the main-text implementation.}
In the main text, we implement foreground weakening using the linear shrinkage
\begin{equation}
\Delta v_t^{(1)}
=
(1-\lambda_{\mathrm{fg}}M_{\mathrm{fg}})\odot \Delta v_t^{\mathrm{DNA}},
\label{eq:app_linear_shrink}
\end{equation}
which is the quantity used to form the background residual $r_t=M_{\mathrm{bg}}\odot\Delta v_t^{(1)}$
and ultimately the Stage-1 safe correction $\Delta v_t^{\mathrm{safe}}$ (Eq.~\eqref{eq:app_stage1_safe}).
Eq.~\eqref{eq:app_linear_shrink} preserves the same masking structure as the exact shrinkage and can be interpreted as a
first-order approximation of Eq.~\eqref{eq:app_tikh_sol}:
\begin{equation}
\frac{1}{1+\lambda_{\mathrm{fg}}M_{\mathrm{fg}}}
=
1-\lambda_{\mathrm{fg}}M_{\mathrm{fg}}
+ \mathcal{O}(\lambda_{\mathrm{fg}}^2),
\end{equation}
so when $\lambda_{\mathrm{fg}}$ is small, $(1-\lambda_{\mathrm{fg}}M_{\mathrm{fg}})$ closely matches the exact Tikhonov shrink factor.
Moreover, restricting $\lambda_{\mathrm{fg}}\in[0,1]$ (as in the main text) guarantees a nonnegative shrink factor and yields a
simple monotone knob controlling the foreground freedom in practice.

\subsection{Background Anchoring as Soft Parallel-Component Removal}
\label{sec:app_stage1_proj}

Even after foreground weakening, the \emph{background} portion of $\Delta v_t^{(1)}$ may still contain components
that are \emph{parallel} to the source flow $v_t^{\mathrm{src}}$.
Such parallel components typically correspond to low-frequency, transport-like structural motion; when integrated over many steps,
they accumulate into visible background drift.
Stage-1 therefore \emph{softly} suppresses only the background component aligned with the source velocity, using the same
parameter $\rho_{\mathrm{bg}}\in[0,1]$ as in the main text.

\paragraph{Background residual and source direction (aligned with the main text).}
We isolate the background part of the weakened discrepancy and the background source-flow direction:
\begin{equation}
r_t \;=\; M_{\mathrm{bg}}\odot \Delta v_t^{(1)},\qquad
b_t \;=\; M_{\mathrm{bg}}\odot v_t^{\mathrm{src}} .
\label{eq:app_bg_defs}
\end{equation}
Here $r_t$ is the background alignment residual we would otherwise integrate, and $b_t$ specifies the background source-flow direction
along which drift tends to accumulate.

\paragraph{Soft removal of the parallel component.}
We measure the parallel component of $r_t$ along $b_t$ via the standard projection coefficient
\begin{equation}
\alpha_t \;=\; \frac{\langle r_t,b_t\rangle}{\langle b_t,b_t\rangle+\epsilon},
\label{eq:app_alpha}
\end{equation}
so that $\alpha_t b_t$ is exactly the projection of $r_t$ onto $b_t$, and the hard orthogonal projection would be
$r_t^\perp=r_t-\alpha_t b_t$.
Instead of enforcing full orthogonality, Stage-1 removes only a \emph{fraction} $\rho_{\mathrm{bg}}\in[0,1]$ of this parallel component:
\begin{equation}
\tilde r_t
\;=\;
r_t - \rho_{\mathrm{bg}}\,\alpha_t\, b_t
\;=\;
(1-\rho_{\mathrm{bg}})\,r_t + \rho_{\mathrm{bg}}\,r_t^\perp .
\label{eq:app_bg_soft_remove}
\end{equation}
It has a clear interpretation: $\rho_{\mathrm{bg}}=0$ keeps the original background residual ($\tilde r_t=r_t$),
while $\rho_{\mathrm{bg}}=1$ recovers the hard orthogonal projection ($\tilde r_t=r_t^\perp$).

\paragraph{Assembling the Stage-1 safe correction.}
Finally, Stage-1 combines the (weakened) foreground correction with the softly anchored background correction:
\begin{equation}
\Delta v_t^{\mathrm{safe}}
=
(1-M_{\mathrm{bg}})\odot \Delta v_t^{(1)}
+
\tilde r_t,
\qquad
\tilde r_t = r_t - \rho_{\mathrm{bg}}\,\alpha_t\, b_t .
\label{eq:app_stage1_safe}
\end{equation}
Eq.~\eqref{eq:app_stage1_safe} is identical to the Stage-1 construction in the main text
(Eq.~\eqref{eq:stage1_safe}) and makes explicit that Stage-1 anchors the background by partially
removing only the drift-inducing component parallel to the source flow, while leaving the remaining (approximately orthogonal)
background correction intact.

\section{Stage-1 Recursion and Stored Offsets}
\label{sec:app_stage1_recursion}

This section instantiates the Stage-1 recursions in the timestep-based notation of
Appendix~\ref{sec:app_notation}.
At each timestep $t$, the scheduler provides (i) an evaluable interpolation/proxy point $Z_t^{*}$
(as defined in the main text / preliminaries) and (ii) a per-step integration increment $\Delta_t>0$
so that a generic Euler-type update reads $Z_{t+1}=Z_t+\Delta_t\,\widehat v_t$ (Eq.~\eqref{eq:app_euler_timestep}).
Stage-1 applies the derived safe alignment correction $\Delta v_t^{\mathrm{safe}}$ as an additive correction at timestep $t$.

\paragraph{Latent recursion in Stage-1.}
Using $\Delta v_t^{\mathrm{safe}}$, Stage-1 forms a corrected latent by adding the step-size--scaled correction to
the proxy point:
\begin{equation}
Z_t \;=\; Z_t^{*} + \Delta_t\,\Delta v_t^{\mathrm{safe}} .
\label{eq:app_stage1_latent_update}
\end{equation}

\paragraph{Auxiliary noise-state recursion.}
Following DNAEdit's noise-space bookkeeping, we maintain an auxiliary noise state $S_t$ aligned with timestep $t$.
Intuitively, $S_t$ tracks the noise endpoint of the linear interpolation used to construct $Z_t^{*}$ and is updated
using the same safe correction.
Consistent with the Stage-1 update used in the main text, we update
\begin{equation}
S_t \;=\; S_{t+1} + \sigma_{t+1}\,\Delta v_t^{\mathrm{safe}} ,
\label{eq:app_stage1_noise_update}
\end{equation}
where $\sigma_{t+1}$ is the scheduler-provided noise level associated with timestep $t+1$
(the same quantity used to construct $Z_t^{*}$ in the main text).
Eq.~\eqref{eq:app_stage1_noise_update} is the exact recursion implemented in Stage-1; it ensures that the proxy-point construction
and the stored offsets remain consistent across steps.

\paragraph{Stored offset (interface to Stage-2).}
We cache the per-timestep offset between the corrected latent and the proxy point:
\begin{equation}
\Delta x_t^{\mathrm{safe}}
\;\triangleq\;
Z_t - Z_t^{*}
=
\Delta_t\,\Delta v_t^{\mathrm{safe}} .
\label{eq:app_dx}
\end{equation}
Because Euler-type discretizations are additive in the applied velocity (Eq.~\eqref{eq:app_euler_timestep}),
$\Delta x_t^{\mathrm{safe}}$ fully captures the step-wise contribution of the Stage-1 alignment correction.
Stage-2 therefore reuses $\Delta x_t^{\mathrm{safe}}$ to evaluate prompt-conditioned velocities on offset-aligned proxy states
without re-running Stage-1, making it the sufficient interface from Stage-1 to Stage-2.

\section{Stage-2 Derivation: Offset-Aligned and Constrained Semantic Injection}
\label{sec:app_stage2_derivation}

Stage-2 injects target semantics by (i) evaluating prompt-conditioned velocities on an offset-aligned proxy state,
and (ii) applying constrained, spatially supported updates that suppress target-induced background residuals at each step.
All symbols and indices in this section follow the main text and the timestep-based update in
Eq.~\eqref{eq:app_euler_timestep}.

\subsection{Offset-Aligned Evaluation and Stability}
\label{sec:app_stage2_offset}

\paragraph{Offset-aligned proxy state (aligned with the main text).}
At timestep $t$, the edited trajectory provides the current latent $Z_t^{\mathrm{edit}}$.
To evaluate prompt-conditioned velocities consistently with Stage-1, we reuse the cached Stage-1 offset
$\Delta x_t^{\mathrm{safe}}$ and form the offset-aligned proxy state
\begin{equation}
Z_t^{*\mathrm{edit}}
=
Z_t^{\mathrm{edit}} + \Delta x_t^{\mathrm{safe}},
\qquad
v_t^{\mathrm{tgt}}=v_\theta(Z_t^{*\mathrm{edit}},\psi_{\mathrm{tgt}}).
\label{eq:app_offset}
\end{equation}
We also reuse the cached source velocity $v_t^{\mathrm{src}}=v_\theta(Z_t^{*},\psi_{\mathrm{src}})$ from Stage-1
to define the prompt-induced semantic displacement
$\Delta v_t = v_t^{\mathrm{tgt}}-v_t^{\mathrm{src}}$ (as in the main text).

\paragraph{Why controlling $\Delta x_t^{\mathrm{safe}}$ stabilizes semantic differences.}
Assume $v_\theta(\cdot,\psi)$ is locally Lipschitz around $Z_t^{\mathrm{edit}}$ for both $\psi_{\mathrm{src}}$ and $\psi_{\mathrm{tgt}}$.
Then the semantic difference evaluated at the offset-aligned state is stable w.r.t. the offset magnitude:
\begin{equation}
\|\Delta v_t(Z_t^{\mathrm{edit}}+\Delta x_t^{\mathrm{safe}})-\Delta v_t(Z_t^{\mathrm{edit}})\|
\le
(L_{\mathrm{tgt},t}+L_{\mathrm{src},t})\|\Delta x_t^{\mathrm{safe}}\|.
\end{equation}
Thus, Stage-1's suppression of spurious (especially background) components in $\Delta x_t^{\mathrm{safe}}$
directly improves the stability of $\Delta v_t$ used in Stage-2.

\subsection{First Projection: Transport-Orthogonal Semantic Direction}
\label{sec:app_stage2_proj}

semantic direction by projecting the \emph{prompt difference} $\Delta v_t$ onto the subspace orthogonal to $v_t^{\mathrm{src}}$:
\begin{equation}
\Delta v_t^\perp
=
\Delta v_t
-
\frac{\langle \Delta v_t,v_t^{\mathrm{src}}\rangle}
{\langle v_t^{\mathrm{src}},v_t^{\mathrm{src}}\rangle+\epsilon}
\,v_t^{\mathrm{src}}.
\label{eq:app_stage2_proj}
\end{equation}
Eq.~\eqref{eq:app_stage2_proj} is the unique minimizer of the constrained least-squares problem
$\min_u \|u-\Delta v_t\|_2^2\ \text{s.t.}\ \langle u,v_t^{\mathrm{src}}\rangle=0$,
and provides the semantic direction used for controlled accumulation (e.g., the reference update).

\subsection{Boundary High-Pass Operator and Confidence Gate}
\label{sec:appendix_gate}

\paragraph{High-pass operator on the boundary ring.}
Even when semantic injection is restricted to the object support, slow and low-frequency motion near the boundary may accumulate across steps,
leading to halos or contour wobble.
We optionally apply a lightweight high-pass operator only on the boundary ring to suppress such low-frequency drift while preserving local details.

For spatial latents (SD3.5, $x\in\mathbb{R}^{B\times C\times H\times W}$), we use a residual high-pass filter
\begin{equation}
\mathrm{HP}_{\mathrm{sp}}(x)\;\triangleq\; x - \mathrm{AvgPool}_{k}(x),
\label{eq:app_hp_spatial}
\end{equation}
where $\mathrm{AvgPool}_{k}(\cdot)$ is the local averaging operator with window size $k\times k$.

For packed token latents (FLUX, $x\in\mathbb{R}^{B\times L\times D}$), we use a token-wise analogue:
\begin{equation}
\mathrm{HP}_{\mathrm{tok}}(x)\;\triangleq\; x - \mathrm{Mean}_{\text{tokens}}(x),
\label{eq:app_hp_token}
\end{equation}
where $\mathrm{Mean}_{\text{tokens}}(\cdot)$ averages over the token dimension $L$.

\paragraph{Scalar confidence gate.}
Projected residuals can be unreliable at large noise levels, and repeatedly integrating such outliers may amplify artifacts.
We optionally apply a scalar confidence gate $g_t\in[0,1]$ to downweight residual updates whose magnitude is abnormally large.

Let $x_t$ denote the residual to be integrated at step $t$ (after projection and, if enabled, boundary high-pass filtering).
We compute its RMS magnitude
\begin{equation}
m_t \;\triangleq\; \sqrt{\mathrm{Mean}\!\left(x_t^2\right)},
\label{eq:app_gate_rms}
\end{equation}
maintain an EMA $\bar m_t \leftarrow \mu\,\bar m_{t-1} + (1-\mu)\, m_t$ with decay $\mu$,
form the ratio $r_t = m_t/(\bar m_t+\epsilon)$,
and map it to a smooth gate
\begin{equation}
g_t \;\triangleq\; \sigma\!\left(\frac{\kappa - r_t}{\tau}\right).
\label{eq:app_gate_sigmoid}
\end{equation}
In all experiments, the gate and high-pass settings are fixed and shared across backbones: $\mu=0.95$, $\kappa=2$, $\tau=0.5$, and window size $k=3$ for $\mathrm{HP}_{\mathrm{sp}}$ in Eq.~\eqref{eq:app_hp_spatial}.

\section{Stage-2 Recursion: Reference Constraint and Field-Level Projection}
\label{sec:app_stage2_recursion}

Stage-2 maintains an editing trajectory $Z_t^{\mathrm{edit}}$ and a reference trajectory $Z_t^{\mathrm{ref}}$ (as defined in the main text).
The reference trajectory accumulates controlled semantic displacement within $W_t^{\mathrm{ref}}$,
while the edited trajectory integrates a field that equals $v_t^{\mathrm{src}}$ on the background.

\paragraph{Why a second projection is applied (aligned with the main text).}
Eq.~\eqref{eq:app_stage2_proj} projects the prompt difference $\Delta v_t$ to obtain a transport-orthogonal semantic direction
used for controlled accumulation (e.g., the reference update).
However, the final integrated field is built from the reference-guided velocity $v_t^{\mathrm{fg}}$,
which is an intermediate guidance signal and is not guaranteed to be transport-orthogonal.
To suppress target-induced background residuals at each step, we therefore apply projection again at the field level and inject only the
source-orthogonal component of $(v_t^{\mathrm{fg}}-v_t^{\mathrm{src}})$ inside the object region.

\paragraph{Final edited field and timestep update.}
The final edited velocity field matches the main text:
\begin{equation}
v_t^{\mathrm{edit}}
=
v_t^{\mathrm{src}}
+
W_t^{\mathrm{edit}}\odot
\Big(
g_t\,\mathrm{Proj}^\perp_{v_t^{\mathrm{src}}}
\big(v_t^{\mathrm{fg}}-v_t^{\mathrm{src}}\big)
\Big).
\label{eq:app_final_edit_field}
\end{equation}
The edited latent is then updated by the timestep-based Euler step (Eq.~\eqref{eq:app_euler_timestep}):
\begin{equation}
Z_{t+1}^{\mathrm{edit}} = Z_t^{\mathrm{edit}} + \Delta_t\, v_t^{\mathrm{edit}}.
\label{eq:app_stage2_euler}
\end{equation}

\begin{figure}[!t]
    \centering
    \includegraphics[width=1.0\linewidth]{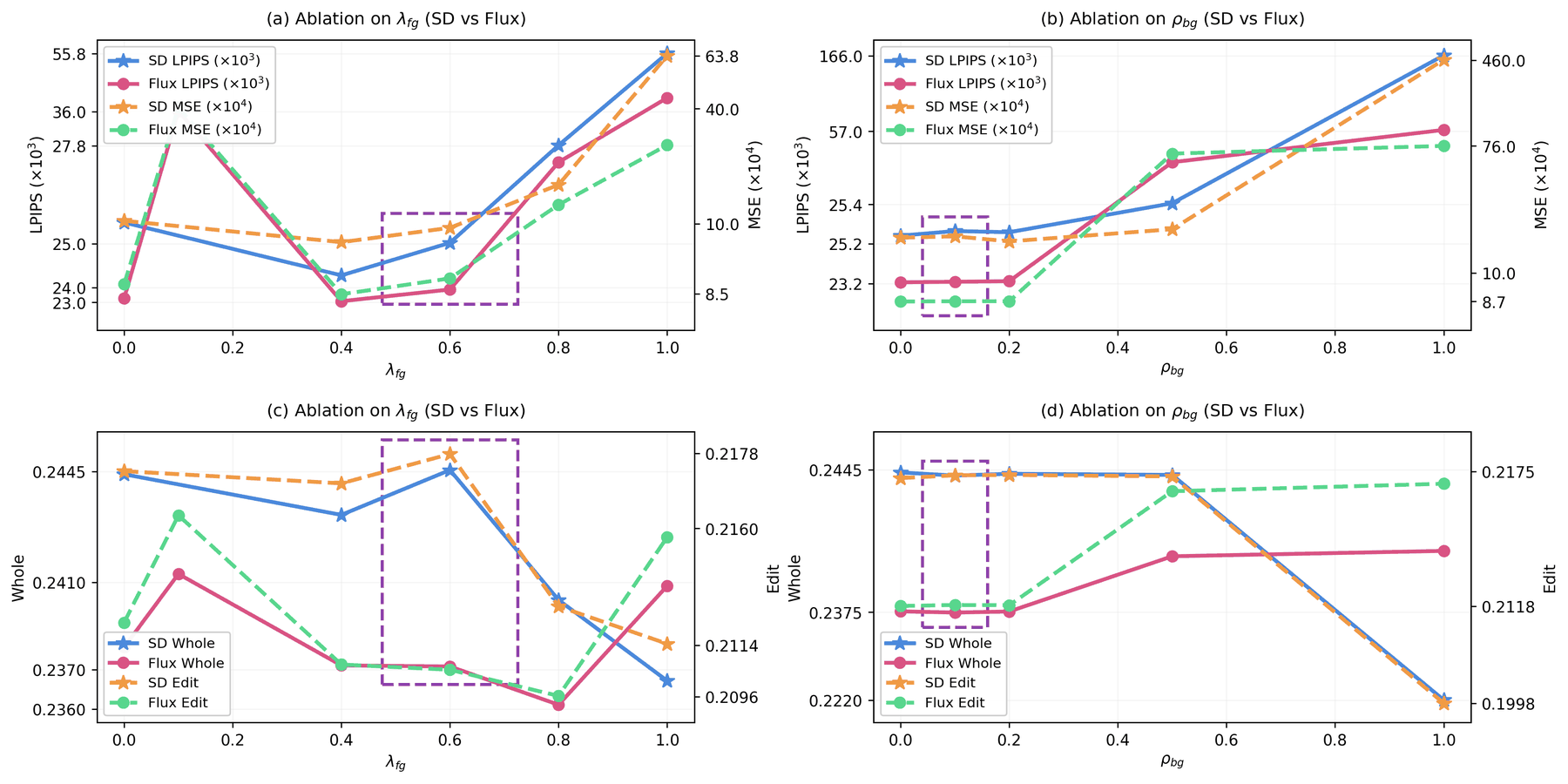}
    
    \caption{Stage-1 hyperparameter sweep on PIE-Bench for SD3.5 and FLUX.
    We vary foreground relaxation $\lambda_{fg}$ and background suppression $\rho_{bg}$ and report background preservation
    (LPIPS$\downarrow$, MSE$\downarrow$ on non-target regions; scaled as shown) together with semantic alignment
    (CLIP$_\text{whole}$/CLIP$_\text{edit}$).
    The dashed boxes highlight the stable region around our default setting ($\lambda_{fg}{=}0.6$, $\rho_{bg}{=}0.1$).}
    \label{fig:fg_bg}
\end{figure}
% =========================================================
\section{Additional Studies}
\label{sec:ablation}

This appendix reports additional studies.
All results follow the same setup and evaluation protocol as Sec.~\ref{sec:exp_setup} and Tab.~\ref{tab:pie_full}.

% =========================================================
% Table 4: fg_dilate (scaled to match Table 1)
% =========================================================
\begin{table}[t]
\caption{Ablation on the interior-support dilation radius $r_{\mathrm{fg}}$ on PIE-Bench (scaled to match Tab.~\ref{tab:pie_full}). The radius controls a preservation--editability trade-off; $r_{\mathrm{fg}}=2$ is the default operating point used in all main experiments.}
\label{tab:ablation_fg_dilate}
\centering
\small
\setlength{\tabcolsep}{4pt}
\resizebox{\linewidth}{!}{
\begin{tabular}{l|c|c|c|c|c|c|c}
\toprule
\multirow{2}{*}{Setting} &
\multicolumn{1}{c|}{\textbf{Structure}} &
\multicolumn{4}{c|}{\textbf{Background Preservation}} &
\multicolumn{2}{c}{\textbf{CLIP Similarity$\uparrow$}} \\
\cmidrule(lr){2-2} \cmidrule(lr){3-6} \cmidrule(lr){7-8}
& Distance$\downarrow$ &
PSNR$\uparrow$ & LPIPS$\downarrow$ & MSE$\downarrow$ & SSIM$\uparrow$ &
CLIP$_\text{whole}\uparrow$ & CLIP$_\text{edit}\uparrow$ \\
\midrule
$r_{\mathrm{fg}}{=}3$ 
& 4.42 & 32.22 & 25.60 & 10.26 & 92.86 & \textbf{24.29} & 21.50 \\
$r_{\mathrm{fg}}{=}1$ 
& \textbf{3.15} & \textbf{34.10} & \textbf{21.68} & \textbf{7.86} & \textbf{93.35} & 23.83 & 21.25 \\
$r_{\mathrm{fg}}{=}0$ 
& 3.80 & 33.26 & 23.05 & 8.78 & 93.17 & 24.10 & 21.45 \\
$r_{\mathrm{fg}}{=}2$ (default) 
& 4.11 & 32.64 & 24.40 & 9.61 & 93.00 & 24.21 & \textbf{21.51} \\
\bottomrule
\end{tabular}
}
\end{table}

% =========================================================
% Table 5: beta_max (scaled to match Table 1)
% =========================================================
\begin{table}[t]
\caption{Ablation on maximum ring coupling $\beta_{\max}$ in $\beta(\sigma_t)$ (Eq.~\eqref{eq:app_beta}) on PIE-Bench (scaled to match Tab.~\ref{tab:pie_full}).}
\label{tab:ablation_beta_max}
\centering
\small
\setlength{\tabcolsep}{4pt}
\resizebox{\linewidth}{!}{
\begin{tabular}{l|c|c|c|c|c|c|c}
\toprule
\multirow{2}{*}{Setting} &
\multicolumn{1}{c|}{\textbf{Structure}} &
\multicolumn{4}{c|}{\textbf{Background Preservation}} &
\multicolumn{2}{c}{\textbf{CLIP Similarity$\uparrow$}} \\
\cmidrule(lr){2-2} \cmidrule(lr){3-6} \cmidrule(lr){7-8}
& Distance$\downarrow$ &
PSNR$\uparrow$ & LPIPS$\downarrow$ & MSE$\downarrow$ & SSIM$\uparrow$ &
CLIP$_\text{whole}\uparrow$ & CLIP$_\text{edit}\uparrow$ \\
\midrule
$\beta_{\max}{=}0.60$ & \textbf{4.10} & \textbf{32.64} & \textbf{24.39} & \textbf{9.60} & \textbf{93.00} & 24.20 & 21.51 \\
$\beta_{\max}{=}0.15$ & 6.10 & 32.14 & 26.84 & 10.73 & 92.70 & 24.39 & \textbf{21.73} \\
$\beta_{\max}{=}0.00$ (no ring coupling) & 6.12 & 32.13 & 26.87 & 10.74 & 92.69 & \textbf{24.40} & \textbf{21.73} \\
$\beta_{\max}{=}0.45$ (default) & 4.11 & \textbf{32.64} & 24.40 & 9.61 & \textbf{93.00} & 24.21 & 21.51 \\
\bottomrule
\end{tabular}
}
\end{table}

% =========================================================
% Table 6: bd_dilate (scaled to match Table 1)
% =========================================================
\begin{table}[t]
\caption{Ablation on boundary-band dilation radius $r_{\mathrm{bd}}$ used to form $M_{\mathrm{ring}}$ (Eq.~\ref{eq:appendix_ring}) on PIE-Bench (scaled to match Tab.~\ref{tab:pie_full}).}
\label{tab:ablation_bd_dilate}
\centering
\small
\setlength{\tabcolsep}{5pt}
\resizebox{\linewidth}{!}{
\begin{tabular}{l|c|c|c|c|c|c|c}
\toprule
\multirow{2}{*}{Setting} &
\multicolumn{1}{c|}{\textbf{Structure}} &
\multicolumn{4}{c|}{\textbf{Background Preservation}} &
\multicolumn{2}{c}{\textbf{CLIP Similarity$\uparrow$}} \\
\cmidrule(lr){2-2} \cmidrule(lr){3-6} \cmidrule(lr){7-8}
&Distance$\downarrow$ &
PSNR$\uparrow$ & LPIPS$\downarrow$ & MSE$\downarrow$ & SSIM$\uparrow$ &
CLIP$_\text{whole}\uparrow$ & CLIP$_\text{edit}\uparrow$ \\
\midrule
$r_{\mathrm{bd}}{=}2$ & 4.12 & 32.63 & 24.42 & 9.62 & \textbf{93.00} & 24.19 & 21.51 \\
$r_{\mathrm{bd}}{=}6$ & \textbf{4.09} & \textbf{32.65} & \textbf{24.39} & \textbf{9.59} & \textbf{93.00} & 24.19 & 21.51 \\
$r_{\mathrm{bd}}{=}0$ (no ring band) & 4.13 & 32.62 & 24.46 & 9.63 & 92.99 & 24.19 & \textbf{21.52} \\
$r_{\mathrm{bd}}{=}4$ (default) & 4.11 & 32.64 & 24.40 & 9.61 & \textbf{93.00} & \textbf{24.21} & 21.51 \\
\bottomrule
\end{tabular}
}
\end{table}
\subsection{Spatial Supports from the Object Prior}
\label{sec:appendix_supports}

We briefly restate how the object prior (SAM3 mask) is converted into the spatial supports used in Sec.~\ref{sec:stage2}.
Let $M\in\{0,1\}^{H\times W}$ denote the binary object mask (foreground indicator). We define two dilation radii:
$r_{\mathrm{fg}}$ controls the interior object support, and $r_{\mathrm{bd}}$ controls the thickness of the boundary band.
We first form a dilated interior mask
\begin{equation}
\widetilde M_{\mathrm{fg}} = \mathrm{Dilate}(M;\, r_{\mathrm{fg}}),
\qquad
W_t^{\mathrm{edit}} \triangleq \widetilde M_{\mathrm{fg}},
\label{eq:appendix_w_edit}
\end{equation}
which specifies the spatial region where semantic injection is permitted in Eq.~\eqref{eq:final_edit_velocity}.
Next, we form an outer dilated mask and the boundary ring:
\begin{equation}
\widetilde M_{\mathrm{bd}} = \mathrm{Dilate}(\widetilde M_{\mathrm{fg}};\, r_{\mathrm{bd}}),
\qquad
M_{\mathrm{ring}} = \mathrm{clip}\!\big(\widetilde M_{\mathrm{bd}}-\widetilde M_{\mathrm{fg}},0,1\big).
\label{eq:appendix_ring}
\end{equation}
Finally, we define the reference support that couples a thin boundary band with a time-adaptive strength:
\begin{equation}
W_t^{\mathrm{ref}}
=
\mathrm{clip}\!\Big(W_t^{\mathrm{edit}} + \beta(t)\, M_{\mathrm{ring}},\,0,1\Big),
\qquad
\beta(t)=\beta_{\max}\Big(\frac{\sigma_t}{\sigma_{t_s}}\Big)^{\gamma}.
\label{eq:appendix_w_ref}
\end{equation}
Hence the boundary transition support in Eq.~\eqref{eq:app_w_ring} is $W_t^{\mathrm{ring}}=\mathrm{clip}(W_t^{\mathrm{ref}}-W_t^{\mathrm{edit}},0,1)=\beta(t)\,M_{\mathrm{ring}}$, i.e., the ring band weighted by the time-adaptive coupling.
This construction instantiates the spatial support $W_t$ in Eq.~\eqref{eq:where_how_factorization}.
Specifically, $M_{\mathrm{fg}}$ defines the editable interior support $W_t^{\mathrm{edit}}$, while $M_{\mathrm{ring}}$ is used only in the reference support $W_t^{\mathrm{ref}}$ through the fixed boundary-coupling schedule in Eq.~\eqref{eq:app_beta}.
These supports are then used for velocity-level control in Stage-2, including the reference-anchor update in Eq.~\eqref{eq:ref_update_stage2} and the final object-localized velocity in Eq.~\eqref{eq:final_edit_velocity}.

\subsection{Qualitative Transfer to FlowEdit}
\label{sec:appendix_flowedit_transfer}

\begin{figure}[t]
\centering
\vspace{-2mm}
\setlength{\tabcolsep}{0pt}

{\footnotesize
\begin{tabularx}{\linewidth}{@{} V *{3}{K} @{\hspace{3mm}} V *{3}{K} @{}}

& \textbf{Original} & \textbf{FlowEdit} & \textbf{OAVC on FlowEdit}
&& \textbf{Original} & \textbf{FlowEdit} & \textbf{OAVC on FlowEdit} \\[0.01mm]

\raisebox{0.1pt}{\vlabel{\large \textit{cake$\rightarrow$ice cream}}} &
\imgcell{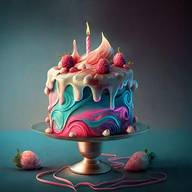} &
\imgcell{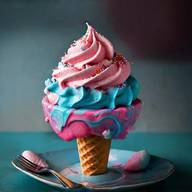} &
\imgcell{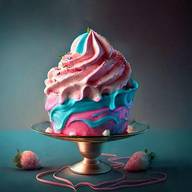} &
\raisebox{-3pt}{\vlabel{\large \textit{tiger$\rightarrow$cat}}} &
\imgcell{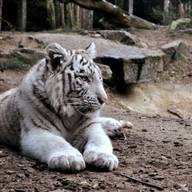} &
\imgcell{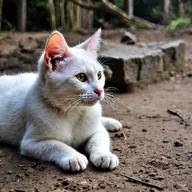} &
\imgcell{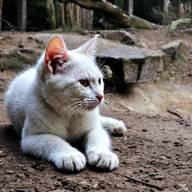} \\ \noalign{\vskip 0.5mm}

\raisebox{-2pt}{\vlabel{\large \textit{plaid$\rightarrow$floral}}} &
\imgcell{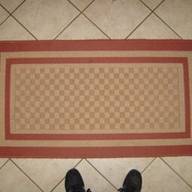} &
\imgcell{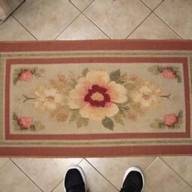} &
\imgcell{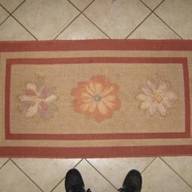} &
\raisebox{-3pt}{\vlabel{\large \textit{add a car}}} &
\imgcell{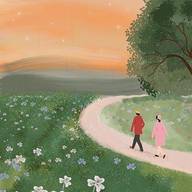} &
\imgcell{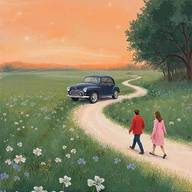} &
\imgcell{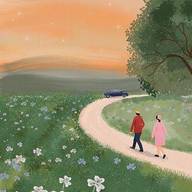} \\ \noalign{\vskip 0.5mm}

\raisebox{-3pt}{\vlabel{\large \textit{delete flower}}} &
\imgcell{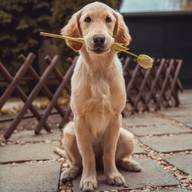} &
\imgcell{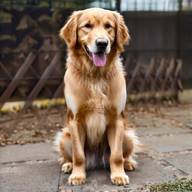} &
\imgcell{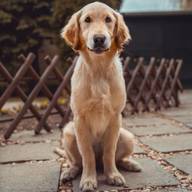} &
\raisebox{-3pt}{\vlabel{\large \textit{trees$\rightarrow$a city}}} &
\imgcell{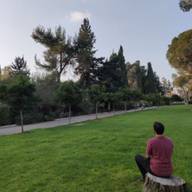} &
\imgcell{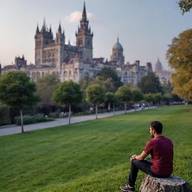} &
\imgcell{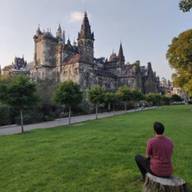} \\

\end{tabularx}
}

\vspace{0.6mm}
\caption{Qualitative transfer of our Stage-2 \emph{where/how} control to FlowEdit.
Left: SD3. Right: FLUX. Each column shows (Original, FlowEdit, and OAVC applied to FlowEdit) on three examples.}
\label{fig:appendix_flowedit_transfer_qual}
\end{figure}

In Sec.~\ref{sec:exp_main} we quantitatively evaluate whether the proposed \emph{where/how} control can be transferred to another velocity-based editor, FlowEdit.
Here we provide additional qualitative results in Fig.~\ref{fig:appendix_flowedit_transfer_qual}.
We apply only the Stage-2 controlled injection to FlowEdit by using the same object support $W_t^{\mathrm{edit}}$ and the same safe injection operator when forming the prompt-induced update.
This transfer does not include Stage-1, since FlowEdit does not construct the same noise-aligned reference interface as OAVC.

The visualization shows that applying OAVC-style control to FlowEdit can better localize the intended edit and reduce unnecessary changes in non-target regions.
For example, in object replacement and object removal cases, FlowEdit often changes surrounding structures or introduces global appearance shifts, while the controlled variant better preserves the original layout.
These qualitative results are consistent with the transfer robustness results in Tab.~\ref{tab:flowedit_transfer}: the \emph{where/how} mechanism is broadly useful beyond our base pipeline, while full stability still depends on the alignment quality of the underlying editor.

\subsection{Ablation on Interior Support Radius $r_{\mathrm{fg}}$}
\label{sec:appendix_fg}

Tab.~\ref{tab:ablation_fg_dilate} studies the sensitivity to the interior-support dilation radius $r_{\mathrm{fg}}$ in Eq.~\eqref{eq:appendix_w_edit}.
Increasing $r_{\mathrm{fg}}$ expands the region where semantic injection is allowed, which can improve edit coverage but may also bring the injection closer to background structures near the boundary.
The sweep therefore reflects a preservation--editability trade-off rather than a universally optimal dilation value: $r_{\mathrm{fg}}=1$ gives the strongest aggregate structure/background preservation, whereas $r_{\mathrm{fg}}=2$ gives slightly higher CLIP$_{\mathrm{edit}}$ and additional spatial freedom for shape-changing edits.
We use $r_{\mathrm{fg}}=2$ as the default operating point in all main experiments; excessive expansion at $r_{\mathrm{fg}}=3$ provides no further semantic gain and weakens preservation.

\subsection{Ablation on Maximum Ring Coupling}
\label{sec:appendix_beta}

Tab.~\ref{tab:ablation_beta_max} ablates $\beta_{\max}$ in Eq.~\eqref{eq:appendix_w_ref}.
When $\beta_{\max}=0$, the ring coupling is removed and the method reduces to a hard spatial split between edited and preserved regions.
This increases Struct.\ Dist.\ and worsens LPIPS/MSE, consistent with boundary artifacts (e.g., halos and contour wobble) accumulating under multi-step integration.
Moderate coupling improves stability, and the default $\beta_{\max}=0.45$ provides the best overall balance while being robust to nearby values.

\subsection{Ablation on Boundary Band Thickness $r_{\mathrm{bd}}$}
\label{sec:appendix_bd}

Tab.~\ref{tab:ablation_bd_dilate} studies the boundary-band dilation radius $r_{\mathrm{bd}}$ in Eq.~\eqref{eq:appendix_ring}.
We find the method is not sensitive to a wide range of ring thicknesses: varying $r_{\mathrm{bd}}$ yields nearly identical metrics.
Removing the explicit ring band (effectively collapsing $M_{\mathrm{ring}}$) slightly degrades stability-related measures.
This indicates that explicitly treating the boundary region differently from the object interior is beneficial, while the precise band thickness is not critical.

\section{More Qualitative Results}
\label{sec:appendix_more_qual}

This appendix provides additional visual comparisons for both image and video editing.
Figs.~\ref{fig:qualitative_results_v2}, \ref{fig:qualitative_results_v3}, and \ref{fig:sd_flux_combined_v2} report more qualitative results on PIE-Bench, complementing the main-paper examples.
Fig.~\ref{fig:davis} shows qualitative results on DAVIS for video editing.
All experiments in this work were run on a single NVIDIA GPU with 48~GB of memory.

\subsection{More Qualitative Results for Image Editing}
\label{sec:appendix_more_qual_image}
\begin{figure}[t]
\centering
\vspace{-2mm}
\setlength{\tabcolsep}{0pt}

{\scriptsize
\begin{tabularx}{\linewidth}{@{} V *{10}{K} @{}}
\noalign{\vskip 2mm} 
& \textbf{Source} & \textbf{DNAEdit-SD3} & \textbf{Ours-SD3} & \textbf{DNAEdit-FLUX} & \textbf{Ours-FLUX} & \textbf{FlowEdit} & \textbf{UniEdit} & \textbf{FireFlow} & \textbf{Inf} & \textbf{RF-Solver} \\[0.01mm]

% ---------------- Row 1 ----------------
\vlabel{Cake:\\Round $\rightarrow$ Square} & 
\imgcell{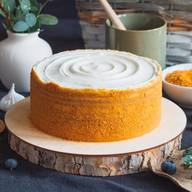}&
\imgcell{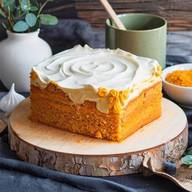}&
\imgnode{s_top}{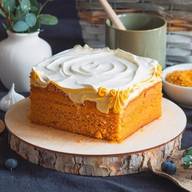}&
\imgcell{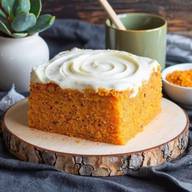}&
\imgnode{f_top}{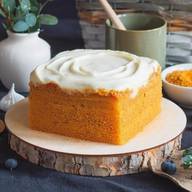}&
\imgcell{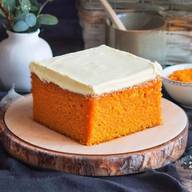}&
\imgcell{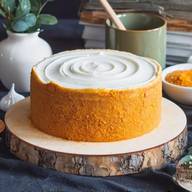}&
\imgcell{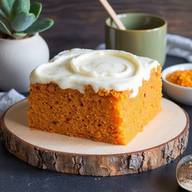}&
\imgcell{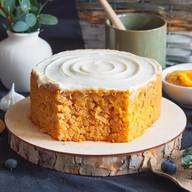}&
\imgcell{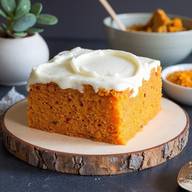} \\ \noalign{\vskip 0.5mm}

% ---------------- Row 2 ----------------
\raisebox{-5pt}{\vlabel{Plate:\\Fruits $\rightarrow$ Pizza}} & 
\imgcell{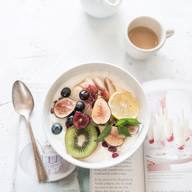}&
\imgcell{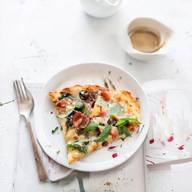}&
\imgcell{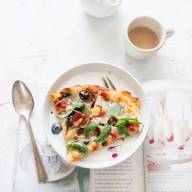}&
\imgcell{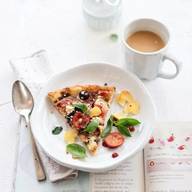}&
\imgcell{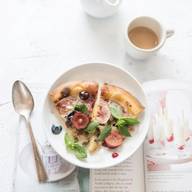}&
\imgcell{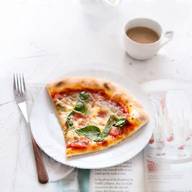}&
\imgcell{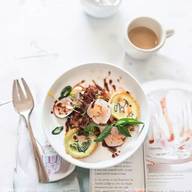}&
\imgcell{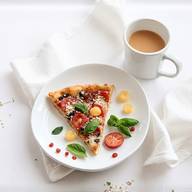}&
\imgcell{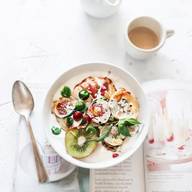}&
\imgcell{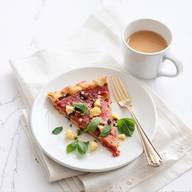} \\ \noalign{\vskip 0.5mm}

% ---------------- Row 3 ----------------
\raisebox{-5pt}{\vlabel{Eyes:\\Opened$\rightarrow$Closed}} & 
\imgcell{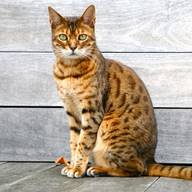}&
\imgcell{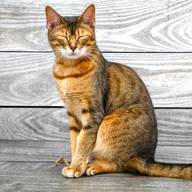}&
\imgcell{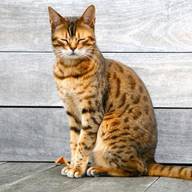}&
\imgcell{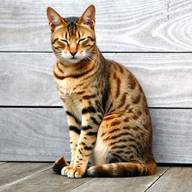}&
\imgcell{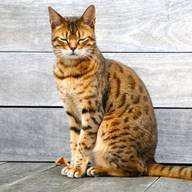}&
\imgcell{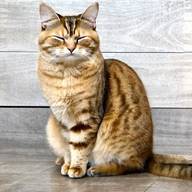}&
\imgcell{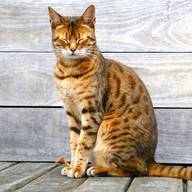}&
\imgcell{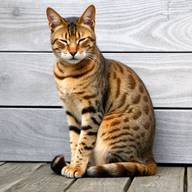}&
\imgcell{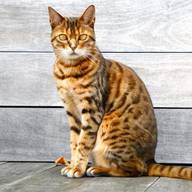}&
\imgcell{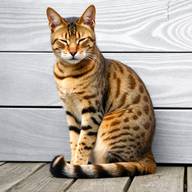} \\ \noalign{\vskip 0.5mm}

% ---------------- Row 4 ----------------
\raisebox{-5pt}{\vlabel{Eyes:\\Green $\rightarrow$ Blue}} & 
\imgcell{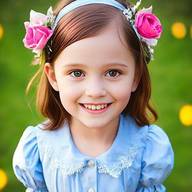}&
\imgcell{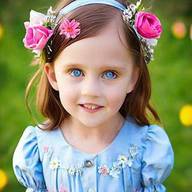}&
\imgcell{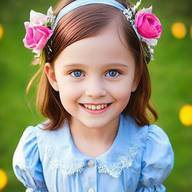}&
\imgcell{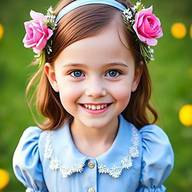}&
\imgcell{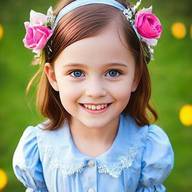}&
\imgcell{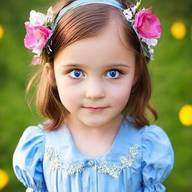}&
\imgcell{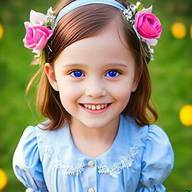}&
\imgcell{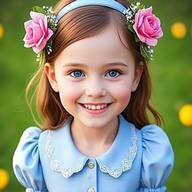}&
\imgcell{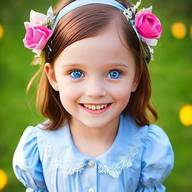}&
\imgcell{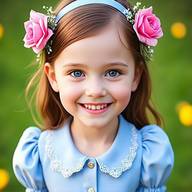} \\ \noalign{\vskip 0.5mm}

% ---------------- Row 5 ----------------
\raisebox{-5pt}{\vlabel{Glowing\\jar$\rightarrow$Crystal ball}} & 
\imgcell{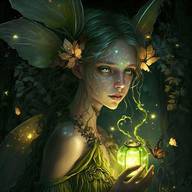}&
\imgcell{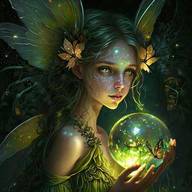}&
\imgcell{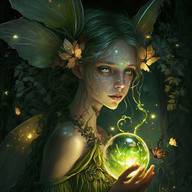}&
\imgcell{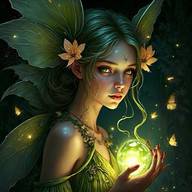}&
\imgcell{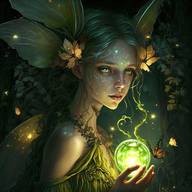}&
\imgcell{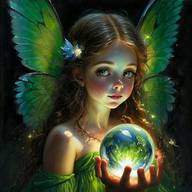}&
\imgcell{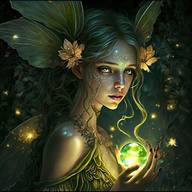}&
\imgcell{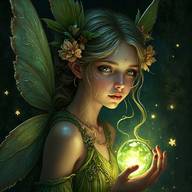}&
\imgcell{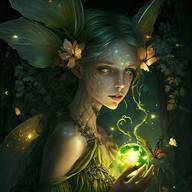}&
\imgcell{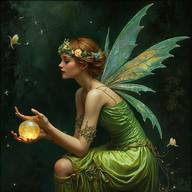} \\ \noalign{\vskip 0.5mm}

% ---------------- Row 6 ----------------
\raisebox{-5pt}{\vlabel{Standing on:\\Rocks$\rightarrow$Boat}} & 
\imgcell{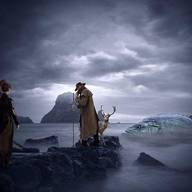}&
\imgcell{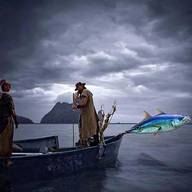}&
\imgcell{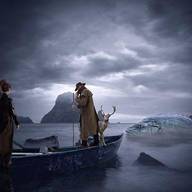}&
\imgcell{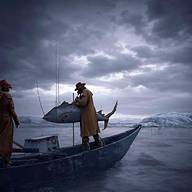}&
\imgcell{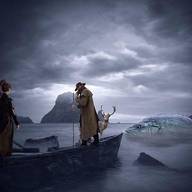}&
\imgcell{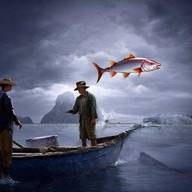}&
\imgcell{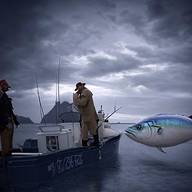}&
\imgcell{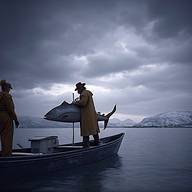}&
\imgcell{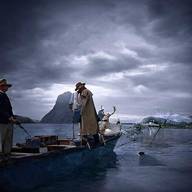}&
\imgcell{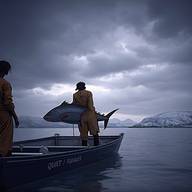} \\ \noalign{\vskip 0.5mm}

% ---------------- Row 7 ----------------
\raisebox{-10pt}{\vlabel{Table:\\Vase$\rightarrow$Plant}} & 
\imgcell{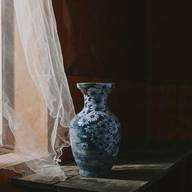}&
\imgcell{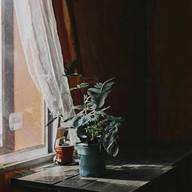}&
\imgcell{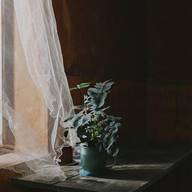}&
\imgcell{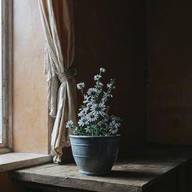}&
\imgcell{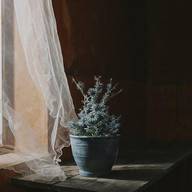}&
\imgcell{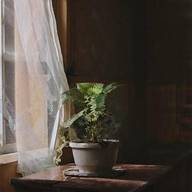}&
\imgcell{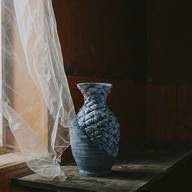}&
\imgcell{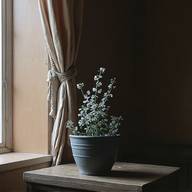}&
\imgcell{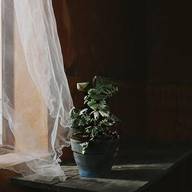}&
\imgcell{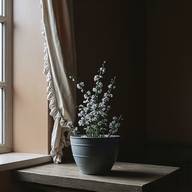} \\ \noalign{\vskip 0.5mm}

% ---------------- Row 8 ----------------
\vlabel{Woods:\\House$\rightarrow$Monster} & 
\imgcell{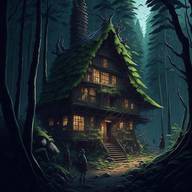}&
\imgcell{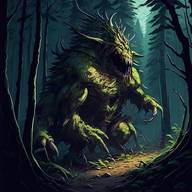}&
\imgcell{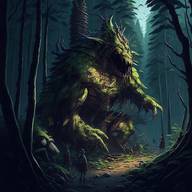}&
\imgcell{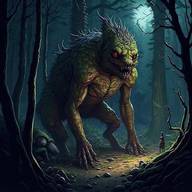}&
\imgcell{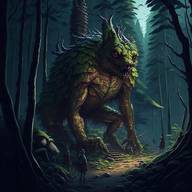}&
\imgcell{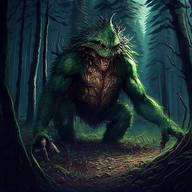}&
\imgcell{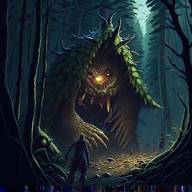}&
\imgcell{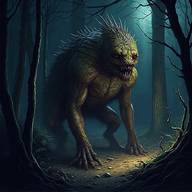}&
\imgcell{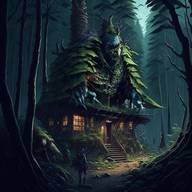}&
\imgcell{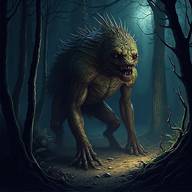} \\ \noalign{\vskip 0.5mm}

% ---------------- Row 9 ----------------
\raisebox{-5pt}{\vlabel{+Exaggerated\\hair band}} & 
\imgcell{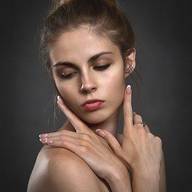}&
\imgcell{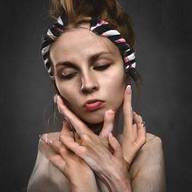}&
\imgcell{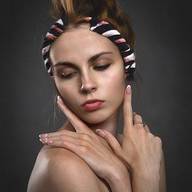}&
\imgcell{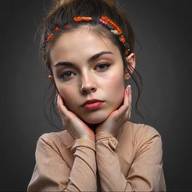}&
\imgcell{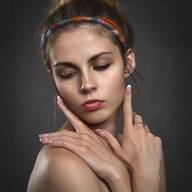}&
\imgcell{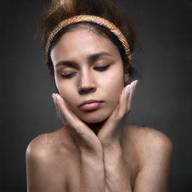}&
\imgcell{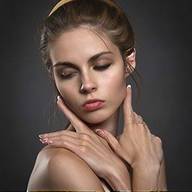}&
\imgcell{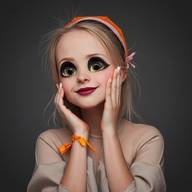}&
\imgcell{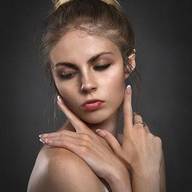}&
\imgcell{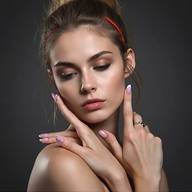} \\ \noalign{\vskip 0.5mm}

% ---------------- Row 10 ----------------
\raisebox{-10pt}{\vlabel{Woman:\\-Backpack}} & 
\imgcell{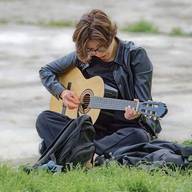}&
\imgcell{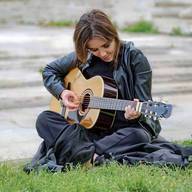}&
\imgnode{s_bot}{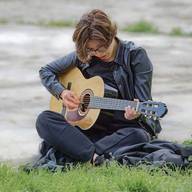}&
\imgcell{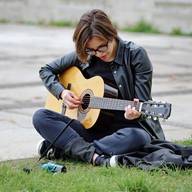}&
\imgnode{f_bot}{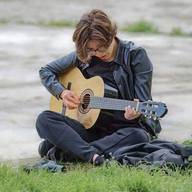}&
\imgcell{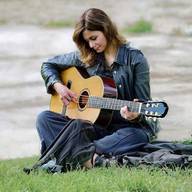}&
\imgcell{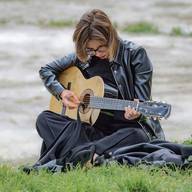}&
\imgcell{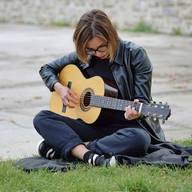}&
\imgcell{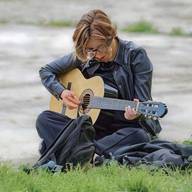}&
\imgcell{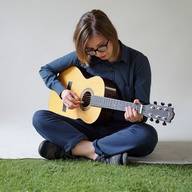}    \\ \relax

\end{tabularx}
}

\vspace{-4mm} 
\caption{Qualitative comparison of image editing methods on 10 samples.}
\label{fig:qualitative_results_v2}
\end{figure}

\begin{figure}[t]
\centering
\vspace{-2mm}
\setlength{\tabcolsep}{0pt}

{\scriptsize
\begin{tabularx}{\linewidth}{@{} V *{10}{K} @{}}
\noalign{\vskip 2mm} 
& \textbf{Source} & \textbf{DNAEdit-SD3} & \textbf{Ours-SD3} & \textbf{DNAEdit-FLUX} & \textbf{Ours-FLUX} & \textbf{FlowEdit} & \textbf{UniEdit} & \textbf{FireFlow} & \textbf{Inf} & \textbf{RF-Solver} \\[0.01mm]

\raisebox{-10pt}{\vlabel{Mountain:\\-Paraglider}} & 
\imgcell{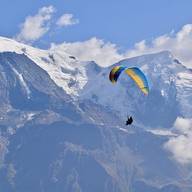}&
\imgcell{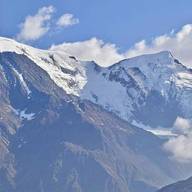}&
\imgnode{s_top}{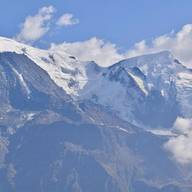}&
\imgcell{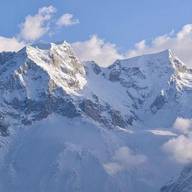}&
\imgnode{f_top}{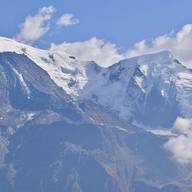}&
\imgcell{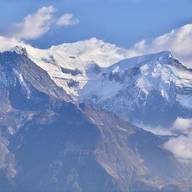}&
\imgcell{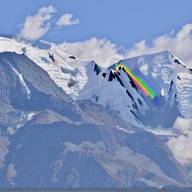}&
\imgcell{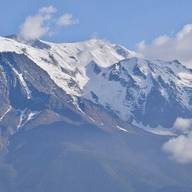}&
\imgcell{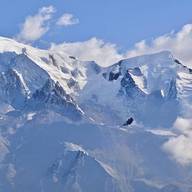}&
\imgcell{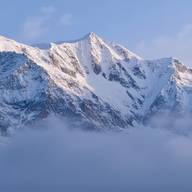} \\ \noalign{\vskip 0.5mm}

% ---------------- Row 11 ----------------
\raisebox{-5pt}{\vlabel{Umbrella:\\\textcolor{pink}{Pink}$\rightarrow$\textcolor{yellow!60!brown}{Yellow}}} & 
\imgcell{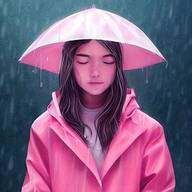}&
\imgcell{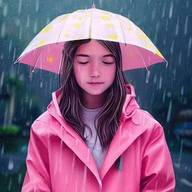}&
\imgcell{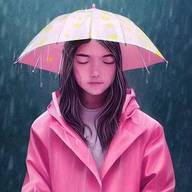}&
\imgcell{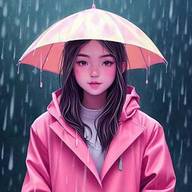}&
\imgcell{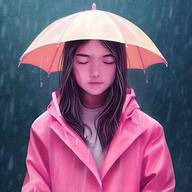}&
\imgcell{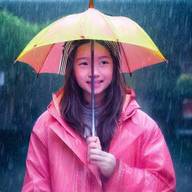}&
\imgcell{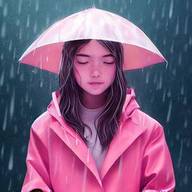}&
\imgcell{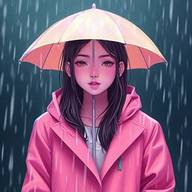}&
\imgcell{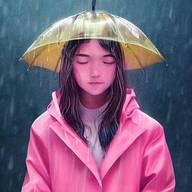}&
\imgcell{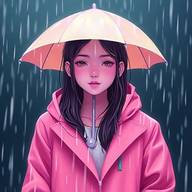} \\ \noalign{\vskip 0.5mm}

% ---------------- Row 12 ----------------
\raisebox{-5pt}{\vlabel{Lipstick:\\\textcolor{red}{Red}$\rightarrow$\textcolor{purple}{Purple}}} & 
\imgcell{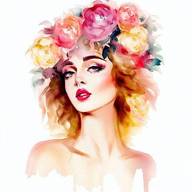}&
\imgcell{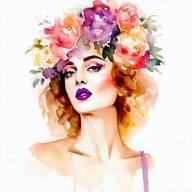}&
\imgcell{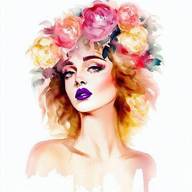}&
\imgcell{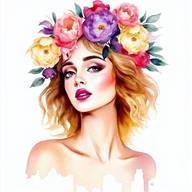}&
\imgcell{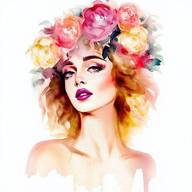}&
\imgcell{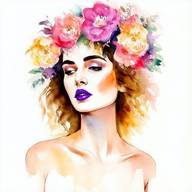}&
\imgcell{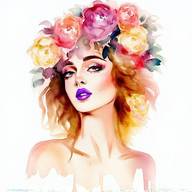}&
\imgcell{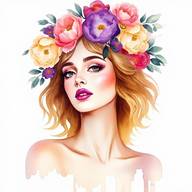}&
\imgcell{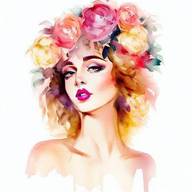}&
\imgcell{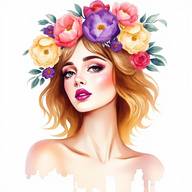} \\ \noalign{\vskip 0.5mm}

% ---------------- Row 13 ----------------
\raisebox{-10pt}{\vlabel{Chairs:\\+Plastic}} & 
\imgcell{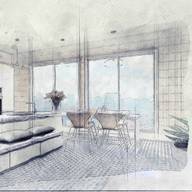}&
\imgcell{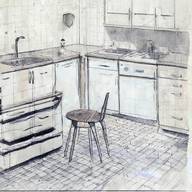}&
\imgcell{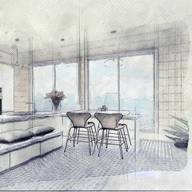}&
\imgcell{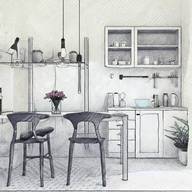}&
\imgcell{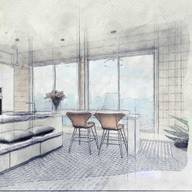}&
\imgcell{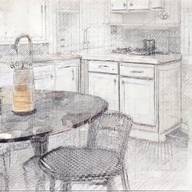}&
\imgcell{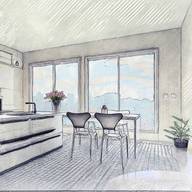}&
\imgcell{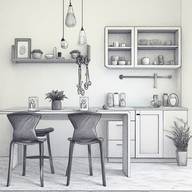}&
\imgcell{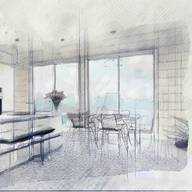}&
\imgcell{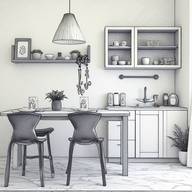} \\ \noalign{\vskip 0.5mm}

\raisebox{-5pt}{\vlabel{Background:\\Trees$\rightarrow$Elves}} & 
\imgcell{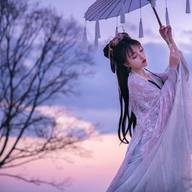}&
\imgcell{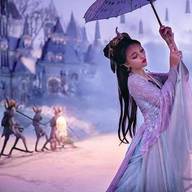}&
\imgnode{s_bot}{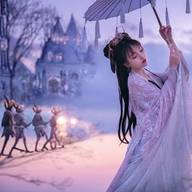}&
\imgcell{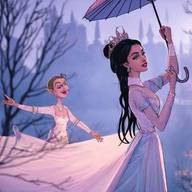}&
\imgnode{f_bot}{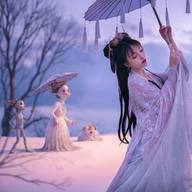}&
\imgcell{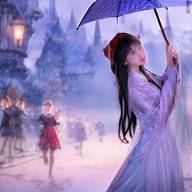}&
\imgcell{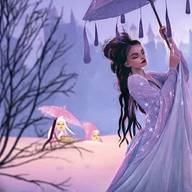}&
\imgcell{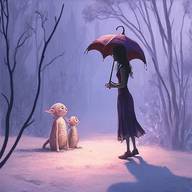}&
\imgcell{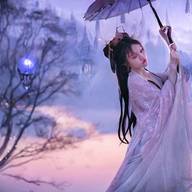}&
\imgcell{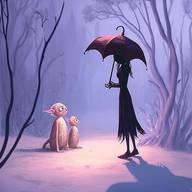}    \\ \relax

\end{tabularx}
}

\vspace{-4mm} 
\caption{Additional qualitative comparison of image editing methods on 5 PIE-Bench samples.}
\label{fig:qualitative_results_v3}
\end{figure}

Figs.~\ref{fig:qualitative_results_v2}, \ref{fig:qualitative_results_v3}, and \ref{fig:sd_flux_combined_v2} present additional image-editing results on PIE-Bench across diverse edit types
(e.g., object replacement, attribute/material changes, deletion/addition, and local semantic edits).
We observe that several baselines may exhibit background drift, boundary artifacts (e.g., halos/color bleeding),
or occasional mis-edits under challenging prompts.
In contrast, OAVC better confines semantic changes to the intended region and preserves non-target content,
yielding cleaner boundaries and stronger structure/background consistency.
Notably, for human-centric cases, OAVC more reliably preserves identity-related characteristics while applying the requested edits.

\begin{figure}[t]
    \centering
    \includegraphics[width=\textwidth]{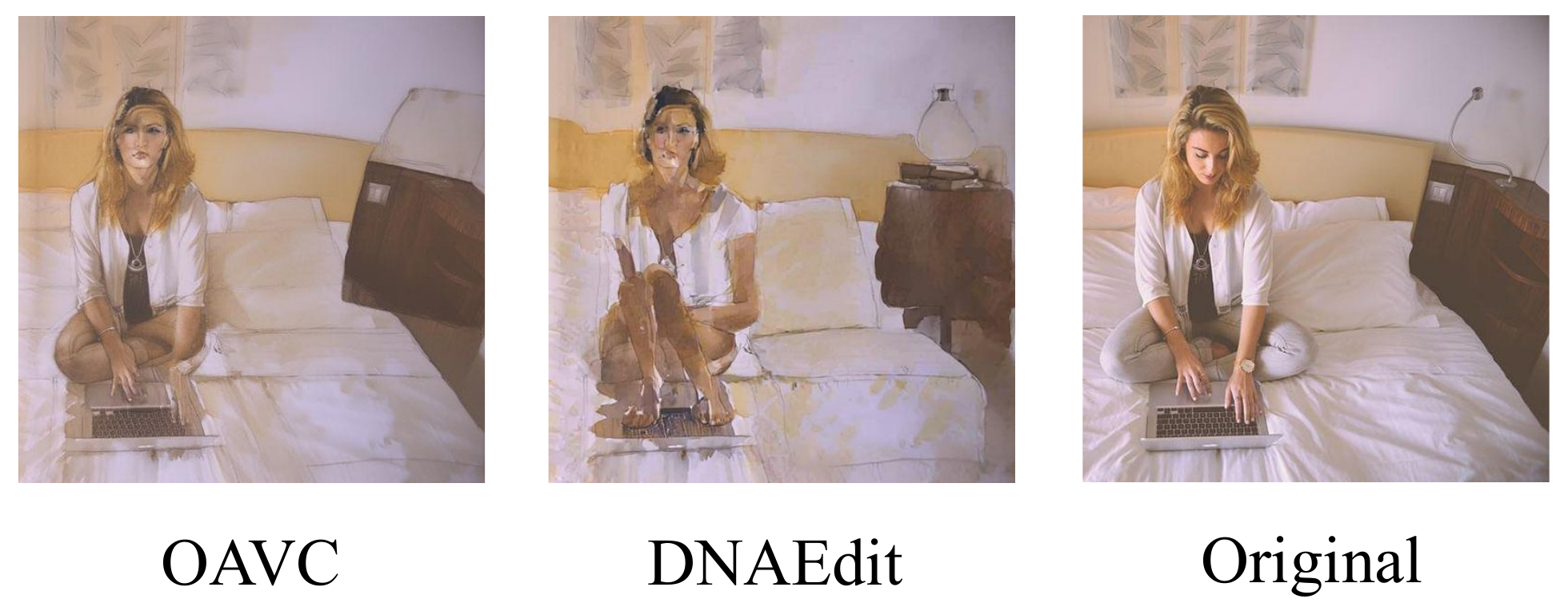}
    \caption{
    Representative weak-benefit case 1 for OAVC.
    }
    \label{fig:failure_global_edit_1}
\end{figure}

\begin{figure}[t]
    \centering
    \includegraphics[width=\linewidth]{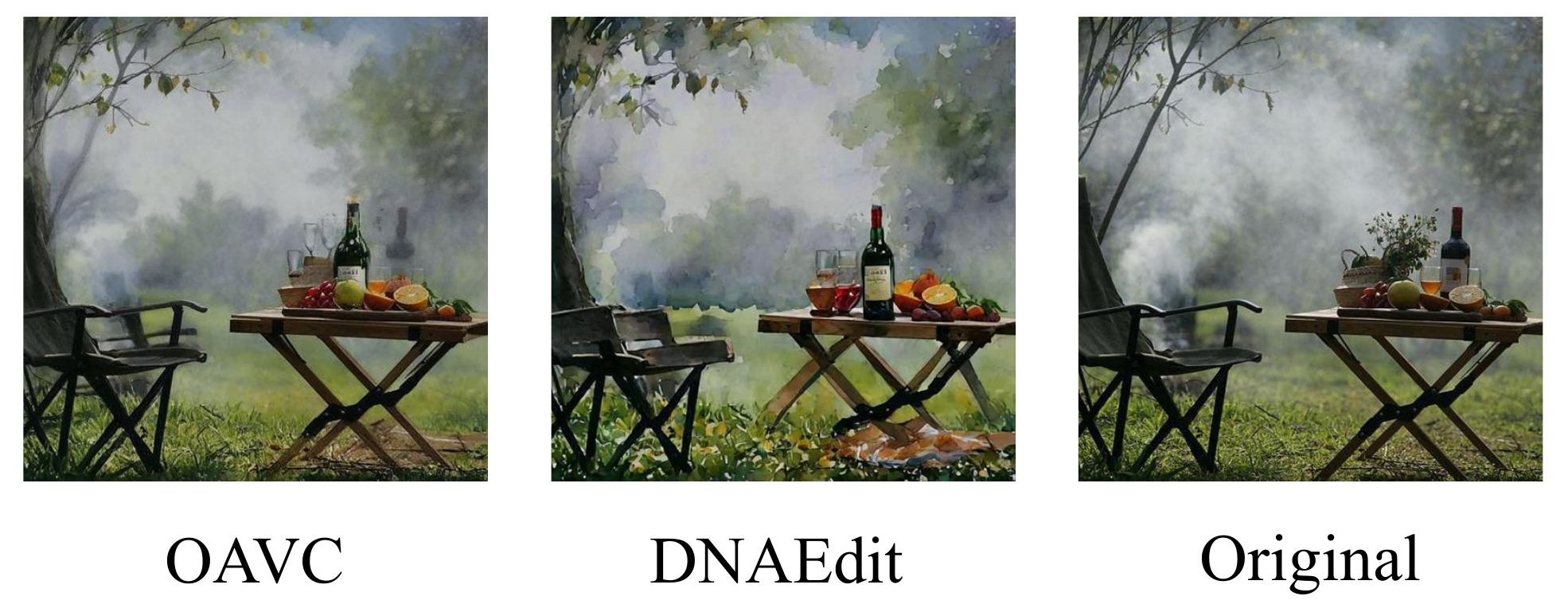}
    \caption{
    Representative weak-benefit case 2 for OAVC.
    }
    \label{fig:failure_global_edit_2}
\end{figure}
\subsection{More Qualitative Results for Video Editing on DAVIS}
\label{sec:appendix_more_qual_video}
\begin{figure}[t]
    \centering
    \includegraphics[width=1.0\linewidth]{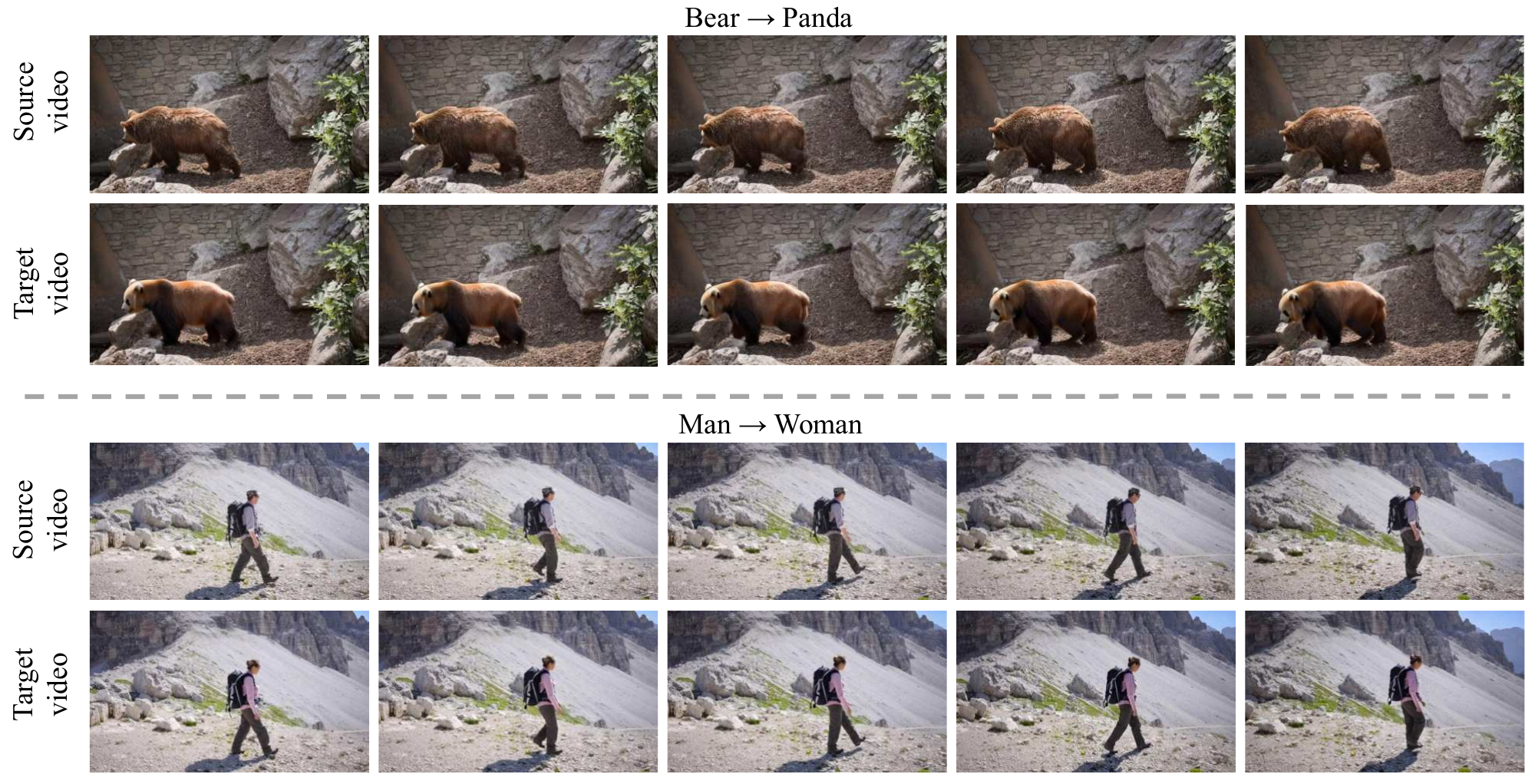}
    
    \caption{Qualitative video editing results on DAVIS using Wan2.1. OAVC reduces background drift and improves temporal consistency by constraining semantic updates to the target object region (and a safe injection direction), compared to global velocity steering baselines.}
    \label{fig:davis}
\end{figure}

Fig.~\ref{fig:davis} provides additional video-editing results on DAVIS.
Following prior work, we use the Rectified-Flow text-to-video backbone Wan2.1 for DAVIS experiments.
Across challenging sequences, OAVC reduces temporal drift and improves frame-to-frame consistency,
while maintaining the original motion patterns and scene layout compared to global-steering baselines.

\section{Additional Limitation Visualizations}
\label{sec:app_limitations}

Our improvements rely on a non-trivial \emph{object support} that localizes where semantic velocity updates are allowed to act.
When the mask degenerates to a near-global support (e.g., $M_{\mathrm{fg}}\approx \mathbf{1}$), the spatial selectivity of Stage-2 vanishes:
$W_t^{\mathrm{edit}}\approx W_t^{\mathrm{ref}}\approx \mathbf{1}$ makes the controlled injection equivalent to global velocity steering.
In this regime, OAVC effectively reduces to the reference/offset stabilization in Stage-1 and behaves similarly to prior global editors (e.g., DNAEdit), offering limited advantage for edits that are inherently global.

Figs.~\ref{fig:failure_global_edit_1} and \ref{fig:failure_global_edit_2} show representative weak-benefit cases.
For near-global transformations such as watercolor painting style transfer, the intended change affects most of the image rather than a localized object.
As a result, the object/background separation becomes ill-defined, background anchoring becomes less informative, and boundary-aware object control has limited effect.
These examples visually complement the scope discussion in Sec.~\ref{sec:main_limitations}.

Consequently, OAVC is not designed to be optimal for full-image transformations such as global style transfer, overall color grading, or domain-level appearance changes, where the desired modification intentionally spans the entire image.
Extending our framework to these cases likely requires alternative priors beyond object masks (e.g., multi-region semantic layouts or frequency/texture-specific supports) and revised constraints that balance global appearance changes with structure preservation.
Our current post-hoc control uses a single final latent hard blend; stronger pixel-space or multi-step compositing controls remain to be evaluated.
The reported Structure Distance is a global correspondence metric, so the present experiments do not isolate whether every improvement is foreground- or boundary-localized.
Finally, the human study was not powered for reliable per-edit-category analysis of the elevated ``Partially'' rate; the aggregate results should not be interpreted as equivalence in edit strength.

\newlength{\trcolsep}
\setlength{\trcolsep}{1.0mm}
\newlength{\trrowsep}
\setlength{\trrowsep}{2.0mm}
\newlength{\trcellw}

\newcolumntype{T}{>{\centering\arraybackslash}m{\trcellw}}

\newcommand{\trimg}[1]{%
  \begin{minipage}[c]{\trcellw}%
    \centering
    \includegraphics[width=\linewidth]{#1}%
  \end{minipage}%
}

\newlength{\sdcolsep}
\setlength{\sdcolsep}{1.0mm}   % horizontal gap between cells (smaller = safer)
\newlength{\sdrowsep}
\setlength{\sdrowsep}{2.0mm}   % vertical gap between rows

% ---- cell width (set INSIDE each minipage using \linewidth) ----
\newlength{\sdcellw}
\newcolumntype{S}{>{\centering\arraybackslash}p{\sdcellw}}

% ---- robust definition of \imgcell (works whether previously defined or not) ----
\makeatletter
\@ifundefined{imgcell}{
  \newcommand{\imgcell}[1]{\includegraphics[width=\linewidth]{#1}}
}{
  \renewcommand{\imgcell}[1]{\includegraphics[width=\linewidth]{#1}}
}
\makeatother

% ---- your root ----
\newcommand{\sdroot}{Image1}

\begin{figure}[t]
\centering
\vspace{-2mm}
\setlength{\tabcolsep}{0pt}

{\scriptsize
\begin{tabularx}{\linewidth}{@{} V *{4}{K} @{\hspace{3mm}} V *{4}{K} @{}}

% & \multicolumn{4}{c}{\textbf{Stable Diffusion (SD3)}} && \multicolumn{4}{c}{\textbf{FLUX.1-dev}} \\
% \cmidrule(lr){2-5} \cmidrule(lr){7-10}

& \textbf{Source} & \textbf{FlowEdit} & \textbf{DNAEdit} & \textbf{Ours} 
&& \textbf{Source} & \textbf{UniEdit} & \textbf{DNAEdit} & \textbf{Ours} \\[0.01mm]

% ---------------- Row 1 ----------------
\raisebox{10pt}{\vlabel{Hold:\\Torch $\rightarrow$ Flower}} & 
\imgcell{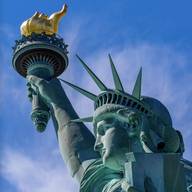} &
\imgcell{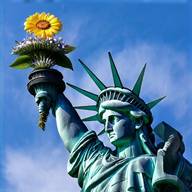} &
\imgcell{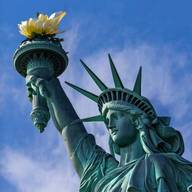} &
\imgcell{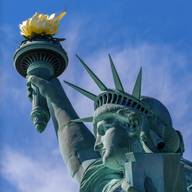} &
\raisebox{7pt}{\vlabel{Chair:\\Cat $\rightarrow$ Dog}} &
\imgcell{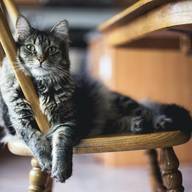} &
\imgcell{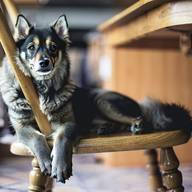} &
\imgcell{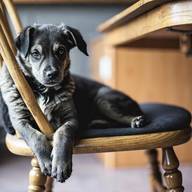} &
\imgcell{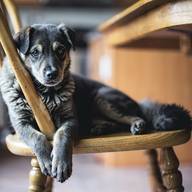} \\ \noalign{\vskip 0.5mm}

% ---------------- Row 2 ----------------
\raisebox{10pt}{\vlabel{Woman:\\Jacket $\rightarrow$ Blouse}} & 
\imgcell{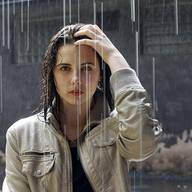} &
\imgcell{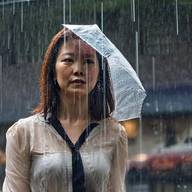} &
\imgcell{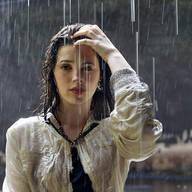} &
\imgcell{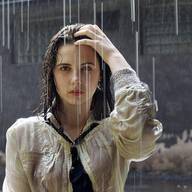} &
\raisebox{10pt}{\vlabel{Street:\\Car $\rightarrow$ Motorcycle}} &
\imgcell{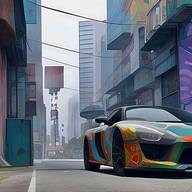} &
\imgcell{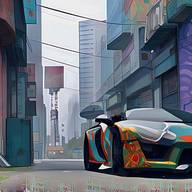} &
\imgcell{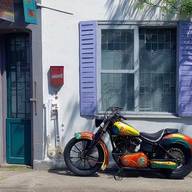} &
\imgcell{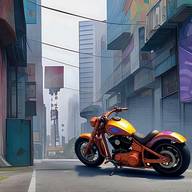} \\ \noalign{\vskip 0.5mm}

% ---------------- Row 3 ----------------
\raisebox{10pt}{\vlabel{Table:\\Pizza $\rightarrow$ Noodles}} & 
\imgcell{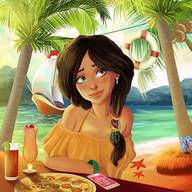} &
\imgcell{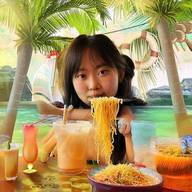} &
\imgcell{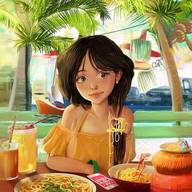} &
\imgcell{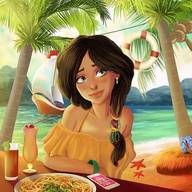} &
\raisebox{10pt}{\vlabel{Seat:\\Bench $\rightarrow$ Sofa}} &
\imgcell{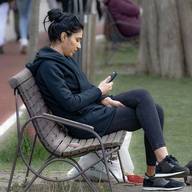} &
\imgcell{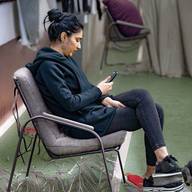} &
\imgcell{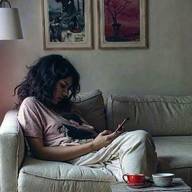} &
\imgcell{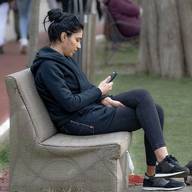} \\ \noalign{\vskip 0.5mm}

% ---------------- Row 4 ----------------
\raisebox{10pt}{\vlabel{Hold:\\Flower $\rightarrow$ Bear}} & 
\imgcell{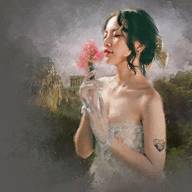} &
\imgcell{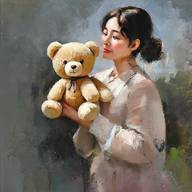} &
\imgcell{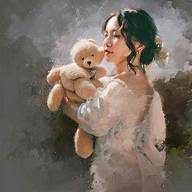} &
\imgcell{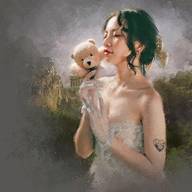} &
\raisebox{10pt}{\vlabel{Path:\\ - husky dog}} &
\imgcell{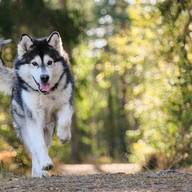} &
\imgcell{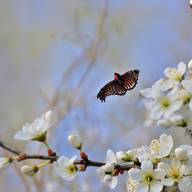} &
\imgcell{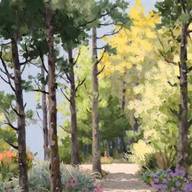} &
\imgcell{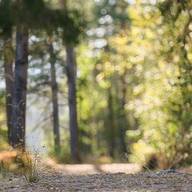} \\ \noalign{\vskip 0.5mm}

% ---------------- Row 5 ----------------
\raisebox{10pt}{\vlabel{Easter eggs:\\Teacup $\rightarrow$ Glass}} & 
\imgcell{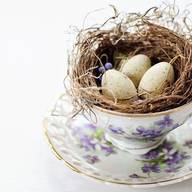} &
\imgcell{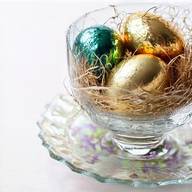} &
\imgcell{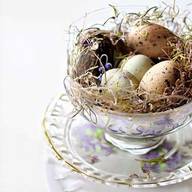} &
\imgcell{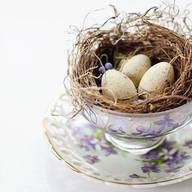} &
\raisebox{5pt}{\vlabel{Girl:\\- Dog}} &
\imgcell{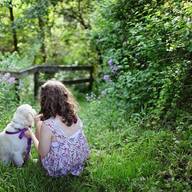} &
\imgcell{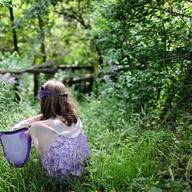} &
\imgcell{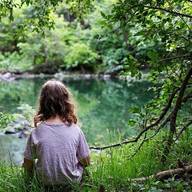} &
\imgcell{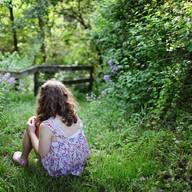} \\ \noalign{\vskip 0.5mm}

% ---------------- Row 6 ----------------
\raisebox{5pt}{\vlabel{Table:\\- Laptop}} & 
\imgcell{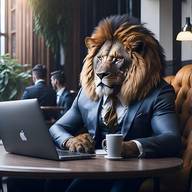} &
\imgcell{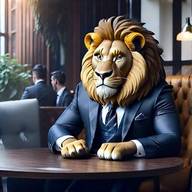} &
\imgcell{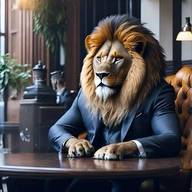} &
\imgcell{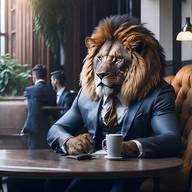} &
\raisebox{5pt}{\vlabel{Cake:\\Round $\rightarrow$ Star}} &
\imgcell{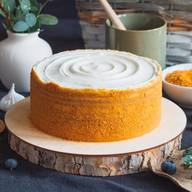} &
\imgcell{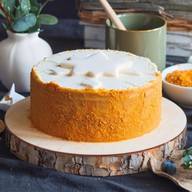} &
\imgcell{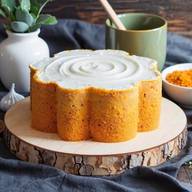} &
\imgcell{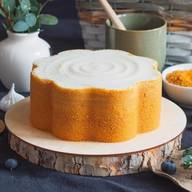} \\ \noalign{\vskip 0.5mm}

% ---------------- Row 7 ----------------
\raisebox{5pt}{\vlabel{Road:\\- Bus}} & 
\imgcell{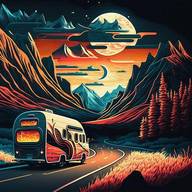} &
\imgcell{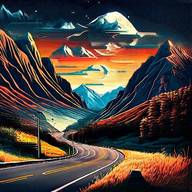} &
\imgcell{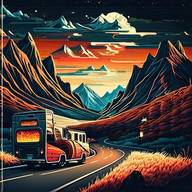} &
\imgcell{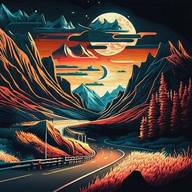} &
\vlabel{Road:\\+ Flowers} &
\imgcell{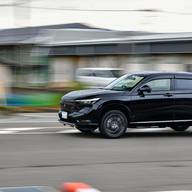} &
\imgcell{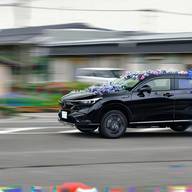} &
\imgcell{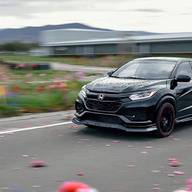} &
\imgcell{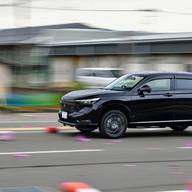} \\ \noalign{\vskip 0.5mm}

% ---------------- Row 8 ----------------
\raisebox{10pt}{\vlabel{Tree:\\Dead $\rightarrow$ Blooming}} & 
\imgcell{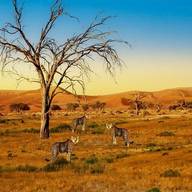} &
\imgcell{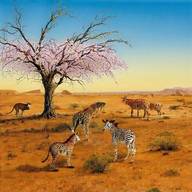} &
\imgcell{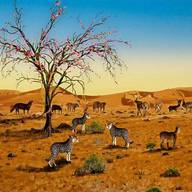} &
\imgcell{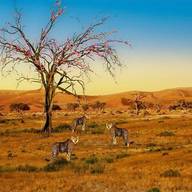} &
\raisebox{5pt}{\vlabel{Colorful eggs:\\Rabbit $\rightarrow$ Cat}} &
\imgcell{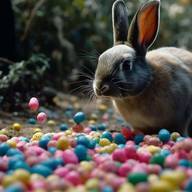} &
\imgcell{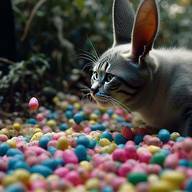} &
\imgcell{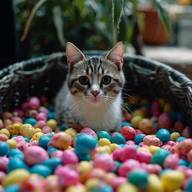} &
\imgcell{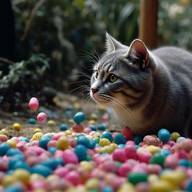} \\ \noalign{\vskip 0.5mm}

% ---------------- Row 9 ----------------
\raisebox{10pt}{\vlabel{Background:\\Dirt $\rightarrow$ Snow}} & 
\imgcell{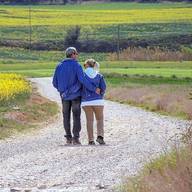} &
\imgcell{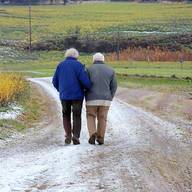} &
\imgcell{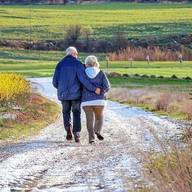} &
\imgcell{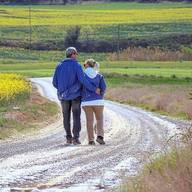} &
\raisebox{10pt}{\vlabel{Head:\\Garland $\rightarrow$ Hat}} &
\imgcell{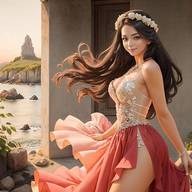} &
\imgcell{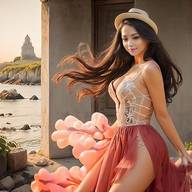} &
\imgcell{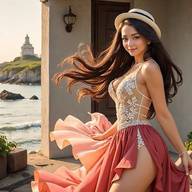} &
\imgcell{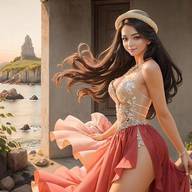} \\

\end{tabularx}
}

\vspace{0.01mm}
\caption{Qualitative comparison on Stable Diffusion (left) and FLUX (right).}
\label{fig:sd_flux_combined_v2}
\end{figure}

%%%%%%%%%%%%%%%%%%%%%%%%%%%%%%%%%%%%%%%%%%%%%%%%%%%%%%%%%%%%

\FloatBarrier
% \bibliographystyle{ACM-Reference-Format}
% \bibliography{references}

\end{document}